\documentstyle[epsbox,macros,twoside,11pt]{report}
 
\title{\LARGE \bf Anaphora Resolution in Japanese Sentences Using 
Surface Expressions and Examples}
\author{\rule{0mm}{60mm}{\Large Masaki Murata}}
\date{\rule{0mm}{40mm}December 1996}

\begin{document}
\renewcommand{\topfraction}{1}

\maketitle

\pagenumbering{roman}

\chapter*{Abstract}

\addcontentsline{toc}{chapter}{Abstract}

Anaphora resolution is one of the major problems 
in natural language processing. 
It is also 
one of the important tasks 
in machine translation and man/machine dialogue. 
We solve the problem by using 
{\it surface expressions} and {\it examples}. 
{\it Surface expressions} are the words in sentences 
which provide clues for anaphora resolution. 
{\it Examples} are linguistic data which 
are actually used 
in conversations and texts.  
The method using surface expressions and examples is 
a practical method.  

This thesis handles almost all kinds of anaphora. 
\begin{enumerate}
\item 
{ The referential property and number of a noun phrase}

\item 
{ Noun phrase direct anaphora}

\item 
{ Noun phrase indirect anaphora   }

\item 
\label{enum:abs_pro_ana}
{ Pronoun anaphora  }

\item
{ Verb phrase ellipsis}

\end{enumerate}
Pronoun anaphora 
has been investigated by many researchers 
\cite{Tanaka1} \, \cite{kameyama1} 
\cite{yamamura92_ieice} \cite{takada1} \cite{nakaiwa}. 
We used their results in addition to our new methods. 
In other areas of anaphora resolution, 
there are scarcely any empirical works and 
thus this thesis breaks new ground. 
In this thesis, 
the above five computer anaphora resolutions are described 
in Chapter \ref{chap:ref} through Chapter \ref{chap:0verb}. 

Chapter \ref{chap:ref}
shows that the referential property and number of noun phrases 
can be estimated fairly reliably by the words 
in Japanese sentences (surface expressions). 
The referential property and number of a noun phrase 
are basic factors in anaphora resolution. 
The system can grasp the outline of the referent of the noun phrase 
by using the referential property and number of a noun phrase. 
Many rules for the estimation of the referential property and number 
are written in forms similar 
to rewriting rules in expert systems with scores. 
We tested and verified 
the effectiveness of this method. 

Chapter \ref{chap:noun} describes a method for 
estimating the referent of a noun phrase in Japanese sentences 
using referential properties, modifiers, and possessors of noun phrases. 
In this analysis, 
referential properties are very important. 
For example, if the referential property of a noun phrase is definite, 
the noun phrase can refer to a previous noun phrase, and 
if the referential property of a noun phrase is indefinite, 
the noun phrase cannot refer to a previous noun phrase. 
Furthermore, we more precisely estimated referents of noun phrases 
using modifiers and possessors of noun phrases. 
We verified in our experiment 
the effectiveness of using 
referential properties, modifiers, and possessors of noun phrases. 

Chapter \ref{chap:indian} describes 
how to resolve indirect anaphora resolution. 
A noun phrase can indirectly refer to an entity that has 
already been mentioned before. For example, ``There is a house. 
The roof is white.'' indicates that ``the roof'' 
is associated with ``a house'', 
which was previously mentioned. 
When we analyze indirect anaphora, 
we need a case frame dictionary for nouns 
containing information about relationships 
between two nouns. 
But no noun case frame dictionary exists at present. 
Therefore, we used examples of ``X of Y'' 
and a verb case frame dictionary. 
We tested and verified that 
the information of ``X of Y''  
is useful 
when we cannot make use of a noun case frame dictionary. 
We also proposed how to construct a noun case frame dictionary 
from examples of ``X of Y''. 

Chapter \ref{chap:deno} describes
how to estimate the referent of a pronoun in Japanese sentences. 
In conventional work, 
semantic markers have been used 
for semantic constraints. 
We used examples for semantic constraints 
and showed in our experiments that 
examples are as useful as semantic markers. 
We also proposed many new methods for estimating referents of pronouns. 
We experimented with pronoun resolutions on some texts 
and verified the effectiveness of our methods. 

Chapter \ref{chap:0verb} describes 
the method of 
resolving verb phrase ellipsis using 
surface expressions and examples. 
When the referent of a verb phrase ellipsis appears in the sentences, 
the structure of the elliptical sentence is commonly 
in a typical form and 
the resolution is done by using surface expressions. 
When the referent does not exist in the sentences, 
the system resolved the elliptical sentence using examples. 
As the result of the experiment, 
we obtained a high accuracy rate. 

\tableofcontents
\listoffigures
\listoftables
\newpage
\pagenumbering{arabic}
\setcounter{page}{1}

\chapter{Introduction}

\section{Anaphora Resolution}

Natural language understanding is one of many researchers' dreams 
and has been investigated 
in many areas such as machine translation and man machine dialogue 
\cite{Winograd} \cite{Nagao_EBMT} \cite{Hirst86a} \cite{Hobbs88a}. 
Let us consider what natural language understanding is. 
Although machines will eventually understand 
natural language and be able to 
talk with humans, they cannot do so at present. 
The first step for natural language understanding is 
that the machine understands the structure of a sentence. 
It has been investigated 
in some areas (morphological analysis, 
syntax analysis, and case analysis), and 
good results have been obtained in some papers 
\cite{juman} \cite{csan2_ieice} \cite{Brill95a}. 
The next step is 
that the machine understands 
the object which a word refers to, 
which is called {\it anaphora resolution}. 
Although this has been investigated by many researchers, 
good results have still not been obtained. 
Therefore we devised 
a practical method to clarify 
how a word refers to an object. 

What kind of tasks are involved in the resolution of 
the object which a word refers to? 
At first, the system must recognize 
what a noun phrase refers to. 
It must also understand 
whether a noun phrase refers to a specified object or 
to a generic object. 
When a noun phrase partly relates to a noun phrase 
which has already been mentioned, 
the system must detect the relation. 
It must also understand 
what a pronoun or an ellipsis refers to. 

The above analyses are very important 
in machine translation and man machine dialogue. 
If an ellipsis is not resolved, 
machine translation and dialogue processing cannot be performed. 
If the reference of a word is resolved, 
the precision of generating articles ``the/a/an'' 
and pronouns ``I/you/he'' in machine translation will increase. 
In dialogue system, 
the number of 
counter questions to users 
is smaller and 
the processing is becoming more smooth.

The following is handled in this thesis: 
\begin{enumerate}
\item 
{\bf The referential property and the number of a noun phrase}

The system judges whether a noun phrase refers to a specific object 
or a generic object and estimates 
the number of the object. 

\begin{equation}
  \begin{minipage}[h]{11.5cm}
    \begin{tabular}[t]{l@{ }l@{ }l@{ }l}
\underline{HON}-TOIUNOWA & NINGEN-NO & SEICHOU-NI & KAKASEMASEN.\\
(book) & (human being) & (growth) & (be necessary)\\
\multicolumn{4}{l}{
(\underline{Books} are necessary for the growth of the human being. )}\\
\end{tabular}

\vspace{0.5cm}

(Desired solution:  ``HON'' refers to books in general. )
  \end{minipage}
\label{eqn:1c_hon_toiu1}
\end{equation}

\item 
{\bf Noun phrase direct anaphora}

The system estimates what a noun phrase represents. 

\begin{equation}
  \begin{minipage}[h]{11.5cm}
    \begin{tabular}[t]{lll}
{OJIISAN}-WA & JIMEN-NI & KOSHI-WO-OROSHIMASHITA.\\
(old man) & (ground) & (sit down)\\
\multicolumn{3}{l}{
({The old man} sat down on the ground.)}\\[0.3cm]
\end{tabular}
\begin{tabular}[t]{lll}
YAGATE & \underline{OJIISAN}-WA & NEMUTTE-SHIMAIMASHITA.\\
(soon) & (old man) & (fall asleep)\\
\multicolumn{3}{l}{
(\underline{The old man} soon fell asleep.)}\\
\end{tabular}

\vspace{0.5cm}

(Desired solution:  The underlined word ``OJIISAN'' refers to 
``OJIISAN'' in the first sentence. )
  \end{minipage}
\label{eqn:1c_ojiisan_jimen_meishi}
\end{equation}

\item 
{\bf Noun phrase indirect anaphora   }

The system estimates the object which a noun phrase indirectly refers to. 
In other words, 
the system detects the object which a noun phrase relates to in context. 

\begin{equation}
  \begin{minipage}[h]{11.5cm}
    \begin{tabular}[t]{lllll}
   KINOU & ARU  & HURUI & IE-NI & ITTA.\\
  (yesterday) & (a certain) & (old) & (house) & (go)\\
\multicolumn{5}{l}{
  (I went to an old house yesterday.)}
    \end{tabular}

\vspace{0.3cm}

  \begin{tabular}[t]{lll}
    \underline{YANE-WA} & HIDOI  & AMAMORIDE ... \\
  (roof) & (badly) & (be leaking)\\
\multicolumn{3}{l}{
  (\underline{The roof} was leaking badly and ... )}
    \end{tabular}

\vspace{0.5cm}

(Desired solution:  The underlined word ``YANE (roof)'' is 
the roof of ``IE (house)'' in the first sentence.)
  \end{minipage}
  \label{eqn:1c_doguse}
\end{equation}

\item 
\label{enum:1c_pro_ana}
{\bf Pronoun anaphora  }

The system estimates what a pronoun represents. 

\begin{equation}
  \begin{minipage}[h]{11.5cm}
    \begin{tabular}[t]{lll}
   KINOU & MIKAN-WO & KATTA .\\
  (yesterday) & (oranges) & (buy) \\
\multicolumn{3}{l}{
  (I bought some oranges yesterday.)}
    \end{tabular}

\vspace{0.3cm}

    \begin{tabular}[t]{lllll}
   TAROU-NO & IE-NI & ITTE & \underline{KORE}-WO & TABETA.\\
  (Taroo's) & (house) & (go) & (this)  & (eat)\\
\multicolumn{5}{l}{
  (I went to Taroo's house and ate \underline{them}.)}\\
    \end{tabular}

\vspace{0.5cm}

(Desired solution:  ``KORE'' refers to ``MIKAN''. )
  \end{minipage}
\label{eqn:mikan1}
\end{equation}

\item
{\bf Verb phrase ellipsis}

The system recovers an omitted verb phrase. 

\begin{equation}
  \begin{minipage}[h]{11.5cm}
    \begin{tabular}[t]{lll}
      SOU & UMAKU IKUTOWA & [OMOWANAI] .\\
      (so) & (succeed so well) & (I don't think)\\
\multicolumn{3}{l}{
  ([I don't think] it will succeed so well. )}
    \end{tabular}

\vspace{0.5cm}

(Desired solution:  ``OMOWANAI (I don't think)'' is recovered.)
  \end{minipage}
\label{eqn:souumaku1}
\end{equation}

\end{enumerate}

The area of ``\ref{enum:1c_pro_ana}. {Pronoun anaphora}'' 
has been investigated by many researchers 
\cite{Tanaka1} \, \cite{kameyama1} \, 
\cite{yamamura92_ieice} \, \cite{takada1}\\ \cite{nakaiwa}. 
We used their results in addition to our new methods. 
In the other areas of anaphora resolution, 
there are scarcely any empirical works. So 
this thesis breaks new ground in this regard.

\section{The Method Using Surface Expressions and Examples}


In this thesis, 
we have used much available information available for anaphora resolution. 
We emphasize {\it surface expressions} and {\it examples}. 

{\it Examples} are linguistic data which 
are actually used 
in conversations and texts.  
By using examples 
we can resolve many linguistic problems. 
For example, 
suppose that 
we want to clarify 
the thing 
which ``KORE (this)'' 
represents 
in the following sentences.  
\begin{equation}
  \begin{minipage}[h]{11.5cm}
    \begin{tabular}[t]{lll}
   KINOU & MIKAN-WO & KATTA .\\
  (yesterday) & (oranges) & (buy) \\
\multicolumn{3}{l}{
  (I bought some oranges yesterday.)}
    \end{tabular}

\vspace{0.3cm}

    \begin{tabular}[t]{lllll}
   TAROU-NO & IE-NI & ITTE & \underline{KORE}-WO & TABETA.\\
  (Taroo's) & (house) & (go) & (this)  & (eat)\\
\multicolumn{5}{l}{
  (I went to Taroo's house and ate \underline{them}.)}
    \end{tabular}
  \end{minipage}
\label{eqn:1c_mikan2}
\end{equation}
In this case, 
we gather examples such as 
``RINGO-WO TABERU (I eat apples)'' and
``KEIKI-WO TABERU (I eat cakes)'', 
and extract ``RINGO (apple)'' and ``KEIKI (cake)'' as the things 
which correspond to ``KORE (this)''. 
Since ``MIKAN (orange)'' is semantically similar to 
``RINGO (apple)'' and ``KEIKI (cake)'' 
in terms of food, 
we find that 
it is the antecedent of ``KORE (this)''. 
The method using examples has a wide application. 
If we discover examples  
which are analogous to the form of a problem, 
we can immediately use examples to solve the problem\footnote{
The method of using examples, which is called {\it Example-based approach}, 
was proposed for the purpose of machine translation \cite{Nagao_EBMT}. 
Although this method is used by many researchers in machine translation, 
it is not used in anaphora resolution to our knowledge. 
}. 

{\it Surface expressions} are the clue words in sentences 
which are used in anaphora resolution. 
For example, 
suppose that 
we want to clarify 
the thing 
which ``HON (book)'' refers to in the following sentences. 
\begin{equation}
  \begin{minipage}[h]{11.5cm}
    \begin{tabular}[t]{l@{ }l@{ }l@{ }l}
\underline{HON}-TOIUNOWA & NINGEN-NO & SEICHOU-NI & KAKASEMASEN.\\
(book) & (human being) & (growth) & (be necessary)\\
\multicolumn{4}{l}{
(\underline{Books} are necessary for the growth of human beings. )}\\
\end{tabular}
  \end{minipage}
\label{eqn:1c_hon_toiu2}
\end{equation}
Since there is a surface expression such as ``TOIUNOWA'' 
in this sentence, 
we find that ``HON (book)'' does not refer to 
a specific book but refers to books in general. 
Using surface expressions also has a wide application. 



The surface expressions and examples 
used in this work are as follows. 

\begin{itemize}
\item 
{\bf Surface Expression}
\begin{itemize}
\item 
  words
\item 
  part-of-speech
\item 
  syntax structure
\end{itemize}

\item 
{\bf Example}
\begin{itemize}
\item 
  the case frame of a verb phrase

\item 
  the semantic relation between two nouns. 

\item 
  example sentences 

\end{itemize}
\end{itemize}

\section{The Overview of Later Chapters}

This thesis describes how to resolve 
many problems in anaphora 
by using surface expressions and examples. 

Chapter \ref{chap:ref}
shows that the referential property and number of noun phrases 
can be estimated fairly reliably by the words (surface expressions) 
in Japanese sentences. 
The referential property and number of a noun phrase 
are basic factors in anaphora resolution. 
The system can grasp the outline of the referent of the noun phrase 
by using the referential property and number of a noun phrase. 
Many rules for the estimation of the referential property and number 
are written in forms similar 
to rewriting rules in expert systems with scores. 
We tested and verified 
the effectiveness of this method.

Chapter \ref{chap:noun} describes a method for 
estimating the referent of a noun phrase in Japanese sentences 
using referential properties, modifiers, and possessors of noun phrases. 
In the analysis, 
referential properties are very important. 
For example if the referential property of a noun phrase is definite, 
the noun phrase can refer to a previous noun phrase, and 
if the referential property of a noun phrase is indefinite, 
the noun phrase cannot refer to a previous noun phrase. 
Furthermore we estimated referents of noun phrases 
using modifiers and possessors of noun phrases more precisely. 
We made the experiment and verified that 
it is effective to use 
referential properties, modifiers, and possessors of noun phrases 
for estimating the referent of a noun phrase. 

Chapter \ref{chap:indian} describes 
how to resolve indirect anaphora resolution. 
A noun phrase can indirectly refer to an entity that has 
already been mentioned before. For example, ``There is a house. 
The roof is white.'' indicates that ``the roof'' 
is associated with ``a house'', 
which was mentioned in the previous sentence. 
When we analyze indirect anaphora, 
we need a case frame dictionary for nouns 
containing the information about relations between two nouns. 
But no noun case frame dictionary exists at present. 
Therefore, we used examples of ``X of Y'' 
and a verb case frame dictionary, instead. 
We made some experiments and verified that 
the information of ``X of Y''  
is useful 
when we cannot make use of a noun case frame dictionary. 
We also proposed how to construct a noun case frame dictionary 
from examples of ``X of Y''. 

Chapter \ref{chap:deno} describes
how to estimate the referent of a pronoun in Japanese sentences. 
In conventional work, 
semantic markers have been used 
for semantic constraints. 
We used examples for semantic constraints 
and show by our experiments that 
examples are as useful as semantic markers. 
We also proposed many new methods for estimating referents of pronouns. 
We experimented with pronoun resolutions on some texts 
and verified the effectiveness of our methods. 

Chapter \ref{chap:0verb} describes 
the method of 
resolving verb phrase ellipsis using 
surface expressions and examples. 
When the referent of a verb phrase ellipsis appears in the sentences, 
the structure of the elliptical sentence is commonly 
in a typical form and 
the resolution is done by using surface expressions. 
When the referent does not exist in the sentences, 
the system resolved the elliptical sentence using examples. 
As the result of the experiment, 
we obtained a high accuracy rate. 

Chapter \ref{chap:conc} is concluding remarks.

\mychapter
{\chapter{An Estimate of the Referential Property and the Number of Noun Phrase}}
{\chapter[The Referential Property and the Number]
{An Estimate of the Referential Property and the Number of Noun Phrase}}
\label{chap:ref}

\section{Introduction}

This chapter describes a method for the estimation of 
the referential property and number of a noun phrase
by using surface expressions. 
The referential property of a noun phrase represents 
how the noun phrase denotes the referent. 
The referential property is classified into three types: 
generic, definite and indefinite. 
A definite noun phrase refers to a given object. 
An indefinite noun phrase refers to a new object. 
They correspond to 
a noun phrase with a definite article 
and a noun phrase with an indefinite article 
in English, respectively. 
A generic noun phrase refers to all objects 
which the noun phrase denotes. 
The number of a noun phrase is 
the number of the referent denoted by the noun phrase. 
The number is classified into three types: 
singular, plural, and uncountable. 
The referential property and number of a noun phrase 
are basic factors in anaphora resolution. 
The system can grasp the outline of the referent of the noun phrase 
by using the referential property and number of a noun phrase. 
The referential property and number are also useful 
when the system generates the article 
in translating Japanese nouns into English. 

This chapter shows that 
the referential property and number of noun phrases
can be estimated fairly reliably by words (surface expressions) 
in the sentence.
Many rules for the estimation were written
in forms similar to rewriting rules in expert systems with scores.
Since this method uses scores, 
it is good to deal with 
vague problems like referential properties and numbers. 
We made the experiment estimating the referential property 
and number of the noun phrase 
and verified that our method is effective.

\section{Categories of Referential Property and Number}

\subsection{Categories of Referential Property}
Referential property of a noun phrase here means 
how the noun phrase denotes the subject. 
We classified noun phrases into the following three types 
from the referential property.

{\scriptsize
\[\rm \mbox{\normalsize  noun phrase}
 \left\{ \begin{array}{l}
     \rm \mbox{\normalsize  {\bf generic} noun phrase}\\
         \mbox{\normalsize  {\bf non generic} noun phrase}
            \left\{ \begin{array}{l}
              \rm \mbox{\normalsize  {\bf definite} noun phrase}\\
                  \mbox{\normalsize  {\bf indefinite} noun phrase}
            \end{array}
            \right.
\end{array}
\right.
\] 
}
\paragraph{Generic Noun Phrase}
A noun phrase is classified as generic 
when it denotes all members of the class of the noun phrase 
or the class itself of the noun phrase. 
For example, ``dogs'' in the following sentence is a generic noun phrase.
\begin{equation}
\mbox{\underline{Dogs} are useful.}
  \label{eqn:doguse}
\end{equation}
\paragraph{Definite Noun Phrase}
A noun phrase is classified as definite 
when it denotes a contextually non-ambiguous member 
of the class of the noun phrase. 
For example, ``the dog'' in the following sentence is a definite noun phrase.
\begin{equation}
\mbox{\underline{The dog} went away.}
  \label{eqn:thedogaway}
\end{equation}
\paragraph{Indefinite Noun Phrase}
An indefinite noun phrase denotes 
an arbitrary member of the class of the noun phrase.
For example, the following ``dogs'' is an indefinite noun phrase.
\begin{equation}
\mbox{There are three \underline{dogs}.}
  \label{eqn:threedogs}
\end{equation}

\subsection{Categories of Number}
The number of a noun phrase is 
the number of the subject denoted by the noun phrase.
Categories of number are as follows.
{\scriptsize
\[\rm \mbox{\normalsize  noun phrase}
   \left\{\begin{array}{l}
    \rm \mbox{\normalsize  {\bf countable} noun phrase}
        \left\{\begin{array}{l}
          \rm\mbox{\normalsize  {\bf singular} noun phrase}\\
             \mbox{\normalsize  {\bf plural} noun phrase}
          \end{array}
          \right.\\    
    \mbox{\normalsize  {\bf uncountable} noun phrase}
\end{array}
\right.
\]
}

\paragraph{Singular Noun Phrase}
A noun phrase is classified as singular 
when it denotes a singular member of the class of the noun phrase. 
For example, 
``a book'' in the following sentence is singular. 
\begin{equation}
\mbox{She brought \underline{a book}. }
\end{equation}

\paragraph{Plural Noun Phrase}
A noun phrase is classified as plural 
when it denotes plural members of the class of the noun phrase. 
For example, 
``some books'' in the following sentence is plural. 
\begin{equation}
\mbox{She brought \underline{some books}. }
\end{equation}

\paragraph{Uncountable Noun Phrase}

A noun phrase is classified as uncountable 
when it denotes part of the class of the noun phrase 
which cannot be divided into individuals. 
For example, 
``copper'' in the following sentence 
is used as material and uncountable. 
\begin{equation}
\mbox{\underline{Copper} conducts heat well.}
\end{equation}

\mysection
{\section{How to Estimate Referential Property and Number}}
{\section[How to Estimate]
{How to Estimate Referential Property and Number}}

\begin{figure}[p]

\fbox{
  \begin{minipage}[c]{13cm}
    \begin{center}
      ``KARE(he)-WA SONO(the)-BENGOSHI(lawyer)-NO(of) MUSUKO(son)-NO(of) 
      HITORI(one person)-DESU(is).'' \, \, \\
      (He is one of the sons of the lawyer.) 


 \center{(a):Japanese sentence}
    \end{center}
  \begin{minipage}[c]{13cm}

\baselineskip=0pt
\hspace*{6.6cm}\protect\verb! KARE(he)-WA----|  !\\
\hspace*{2.2cm}\protect\verb! SONO(the)----|!
\hspace*{4.28cm}\protect\verb! |  !\\
\hspace*{2.6cm}\protect\verb! BENGOSHI(lawyer)-NO(of)----| !
\hspace*{0.85cm}\protect\verb! |  !\\
\hspace*{6.2cm}\protect\verb! MUSUKO(son)-NO --|  !\\
\hspace*{7cm}\protect\verb! HITORI(one person)-DESU(is) !

\center{(b):Dependency structure of sentence(a)}
 \end{minipage}

\vspace{5mm}

 \begin{minipage}[c]{13cm}
\small
\baselineskip=12pt
\hspace*{0cm}\protect\verb+( <[noun common-noun +\_\verb+ +\_\verb+ `HITORI' `HITORI']+\\
\hspace*{0.34cm}\protect\verb+   [copula +\_\verb+ copula DESU-line-basic-form `DA' `DESU']+\\
\hspace*{0.34cm}\protect\verb+   [punctuation-mark period +\_\verb+ +\_\verb+ `. ' `. ']> +\\
\hspace*{0.34cm}\protect\verb+   ( <[noun common-noun +\_\verb+ +\_\verb+ `MUSUKO' `MUSUKO']+\\
\hspace*{0.83cm}\protect\verb+      [postpositional-particle noun-connection-postpositional-particle +\_\verb+ +\_\verb+ `NO' `NO']> +\\
\hspace*{0.83cm}\protect\verb+      ( <[noun common-noun +\_\verb+ +\_\verb+ `BENGOSHI' `BENGOSHI']+\\
\hspace*{1.32cm}\protect\verb+         [postpositional-particle noun-connection-postpositional-particle +\_\verb+ +\_\verb+ `NO' `NO']> +\\
\hspace*{1.32cm}\protect\verb+         ( <[demonstrative-adjective   +\_\verb+ +\_\verb+ +\_\verb+ `SONO' `SONO']> )))+\\
\hspace*{0.34cm}\protect\verb+   ( <[noun common-noun +\_\verb+ +\_\verb+ `KARE' `KARE']+\\
\hspace*{0.83cm}\protect\verb+      [postpositional-particle topic-marking-postposition +\_\verb+ +\_\verb+ `WA' `WA']+\\
\hspace*{0.83cm}\protect\verb+      [punctuation-mark komma +\_\verb+ +\_\verb+ `, ' `, ']> ))+
  \end{minipage}
\center{(c):Dependency structure representation of sentence(a)}

\caption{ Example of dependency structure representation}\label{fig:bengoshi_csan}

  \end{minipage}
}
\end{figure}

\begin{figure}[t]

\fbox{  
 \begin{minipage}[c]{13cm}
\small
\baselineskip=12pt
\hspace*{0.83cm}\protect\verb+      ( <[noun -] - >+\\
\hspace*{1.32cm}\protect\verb+         ( <[demonstrative-adjective   +\_\verb+ +\_\verb+ +\_\verb+ `SONO' `SONO']> ) - )+

  \caption{ An expression of the noun modified by ``SONO (the)''}
  \label{fig:sono_jap}
  \end{minipage}
}

\end{figure}

\noindent Heuristic rules for the referential property are given in the form:\\
%
{\small 
 ({\it condition for rule application}) \\
\hspace*{0.5cm}$\Longrightarrow$ \{ 
indefinite({\it possibility, value}) \,
definite({\it possibility, value}) \,
generic({\it possibility, value}) 
\}\\
}
\noindent Heuristic rules for the number are given in the form:\\
%
{\small 
 ({\it condition for rule application})  \\
\hspace*{0.5cm}$\Longrightarrow$ \{ 
singular({\it possibility, value}) \,
plural({\it possibility, value}) \,
uncountable({\it possibility, value})
\}
}

\noindent In {\it condition for rule application}, 
a surface expression is written in the form 
as in Figure~\ref{fig:sono_jap}. 
{\it Possibility} has value 1 when  the
categories: indefinite, definite, generic, singular, plural or uncountable,
are
possible in the context checked by the condition.  
Otherwise the {\it possibility} value is 0. 
{\it Value} means that 
a relative possibility value between 1 and 10 
(integer) is given according to 
the plausibility of the condition that the {\it possibility} is 1.
A larger value means the plausibility is high.

The rules are all heuristic so that the categories are not exclusive.  In a
certain conditional situation both indefinite and generic are possible, and also both singular and plural can co-exist.  In these cases, however, the 
possibility 
values may be different.

Several rules can be applicable to a specific noun in a sentence.  
In this case 
 the possibility values are added for individual categories
and the final decision of a 
category for a noun is done by 
the maximum possibility value. 
An example is given in Section \ref{subsec:abs_rule}.


When determining the referential property and number of nouns, 
the condition part is matched not for a word sequence 
but for a dependency structure of a sentence.
The dependency structure of a sentence (Figure~\ref{fig:bengoshi_csan}(a)) is 
shown in Figure~\ref{fig:bengoshi_csan}(b) 
which is represented as Figure~\ref{fig:bengoshi_csan}(c)\footnote{
This is the result transformed by the system\cite{csan2_ieice}.} 
to which the condition is checked. 
In heuristic rules, 
this expression can include a wild card(represented by ``\verb+-+'') 
which 
%
%
can match any partial dependency structure representations. 
For example, a noun modified by ``SONO(the)'' is expressed 
as in Figure~\ref{fig:sono_jap}. 
There are many other expressions 
such as regular expressions, AND-, OR-, NOT-operators, 
MODee-operator for checking modifier-modifyee relation and so on. 


\vspace{0.3cm}

\subsection*{Algorithm of the Determination of a Category}
The following steps are taken for the decision of a category 
for the referential property and the number. 
\begin{itemize}
\item[(1)]
  Sentences are 
  transformed into dependency structure representations.
\item[(2)]
  Decision is made for each noun from left to right in 
  the sentences transformed 
  into dependency structure representation. 
  This process allows the decision process to make use of 
  the referential property and the number already determined 
  (see \ref{subsec:abs_rule}(c)(d) for example). 
For each noun, the referential property is first determined, 
and then the number. 
This enables the utilization of referential property of a noun
 when analyzing the number of the noun 
(see \ref{subsec:num_rule}(3) for example).
In these processes 
all the applicable rules are used, 
{\it possibility} and {\it value} of each category are computed, 
and the category for the maximum value is obtained. 
An example of the result is shown in Figure~\ref{fig:bengoshi_noun}. 
We can also utilize the global information 
of a document to which a sentence belongs in 
the decision process. 
The {\it condition} part, for example, 
can check whether there are previous identical nouns. 
This information is useful for the determination 
of the referential property. 
\end{itemize}


\begin{figure}[t]
\fbox{
 \begin{minipage}[c]{13cm}
\small
\baselineskip=12pt
\hspace*{0cm}\protect\verb+( <[noun common-noun +\_\verb+ +\_\verb+ `HITORI' `HITORI' indefinite singular]+\\
\hspace*{0.34cm}\protect\verb+   [be-verb +\_\verb+ be-verb DESU-line-basic-form `DA' `DESU']+\\
\hspace*{0.34cm}\protect\verb+   [punctuation-mark period +\_\verb+ +\_\verb+ `. ' `. ']> +\\
\hspace*{0.34cm}\protect\verb+   ( <[noun common-noun +\_\verb+ +\_\verb+ `MUSUKO' `MUSUKO' definite plural]+\\
\hspace*{0.83cm}\protect\verb+      [postpositional-particle noun-connection-postpositional-particle +\_\verb+ +\_\verb+ `NO' `NO']> +\\
\hspace*{0.83cm}\protect\verb+      ( <[noun common-noun +\_\verb+ +\_\verb+ `BENGOSHI' `BENGOSHI' definite singular]+\\
\hspace*{1.32cm}\protect\verb+         [postpositional-particle noun-connection-postpositional-particle +\_\verb+ +\_\verb+ `NO' `NO']> +\\
\hspace*{1.32cm}\protect\verb+         ( <[referential-pronominal   +\_\verb+ +\_\verb+ +\_\verb+ `SONO' `SONO']> )))+\\
\hspace*{0.34cm}\protect\verb+   ( <[noun common-noun +\_\verb+ +\_\verb+ `KARE' `KARE' definite singular]+\\
\hspace*{0.83cm}\protect\verb+      [postpositional-particle sub-postpositional-particle +\_\verb+ +\_\verb+ `WA' `WA']+\\
\hspace*{0.83cm}\protect\verb+      [punctuation-mark komma +\_\verb+ +\_\verb+ `, ' `, ']> ))+

\vspace*{-8pt}

\caption{ The result of analyzing the sentence in Figure~2.1}\label{fig:bengoshi_noun}
  \end{minipage}
}
\end{figure}

\section{Heuristic Rules}

We have written 86 heuristic rules for the referential property and 48 
heuristic rules for the number.  More than half of these rules are just the 
implementation of grammatical properties
explained in standard grammar books of Japanese and English\cite{kanshi}\cite{gentei}\cite{hukusuu}, but 
there are many other heuristic rules which we have created. 
All of the rules are described in Appendix \ref{app:ref_num}. 
Some of the rules are given below.

\subsection{Heuristic Rules for Referential Property}\label{subsec:abs_rule}

\begin{enumerate}
\item[(1)] When a noun is modified by a referential pronoun, KONO(this), SONO(its), etc.,\\
then \,
\{
\mbox{indefinite}  (0, 0)\footnote{ 
(a, b) means the {\it possibility}(a) and the {\it value}(b). 
} \,
\mbox{definite}   (1, 2)  \,
\mbox{generic}  (0, 0)
\}\\
Examples: \underline{KONO}(This) \underline{HON-WA}(book) OMOSHIROI(interesting)\\
\hspace*{3cm} \underline{This book} is interesting.
\item[(2)] When a noun is accompanied by a particle (WA), and the 
predicate is in the past tense,\\
then \,
\{
\mbox{indefinite} (1, 0) \,
\mbox{definite}   (1, 3) \,
\mbox{generic} (1, 1)
\} \\
Example: \underline{INU-WA}(dog) MUKOUE(away there) {IKIMASHITA}(went)\\
\hspace*{3cm} \underline{The dog} went away.
\item[(3)] When a noun is accompanied by a particle (WA), and the 
predicate is in the present tense,\\
then \,
\{
\mbox{indefinite} (1, 0) \,
\mbox{definite}   (1, 2) \,
\mbox{generic} (1, 3)
\}\\
Example: \underline{INU-WA} YAKUNITATSU(useful) DOUBUTSU(animal)  DESU(is)\\
\hspace*{3cm} \underline{Dogs}\footnote{
Both ``a dog'' and ``the dog'' are possible because of the generic subject.
}
 are useful animals.
\item[(4)] When a noun is accompanied by a particle HE (to), MADE (up to) 
or KARA (from),\\
then \,
\{
\mbox{indefinite} (1, 0) \,
\mbox{definite}   (1, 2) \,
\mbox{generic} (1, 0)
\} \\
Example: KARE-WO(he) \underline{KUUKOU-MADE}(airport) MUKAE-NI(to meet) 
YUKIMASHOO(let us go)\\
\hspace*{3cm} Let us go to meet him at \underline{the airport}.

\item[(5)] 
  When a noun phrase is accompanied by a particle NO(of), 
  and it modifies a noun phrase
\footnote{
  When a noun phrase is accompanied by a particle NO(of), 
  it is not always a generic noun phrase. 
  But ``NO'' is likely to accompany 
  old information, 
  a noun phrase with ``NO'' is commonly 
  a definite noun phrase or a generic noun phrase. 
  Since we think that 
  a definite noun phrase 
  can be estimated by the other information, 
  we give a generic noun phrase a higher point value in this rule. 
}
, \\
\{
\mbox{indefinite} (1, 0) \,
\mbox{definite}   (1, 2) \,
\mbox{generic} (1, 3)
\} \\
Example: KARE-WA(he) \underline{KYOUIKU-NO}(education) KACHI-WO(value) 
NINSHIKI-SHITE-IMASEN(do not realize)\\
\hspace*{3cm} He doesn't realize the value of \underline{education}.

\end{enumerate}
There are many other expressions which give some clues 
for the referential property of nouns, such as 
(i) the noun itself,``CHIKYUU (the earth)''[definite], \\
``UCHUU (the universe)''[definite], etc.,
(ii) nouns modified by a numeral 
(Example: KORE-WA(this) ISSATSUNO(one) 
\underline{HON-DESU}(book)[indefinite]. 
(This is \underline{a book}.)), 
(iii) the same noun presented previously 
(Example: KARE-WA(he) JOUYOUSHA(car)-TO(and) TORAKKU-WO(truck) 
ICHIDAI-ZUTSU(by ones) MOTTEIMASUGA(have), 
\underline{JOUYOUSHA}-NIDAKE(car)[definite] 
HOKEN-WO- KAKETEIMASU(be insured). 
(He has a car and a truck, but only the car is insured.)),
(iv) adverb phrases, ``ITSUMO (always)'', ``NIHON-DEWA (in Japan)'', etc. 
(Example: NIHON-DEWA \underline{SHASHOU-WA}(conductor)[generic] 
JOUKYAKU (passenger)-NO(of)  
KIPPU-WO(ticket) SHIRABEMASU(check). 
(In Japan,\\
\underline{the conductor} checks the tickets of the passengers.)),
(v) verbs, ``SUKI(like)'', ``TANOSHIMU(enjoy)'', etc.
 (Example: WATASHI-WA(I) 
\underline{RINGO-GA}(apple) [generic] SUKI-DESU(like). 
(I like \underline{apples}.)). 

In the case of no clues, ``indefinite'' is given to a noun 
as a default value.

Since noun phrases which signify 
family relationships or body-parts such as ``{MUSUKO} (son)''
``{ONAKA} (stomach)'' 
are almost always definite noun phrases, 
we had better use the rule that 
when a noun phrase is a family relationship or a body-part, 
it is judged to be a definite noun phrase. 
Since this rule was made 
after the experiment 
on the test sentences 
in Section \ref{experiment}, 
we did not use it in the experiment. 
To test the effectiveness of this rule 
we made the experiment using this rule. 
The result is that 
the accuracy percentage decreased by 0.4\% in training sentences 
and 
increased by 3\% in test sentences. 
This is because 
in training sentences 
there are unexpectedly many cases that 
a noun phrase which indicates a relative or a body-part 
is used as non-definite. 
In common sentences, 
we should use this rule. 
We used 
{\it Bunrui Goi Hyou}\cite{BGH}
in judging whether 
a noun phrase means kin or body-part. 
The noun phrase the prefix of whose bgh code 
is ``121'' is regarded as relative, 
and ``157'' is regarded as body-part. 

\vspace{3mm}

Let us see an example which has several rule applications for the determination of the referential property of a noun.  ``KUDAMONO (fruit)'' in the following sentence is an example.

\begin{equation}
  \begin{minipage}[h]{13.7cm}
\small
    \begin{tabular}[t]{l@{ }l@{ }l@{ }l@{ }l@{ }l}
WAREWARE-GA & KINOU & TSUMITOTTA & \underline{KUDAMONO}-\underline{WA}
& AJI-GA & IIDESU\\
(We) & (yesterday) & (picked) & (fruit) &  (taste) & (be good)\\
\multicolumn{6}{l}{
(\underline{The fruit} that we picked yesterday tastes delicious.)}
\end{tabular}
  \end{minipage}
\label{eqn:kudamono}
\end{equation}
%

%
%
\hspace*{2cm} 

Seven rules are applied 
for the determination of the definiteness of this noun.  
These are the following:

\begin{itemize}
\item[(a)] When a noun is accompanied by WA, and the corresponding predicate 
has no past tense \\
(KUDAMONO-\underline{WA} AZI-GA \underline{IIDESU}), \\
then 
\{
\mbox{indefinite}  (1, 0) \,
\mbox{definite}  (1, 2)   \,
\mbox{generic}  (1, 3)
\}

\item[(b)] When a noun is modified by an embedded sentence which is in the past tense (TSUMITOTTA),\\
then \,
\{
\mbox{indefinite}  (1,  0) \,
\mbox{definite}    (1,  1) \, 
\mbox{generic}  (1,  0)
\}

\item[(c)] When a noun is modified by an embedded sentence which has a 
definite noun accompanied by WA or GA (WAREWARE-GA),\\
 then \,
\{
\mbox{indefinite}  (1,  0) \,
\mbox{definite}  (1,  1) \,
\mbox{generic}  (1,  0)
\}

\item[(d)] When a noun is modified by an embedded sentence which has a definite noun accompanied by a particle (WAREWARE-GA),\\
 then \,
\{
\mbox{indefinite}  (1,  0) \,
\mbox{definite}  (1,  1) \,
\mbox{generic}  (1,  0)
\}

\item[(e)] When a noun is modified by a phrase which has a pronoun 
(WAREWARE-GA),\\
then \,
\{
\mbox{indefinite}  (1,  0) \,
\mbox{definite}    (1,  1) \,
\mbox{generic}  (1,  0)
\}

\item[(f)] When a noun has an adjective as its predicate 
(KUDAMONO-WA AZI-GA \underline{IIDESU}),\\
then \,
\{
\mbox{indefinite}  (1,  0) \,
\mbox{definite}    (1,  3) \,
\mbox{generic}  (1,  4)
\}

\item[(g)] When a noun is a common noun 
(KUDAMONO),\\
then \,
\{
\mbox{indefinite}  (1,  1) \,
\mbox{definite}    (1,  0) \,
\mbox{generic}  (1,  0)
\}

\end{itemize}

As the result of the application of all these rules, 
we obtained the final score of 
\{
\mbox{indefinite}  (1,  1) \,
\mbox{definite}    (1,  9) \,
\mbox{generic}  (1,  7)
\}
for KUDAMONO, 
and ``definite'' is given as the decision.

\subsection{Heuristic Rules for Number}\label{subsec:num_rule}

\begin{enumerate}
\item[(1)] When a noun is modified by SONO(its), ANO(that), KONO(this),\\
then \, 
\{
\mbox{singular}  (1,  3) \,
\mbox{plural}    (1,  0) \,
\mbox{uncountable}  (1,  1)
\}\\
Example: \underline{ANO}(that) \underline{HON-WO} (book) KUDASAI (give me)\\
\hspace*{3cm} Give me \underline{that book}.
\item[(2)] 
When a noun is accompanied by a particle WA, GA, MO, WO, and there is a numeral x which modifies the predicate of a sentence, and\\
if \, x = 1 , \, then \, 
\{
\mbox{singular}  (1,  2) \,
\mbox{plural}    (1, 0) \,
\mbox{uncountable}  (1,  0)
\} \\
if \, x $\geq$ 2 , \, then \, 
\{
\mbox{singular}  (1,  0) \,
\mbox{plural}    (1,  2) \,
\mbox{uncountable}  (1,  0)
\} \\
Example: \underline{RINGO-WO}(apple) NIKO(two) TABERU(eat)\\
\hspace*{3cm} I eat two \underline{apples}.
\item[(3)]
When a predicate, SUKI(like), TANOSHIMU(enjoy), etc. has a generic noun as an object, and the noun is accompanied by GA(for SUKI), or WO(for TANOSHIMU),\\
then 
\{
\mbox{singular}  (1,  0) \,
\mbox{plural}    (1,  2) \,
\mbox{uncountable}  (1,  0)
\}\\
Example: WATASHI-WA(I) \underline{RINGO-GA}(apple) SUKI-DESU(like)\\
\hspace*{3cm} I like \underline{apples}.

\end{enumerate}

There are many other expressions 
which determine the number of a noun, 
such as (i) nouns modified by a numeral 
(Example: KORE-WA(this) ISSATSUNO(one) \underline{HON-DESU}(book)[singular]. 
(This is \underline{a book}.)), 
(ii) verbs such as ATSUMERU(collect), AFURERU(be full with), 
(Example: WATASHI-WA(I) NEKO-NO(about cat) \underline{HON-WO}(book)[plural] 
 ATSUMETEIMASU(collect). (I collect \underline{books} on cats.))
(iii) adverbs such as NANDO-DEMO(as many times as ...), 
IKURA-DEMO(as much ...) 
(Example: RIYUU-WA(reason)[plural] IKURA-DEMO(as much ...) SHIMESEMASU(give). 
(I can give you a number of reasons.)). 

In the case of no clues, ``singular'' is given as a default value.

\section{ Experiments and Results}
\label{experiment}

Experiments for the determination of the referential property 
and for the number were 
done in the following three texts: typical example sentences 
in a grammar book ``Usage of the English Articles''\cite{kanshi}, 
the complete text of a Japanese popular folk tale 
``The Old Man with a Lump''\cite{kobu},
 a small fragment of an essay ``TENSEI JINGO''. 
The rules were written by referring to these sentences 
which have good English translations.  
These sentences can be regarded as a training set.  
The results of the experiments are shown in Table \ref{tab:learning}.  
Here ``correct'' means that 
the result was correct. 
``Reasonable'' means 
that the result is given, for example, 
as non-generic but the correct answer was definite, etc. 
``Partially correct'' means that the result was 
included in the correct answer.  
``Undecidable'' means that we could not judge 
which category is correct. 
We obtained 85.5\% success rate for the determination 
of the referential properties 
and 89.0\% success rate for the numbers for all these training sentences.  
The scores of these tables show that the heuristic rules are 
effective and applicable to these sentences. 

The modification and addition of rules 
in the experiment of training sentences 
were performed as follows: 
    \begin{enumerate}
    \item 
      \label{enum:error_mod}
      The modification and addition 
      of rules were performed by examining 
      errors. 
      In other words, 
      we looked at the surface expressions 
      near a noun phrase which was incorrectly interpreted, 
      and considered whether 
      we can make a new rule. 
      We also checked 
      whether we could correct 
      this error 
      by modifying the condition and the point of the rule. 

    \item 
      After some modifications and additions of rules were performed, 
      we checked whether the overall precision was higher or lower. 
      When the overall precision was higher, 
      we formally adopted the modifications and additions 
      which were performed in \ref{enum:error_mod}. 
      When the overall precision was lower, 
      we did not perform the modifications and additions, 
      and repeated examinations in \ref{enum:error_mod}. 
    \end{enumerate}

    In addition to this procedure, 
    when we roughly examined some errors and 
    found out a rule by which we could correct these errors, 
    we added the rule to the rule set. 
    Moreover, when we were not certain whether we should add a certain rule, 
    we listed all parts which were used by the rule 
    and decided by looking at them as a whole. 

\begin{table}[t]


\begin{center}

{\small 
\caption{Training sentences}\label{tab:learning}

\vspace*{0.4cm}

\begin{tabular}[c]{|l@{ }|@{\,}r@{ }|@{\,}r@{ }|@{\,}r@{ }|@{\,}r@{ }|@{\,}r@{ }|@{\,}r@{ }|@{\,}r@{ }|@{\,}r@{ }|@{\,}r@{ }|@{\,}r@{ }|} \hline
 & \multicolumn{5}{c|}{Referential property}  &  \multicolumn{5}{c|}{Number}  \\ \hline 
\multicolumn{1}{|c|}{value}  &  indef  &  def &  gener &  other & total &  singl   &  plural    &  uncount &  other &   total  \\ \hline
\multicolumn{11}{|c|}{Usage of the English Articles(140 sentences, 380 nouns)} \\ \hline 
   correct   &      96  &     184  &      58  &       1  &     339  &    274 &    32 &      18  &      25  &  349   \\ 
reasonable&       0  &       3  &       1  &       0  &       4  &       1  &       1  &       1  &       0  &       3  \\ 
  partially correct  &       0  &       0  &       0  &       0  &       0  &       0  &       0  &       0  &      11  &      11  \\ 
  incorrect  &       4  &      25  &       7  &       1  &      37  &       3  &      10  &       0  &       4  &      17   \\ 
\hline 
\% of correct&    96.0  &    86.8  &    87.9  &    50.0  &    89.2  &    98.6  &    74.4  &    94.7  &    62.5  &    91.8    \\ \hline 
\multicolumn{11}{|c|}{The Old Man with a Lump(104 sentences, 267 nouns)} \\ \hline 
   correct   &      73  &     140 &       6  &       1  &    222  &    205  &     24  &       5  &       0  &   234   \\ 
reasonable&       3  &       4  &       0  &       0  &       7 &       2  &       0  &       0  &       0  &       2    \\ 
  partially correct  &       0  &       0  &       0  &       0  &       0   &       0  &       0  &       0  &       7  &       7  \\ 
  incorrect  &      11  &      23  &       4  &       0  &      38    &       1  &      22  &       1  &       0  &      24   \\ 
\hline
\% of correct&   83.9  &   84.0  &   60.0  &  100.0  &   83.2  &   98.7  &   52.2  &   83.3  &    0.0  &   87.6    \\ \hline 
\multicolumn{11}{|c|}{an essay ``TENSEI JINGO''(23 sentences, 98 nouns)} \\ \hline 
   correct   &      25  &     35 &      16  &       0  &    76  &      64  &   13 &       0  &       3  &   80    \\ 
reasonable&       0  &       4  &       2  &       0  &       6  &       2  &       1  &       0  &       0  &       3   \\ 
  partially correct  &       0  &       0  &       0  &       0  &       0   &       0  &       0  &       0  &       6  &       6  \\ 
  incorrect  &       5  &      10  &       1  &       0  &      16   &       1  &       6  &       1  &       1  &       9  \\ 
\hline
\% of correct&    83.3  &    71.4  &    84.2  &    -----  &    77.6   &    95.5  &    65.0  &     0.0  &    30.0  &    81.6   \\ \hline \hline 
 & & & & & & & & & &\\[-0.5cm]
average  & & & & & & & & & &\\[-0.1cm]
\hspace*{3mm}\% of appearance &  29.1  &  57.7   &  12.8   &  0.4  &  100.0 &  74.2  &  14.6   &  3.5  &  7.7   & 100.0  \\[-0.1cm]

\hspace*{3mm}\% of correct &  89.4  &  84.0   &  84.2   &  66.7   &  85.5  &  98.2  &  63.3   &  88.5  &  49.1   &  89.0   \\ \hline 
\end{tabular}
}
\end{center}

\end{table}

\begin{table}[t]
\begin{center}

{\small 

\caption{Test sentences}\label{tab:test}

\vspace*{0.4cm}

\begin{tabular}[c]{|l@{ }|@{\,}r@{ }|@{\,}r@{ }|@{\,}r@{ }|@{\,}r@{ }|@{\,}r@{ }|@{\,}r@{ }|@{\,}r@{ }|@{\,}r@{ }|@{\,}r@{ }|@{\,}r@{ }|} \hline
 & \multicolumn{5}{c|}{Referential property}  &  \multicolumn{5}{c|}{Number}  \\ \hline 
\multicolumn{1}{|c|}{value}  &  indef  &  def &  gener &  other & total &  singl   &  plural    &  uncount &  other &   total  \\ \hline
\multicolumn{11}{|c|}{a folk tale ``TSURU NO ONGAESHI''(263 sentences, 699 nouns)} \\ \hline 
   correct   &     109  &    363  &      13  &      10  &   495   &     610  &      13  &       1  &       1  &     625  \\ 
reasonable&       6  &      25  &       0  &       0  &      31  &      12  &       2  &       0  &       0  &      14   \\ 
  partially correct  &       0  &       0  &       0  &       0  &       0  &       0  &       0  &       0  &       1  &       1   \\ 
  incorrect  &      32  &     135  &       6  &       0  &     173   &       2  &      20  &      37  &       0  &      59  \\ 
\hline
  \% of correct   &    74.2  &    69.4  &    68.4  &   100.0  &    70.8   &    97.8  &    37.1  &     2.6  &    50.0  &    89.4    \\ \hline 
\multicolumn{11}{|c|}{an essay ``TENSEI JINGO''(75 sentences, 283 nouns)} \\ \hline 
   correct   &      75  &    81  &      16  &       0  &   172  &     197  &      13  &       2  &       3  &     215   \\ 
reasonable&       8  &       9  &       1  &       0  &      18 &       3  &       1  &       0  &       0  &       4    \\ 
  partially correct  &       0  &       0  &       0  &       0  &       0   &       0  &       0  &       0  &       3  &       3     \\ 
  incorrect  &      33  &      51  &       9  &       0  &      93  &       3  &      55  &       3  &       0  &      61    \\ 
\hline
  \% of correct   &    64.7  &    57.5  &    61.5  &    ----- &    60.8   &    97.0  &    18.8  &    40.0  &    50.0  &    76.0    \\ \hline 
\multicolumn{11}{|c|}{Pacific Asia in the Post-Cold-War World(22 sentences, 192 nouns)} \\ \hline 
   correct   &      21  &    108  &      11  &       2  &  142 &     157  &       6  &       1  &       1  &     165   \\ 
reasonable&       6  &       7  &       0  &       0  &      13 &       3  &       0  &       0  &       0  &       3   \\ 
  partially correct  &       0  &       0  &       0  &       0  &       0   &       0  &       0  &       0  &       0  &       0   \\ 
  incorrect  &      11  &      24  &       2  &       0  &      37  &       3  &      20  &       1  &       0  &      24    \\ 
\hline
  \% of correct   &    55.3  &    77.7  &    84.6  &   100.0  &    74.0   &    96.3  &    23.1  &    50.0  &   100.0  &    85.9   \\ \hline \hline
 & & & & & & & & & & \\[-0.5cm]
average  & & & & &  & & & & &\\[-0.1cm]
\hspace*{3mm}\% of appearance&  25.6  &  68.4  &  4.9   &  1.0   &  100.0 &  84.3  &  11.1   &  3.8   &  0.8   &  100.0   \\[-0.1cm] 
\hspace*{3mm}\% of correct &  68.1   &  68.7 &  69.0 &  100.0  &  68.9  & 97.4 &  24.6  & 8.9  &  55.6  & 85.6  \\ \hline
\end{tabular}
}
\end{center}
\end{table}

To test the quality of these rules, 
we applied them 
to the following three texts: a Japanese popular folk tale 
``TSURU NO ONGAESHI'' \cite{kobu},
 three small fragments of an essay ``TENSEI JINGO'',
 ``Pacific Asia in the Post-Cold-War World''
(A Quarterly Publication of The International House of Japan Vol.12, 
No.2 Spring 1992).
These test sentences have good English translations.  
The results are shown in Table \ref{tab:test}.
%
%
The success rates for the referential property and the number decreased 
down to 68.9\% and 85.6\% respectively by these test sentences. 
These scores show, however, that the rules are still effective.


\section{Discussion}

\subsubsection{Discussion on the Experiment of the Referential Property}

With respect to referential property, 
the success rate was 85.5\% 
in the training sentences 
by which we elaborated our rule set. 
There was no category which was very bad. 
This indicates that 
our method of using surface expressions 
can estimate the referential properties of many noun phrases.  

The success rate 
was 68.9\% in the test sentences 
on which we fixed our rule set. 
All the categories' success rates were 
uniformly good and more than 60\%. 
The appearance of the definite noun phrase was 74.8\% 
in the experiment of ``TSURU NO ONGAESHI''. 
Therefore, 
if we make rules which handle each noun phrase
as a definite noun phrase, 
the success rate becomes 74.8\%, 
and becomes higher than the success rate of 70.8\% 
in the experiment. 
But this is not good, 
because the success rates of 
indefinite noun phrases and generic noun phrases become 0\%. 
We think that 
it is important that 
all the categories' success rates are 
uniformly good. 

The success rate in training sentences is not good. 
If we modify the rule set, 
the success rate will easily rise. 
But when we try to increase 
the success rate in new sentences, 
it may be necessary to continue to make new rules for new sentences. 

Table \ref{fig:false_tei} and Table \ref{fig:false_sousyou} 
are examples 
which are analyzed incorrectly, 
even if we modify the rule set. 
Table \ref{fig:false_tei} is a set of examples 
which are analyzed incorrectly because 
no key surface expression exists 
and a noun phrase is a definite noun phrase. 
To solve this problem, 
we need the information 
on contexts and situations. 

Table \ref{fig:false_sousyou} are examples 
which are analyzed incorrectly 
when a noun phrase is a generic noun phrase. 
We describe the reason for the error in each example.

\begin{table}[t]
\small

\caption{Examples of definite noun phrases 
analyzed incorrectly (noun phrases whose head words are underlined)
}\label{fig:false_tei}

{

\fbox{
  \begin{minipage}[h]{13cm}
\small

  \begin{tabular}[t]{@{}r@{ }lll}
(1) & KARE-WA & \underline{SHACHOU}-NO & ONIISAN-DESU.\\
& (he) & (president) & (brother) \\
&\multicolumn{3}{l}{
(He is the brother of \underline{the president}.)}\\
\end{tabular}

  \begin{tabular}[t]{@{}r@{ }lllll}
(2) & JHON-WA & \underline{KURASU}-NO & NAKADE & ICHIBAN & SEGATAKATAI.\\
& (John) & (class) & (in) & (the most)  & (tall) \\
&\multicolumn{5}{l}{
(John is the tallest in \underline{my class}.)}\\
\end{tabular}

  \begin{tabular}[t]{@{}r@{ }lllll}
(3) & KANOJO-WA & \underline{TEIBURU}-WO & HUKU-NONI & HUKIN-WO & TSUKATTA.\\
& (she) & (table) & (to dust) & (cloth) & (use) \\
&\multicolumn{5}{l}{
(She used a cloth to dust \underline{the table}.)}\\
\end{tabular}

  \begin{tabular}[t]{@{}r@{ }l@{ }l@{ }l@{ }l}
(4) & \underline{SHIGOTO}-DE & MUZUKASHII-TOKOROGA & ATTA-GA & 
KOKUHUKUSHITA.\\
& (work) & (difficulty) & (exist) & (overcome)  \\
&\multicolumn{4}{l}{
(I overcame a difficulty in my \underline{work}.)}\\
\end{tabular}

  \begin{tabular}[t]{@{}r@{ }lllll}
(5) & WATASHI-WA & \underline{SENSEI}-TO & ONAJI & HON-WO & MOTTE-IMASU.\\
& (I) & (teacher) & (same) & (book)  & (have) \\
&\multicolumn{5}{l}{
(I have the same book as \underline{the teacher} has.)}\\
\end{tabular}

  \begin{tabular}[t]{@{}r@{ }lll}
(6) & KURUMA-WA & \underline{MICHI}-NO-WAKINI & CHUUSHA-SHITE-ARIMASU.\\
& (car) & (along the street) & (be parked) \\
&\multicolumn{3}{l}{
(Cars are parked along \underline{the street}. )}\\
\end{tabular}

  \begin{tabular}[t]{@{}r@{ }llll}
(7) & JONSONKYOUJU-WA & \underline{GAKKAI}-DE & \underline{RONBUN}-WO & YOMIMASHITA.\\
& (Professor Johnson) & (convention) & (technical paper) & (read)\\
&\multicolumn{3}{l}{
(Professor Johnson read \underline{his paper} at \underline{the convention}.)}\\
\end{tabular}

  \end{minipage}
}

}

\end{table}

\begin{table}[t]
\small

\caption{Examples of generic noun phrases  
incorrectly analyzed (underlined noun phrases)}\label{fig:false_sousyou}

{

  \begin{center}

\begin{tabular}{|p{13cm}|} \hline

 (1)When the noun phrase 
 is incorrectly judged as definite because 
 it is modified an embedded sentence \\

  \begin{tabular}[t]{llll}
\underline{SOREJITAI-WO} & \underline{MAMOROU-TO-SHINAI} & \underline{BUNKA-WA} & HOROBIMASU.\\
(itself) & (do not defend) & (culture) & (die)\\
\multicolumn{4}{l}{
(\underline{A culture that does not defend itself} will die. )}\\
\end{tabular}

\vspace{3mm}

(2)When the noun phrase 
 is incorrectly judged as definite because 
the predicate is in the tense \\

  \begin{tabular}[t]{llll}
\underline{CHUUGOKUJIN-WA} & DOKUJI-NO & MOJI-WO & HATSUMEI-SHITA.\\
(Chinese) & (own) & (writing system) & (invent) \\
\multicolumn{4}{l}{
(\underline{The Chinese} invented 
their own writing system.)}\\
\end{tabular}

\vspace{3mm}

(3)When the noun phrase 
 is incorrectly judged as indefinite because 
it is followed by a copula ``DA''\\

  \begin{tabular}[t]{llll}
NIHON-NO & SHAKAI-DEWA & CHICHIOYA-WA & \underline{KACHOU-DESU}.\\
(Japanese) & (society) & (father) & (the head of the house hold) \\
\multicolumn{4}{l}{
(In Japanese society, 
the father is \underline{the head of the household}.)}\\
\end{tabular}

\vspace{3mm}

(4)When the noun phrase 
 is incorrectly judged as indefinite because 
there is no clue\\

  \begin{tabular}[t]{l@{ }l@{ }l@{ }l@{ }l}
\underline{TABEMONO-GA} & OISHIKEREBA & OISHIIHODO, & TAKUSAN & TABEMASU.\\
(food) & (good) & (the more) & (much) & (eat) \\
\multicolumn{5}{l}{
(The better \underline{the food} is, 
the more I eat.)}\\
\end{tabular}\\\hline
\end{tabular}
  \end{center}
}

\end{table}

There were some cases 
where it is difficult 
to analyze using only surface expressions. 
\begin{equation}
  \begin{minipage}[h]{11.5cm}
    \begin{tabular}[t]{llllll}
KORE-WA & \underline{KARE-KARA} & \underline{KARITA} & \underline{JISHO} & DESU.\\
(this)  & (from him) & (borrow) & (dictionary) & (be)\\
\multicolumn{6}{l}{
(This is \underline{the dictionary that 
I borrowed from him}. )}\\
\end{tabular}
  \end{minipage}
\label{eqn:kari_jisyo}
\end{equation}

In this example, 
since 
``WATASHI-GA KARE-KARA KARITA JISHO (the dictionary that 
I borrowed from him)'' is 
modified by the embedded sentence, 
it was judged to be a definite noun phrase. 
But when 
``WATASHI (I)'' borrowed some dictionaries from ``KARE (him)'' 
and ``WATASHI-GA KARE-KARA KARITA JISHO (the dictionary that 
I borrowed from him)'' is one of them, 
it is an indefinite noun phrase. 
Therefore 
it is difficult for 
the system to judge whether a noun phrase is
 a definite noun phrase or an indefinite noun phrase 
unless the system has certain information. 

\subsubsection{Discussion on the Experiment of the Number}

The success rate was 89.0\% 
in training sentences. 
But the success rate of ``plural'' 
was low. 

The success rate was 85.6\% 
in test sentences. 
But the success rates of ``plural'' and ``uncountable'' 
were low. 

The following example is for when 
the plural noun phrase was analyzed incorrectly. 
\begin{equation}
  \begin{minipage}[h]{11.5cm}
    \begin{tabular}[t]{llllll}
CHUUMON-SHITA & \underline{KENCHIKU-ZAIRYOU}-GA & KIMASHITA.\\
(order) & (building material) & (come) \\
\multicolumn{6}{l}{
(\underline{The building materials} you ordered have come in.)}\\
\end{tabular}
  \end{minipage}
  \label{eqn:kentiku}
\end{equation}
The reason for the error is 
that there is no clue word. 
To judge this case to be ``plural'', 
the system must judge 
it by the word ``KENCHIKU-ZAIRYOU (building material)'' itself. 
But ``KENCHIKU-ZAIRYOU (building material)'' is not always ``plural''. 

The following example is a plural noun phrase analyzed properly 
without quantifiers. 
\begin{equation}
  \begin{minipage}[h]{11.5cm}
    \begin{tabular}[t]{lll}
SONO JIKO-NO-ATO & \underline{YAJIUMA}-GA & ATSUMATTE-KIMASHITA.\\
(after the accident)  & (people) & (gather)  \\
\multicolumn{3}{l}{
(\underline{People gathered after the accident})}\\
\end{tabular}
  \end{minipage}
   \label{eqn:jiko}
\end{equation}
``YAJIUMA'' was judged to be ``plural'' 
using the verb ``GA ATSUMARU (gather)''. 
If we make such a rule, 
we can occasionally analyze the number of a noun phrase 
which is not modified by a quantifier. 


\begin{table}[t]
\small
  \leavevmode
    \caption{Examples of verbs which may be used 
      in the estimation of the number of the noun phrase}
  \begin{center}
    \label{tab:num_verb}
\begin{tabular}{|p{13cm}|}\hline
ABIRU (pour water), HUKIKAKERU (sprinkle), 
MABUSU (cover), WAKIDERU (well up), 
SOROERU (put in order), UMORERU (be buried), 
MORERU (leak), KOBORERU (drop, spill), 
MURAGARU (crowd), 
NOMU (drink)\\\hline
\end{tabular}
\end{center}
\end{table}

After the experiment on the training sentences and test sentences, 
we examined the rule 
using verbs such as ``ATSUMARU (gather)'', ``NARABERU (put in order)'', 
and ``ABIRU (pour water)''.
We gathered about 300 verbs from ``Bunrui Goi Hyou'' \cite{BGH} 
which can be used in the estimation of the number. 
The examples are shown in Table \ref{tab:num_verb}. 
We also checked the occurrence of the 
noun phrases which can be analyzed properly by using these verbs. 
There were 21 noun phrases 
in the sentences (526 sentences, 2680 noun phrases, essays of two months) 
of essays ``TENSEI JINGO'' 
which were analyzed properly by the syntactic parser. 
This frequency was low. 
But since the number of the noun phrase 
which can be analyzed properly still increases, 
we must use the rule using verbs 
as in Table \ref{tab:num_verb} for the estimation of the number. 

\section{Summary of this Chapter}

We obtained the correct recognition scores of 85.5\% and 89.0\% 
in the estimation of referential property and number 
respectively for the sentences
which were used for the construction of our rules.
We tested these rules for some other texts,
and obtained the scores of 68.9\% and 85.6\% respectively.

There are two problems 
in the estimation of the referential property. 
One is that although
a human can easily recognize the referential property from the situation, 
the system cannot estimate the referential property.  
If we can make use of situational information, 
we can analyze the problem properly. 

Another problem is with respect to generic noun phrases. 
A generic noun phrase is difficult to be defined 
to discriminate other categories. 
The category may have to be reconstructed. 

With respect to the number of a noun phrase, 
it is easily estimated, 
if it is modified by some surface expressions 
such as quantifiers. 
Since 
a noun phrase is not always modified by quantifiers, 
the estimation of the number is not so easy. 
There are some cases 
when the number is estimated properly 
by verbs such as ``ATSUMERU (gather)'' and 
adverbs such as ``IKURADEMO (as much as one likes)''. 
 
\chapter{An Estimate of Referent of Noun Phrases}
\label{chap:noun}

\section{Introduction}
\label{sec:intro}

This chapter describes 
how to estimate the referent of a noun phrase 
in Japanese sentences. 
It is important to clarify 
referents of noun phrases 
in machine translation. 
For example, 
since the two ``OJIISAN (old man)'' in the following sentences 
have the same referent,
the second ``OJIISAN (old man)'' should be pronominalized 
in translation into English. 
\begin{equation}
  \begin{minipage}[h]{10.5cm}
    \begin{tabular}[t]{lll}
\underline{OJIISAN}-WA & JIMEN-NI & KOSHI-WO-OROSHIMASHITA.\\
(old man) & (ground) & (sit down)\\
\multicolumn{3}{l}{
(\underline{The old man} sat down on the ground.)}\\[0.3cm]
\end{tabular}
\begin{tabular}[t]{lll}
YAGATE & \underline{OJIISAN}-WA & NEMUTTE-SHIMAIMASHITA.\\
(soon) & (old man) & (fall asleep)\\
\multicolumn{3}{l}{
(\underline{He (= the old man)} soon fell asleep.)}\\
\end{tabular}\\
  \end{minipage}
\label{eqn:ojiisan_jimen_meishi}
\end{equation}
When dealing with a situation like this, 
it is necessary that a machine translation system should recognize that 
two ``OJIISAN (old man)'' have the same referents. 
In this chapter, 
we propose a method for determining referents of noun phrases 
using (1)referential properties of noun phrases, 
(2)modifiers in noun phrases,
 and (3)possessors of objects denoted by noun phrases. 

For languages that have articles like English, 
we can guess by using articles 
whether two noun phrases refer to each other or not. 
In contrast, 
for languages that have no articles like Japanese, 
it is difficult to decide 
whether two noun phrases refer to each other. 
We estimated referential properties of noun phrases 
that correspond to articles 
shown in Chapter \ref{chap:ref}.
By using these referential properties, 
our system determines referents of noun phrases 
in Japanese sentences. 
Noun phrases are classified by referential property into 
generic noun phrases, 
definite noun phrases, 
and indefinite noun phrases. 
When the referential property of a noun phrase is a definite noun phrase, 
the noun phrase can refer to a noun phrase that has already appeared. 
When the referential property of a noun phrase 
is an indefinite noun phrase 
or a generic noun phrase, 
the noun phrase cannot refer to a noun phrase that has appeared already. 

It is insufficient to determine referents of noun phrases 
only using referential property. 
This is 
because 
even if the referential property of a noun phrase is a definite noun phrase, 
the noun phrase does not refer to a noun phrase 
which has a different modifier or a possessor. 
Therefore, 
we also use modifiers and possessors of noun phrases 
in determining referents of noun phrases. 

\section{Referential Property of Noun Phrase}

The following is an example of noun phrase anaphora. 
\begin{equation}
  \begin{minipage}[h]{11.5cm}
    \begin{tabular}[t]{lllll}
    \underline{OJIISAN} & TO & OBAASAN-GA & SUNDEORIMASHITA.\\
    (an old man)&(and)&(an old woman)&(lived)\\
\multicolumn{5}{l}{
  (There lived \underline{an old man} and an old woman.)}
    \end{tabular}
    \begin{tabular}[t]{llll}
      \underline{OJIISAN}-WA & YAMA-HE & SHIBAKARI-NI & IKIMASHITA.\\
      (old man)  & (mountain) & (to gather firewood) & (go)\\
\multicolumn{4}{l}{
  (\underline{The old man} went to the mountains to gather firewood.)}
    \end{tabular}
  \end{minipage}
\label{eqn:ojiisan_obaasan_meishi}
\end{equation}
``OJIISAN (old man)'' in the first sentence and 
``OJIISAN (old man)'' in the second sentence 
refer to the same old man, 
and they are in anaphoric relation. 

When the system analyzes the anaphoric relation of noun phrases like this, 
the referential properties of noun phrases are important. 
Referential property of a noun phrase here means 
how the noun phrase denotes the referent. 
Since the second ``OJIISAN (old man)'' 
has the referential property of the definite noun phrase,  
indicating that it refers to the contextually non-ambiguous object, 
the system can recognize 
that it refers to the first ``OJIISAN (old man). 
The referential property plays an important role 
in clarifying anaphoric relation. 

We classified noun phrases 
by referential property 
into the following three types 
as shown in Chapter \ref{chap:ref}. 

{\scriptsize
\[\rm \mbox{\normalsize noun phrase}
 \left\{ \begin{array}{l}
     \rm \mbox{\normalsize {\bf generic} noun phrase}\\
         \mbox{\normalsize {\bf non generic} noun phrase}
            \left\{ \begin{array}{l}
              \rm \mbox{\normalsize {\bf definite} noun phrase}\\
                  \mbox{\normalsize {\bf indefinite} noun phrase}
            \end{array}
            \right.
\end{array}
\right.
\] 
}

\paragraph{Generic noun phrase}
A noun phrase is classified as generic 
when it denotes all members of the class of the noun phrase 
or the class itself of the noun phrase. 
For example, ``INU(dog)'' in the following sentence is a generic noun phrase.
\begin{equation}
  \begin{minipage}[h]{11.5cm}
    \begin{tabular}[t]{ll}
  \underline{INU-WA} & YAKUNI-TACHIMASU.\\
  (dog) & (useful)\\
\multicolumn{2}{l}{
  (\underline{Dogs} are useful.)}
    \end{tabular}
  \end{minipage}
  \label{eqn:3c_doguse}
\end{equation}
A generic noun phrase cannot refer to an indefinite/definite noun phrase. 
Two generic noun phrases can refer to each other. 
%
\paragraph{Definite noun phrase}
A noun phrase is classified as definite 
when it denotes a contextually non-ambiguous member 
of the class of the noun phrase. 
For example, ``INU(dog)'' 
in the following sentence is a definite noun phrase.
\begin{equation}
  \begin{minipage}[h]{11.5cm}
    \begin{tabular}[t]{lll}
  \underline{INU-WA} & MUKOUHE & IKIMASHITA.\\
  (dog) & (away) & (go)\\
\multicolumn{3}{l}{
  (\underline{The dog} went away.)}
\end{tabular}
\end{minipage}
\label{eqn:3c_thedoguse}
\end{equation}
A definite noun phrase can refer to a noun phrase 
that has already appeared. 
\paragraph{Indefinite noun phrase}
An indefinite noun phrase denotes 
an arbitrary member of the class of the noun phrase.
For example, the following ``INU(dog)'' is an indefinite noun phrase.
\begin{equation}
  \begin{minipage}[h]{11.5cm}
    \begin{tabular}[t]{lll}
    \underline{INU-GA} & SANBIKI & IMASU.\\    
    (dog) & (three) & (there is)\\
\multicolumn{3}{l}{
  (There are three \underline{dogs}.)}
\end{tabular}
\end{minipage}
  \label{eqn:3_threedogs}
\end{equation}
An indefinite noun phrase cannot refer to a noun phrase 
that has already appeared. 

\section{How to Estimate Referent of Noun Phrase}
\label{sec:how_to}

To determine referents of noun phrases, 
we made the following three constraints. 
\begin{enumerate}
\item 
Referential property constraint 
\item 
Modifier constraint 
\item 
Possessor constraint 
\end{enumerate}
When two noun phrases 
which have the same head noun 
satisfy these three constraints, 
the system judges that the two noun phrases refer to each other. 
These three constraints are as follows:  

\subsection{Referential Property Constraint}
\label{sec:ref_pro}

First, our system estimates the referential property of a noun phrase
using the method in Chapter \ref{chap:ref}. 
The method estimates a referential property 
using surface expressions in the sentences. 
For example, 
since the second ``OJIISAN (old man)'' 
in the following sentences 
is accompanied by a particle ``WA (topic)'', 
and the predicate is in the past tense, 
it is estimated to be a definite noun phrase. 
\begin{equation}
  \begin{minipage}[h]{11.5cm}
    \begin{tabular}[t]{lll}
\underline{OJIISAN}-WA & JIMEN-NI &
KOSHI-WO-OROSHIMASHITA.\\
(old man) & (ground) & (sit down)\\ 
\multicolumn{3}{l}{
(\underline{The old man} sat down on the ground.)}\\[0.3cm]
\end{tabular}
\begin{tabular}[t]{lll}
YAGATE & \underline{OJIISAN}-WA & 
NEMUTTE-SHIMAIMASHITA.\\
(soon) & (old man) & (fall asleep)\\
\multicolumn{3}{l}{
(\underline{He} soon fell asleep.)}
\end{tabular}
  \end{minipage}
\label{eqn:ojiisan_jimen_meishi_2}
\end{equation}

Next, our system
determines the referent of a noun phrase 
using its estimated referential property. 
When a noun phrase is estimated to be a definite noun phrase, 
our system judges that the noun phrase refers to a previous noun phrase 
which has the same head noun. 
For example, 
the second ``OJIISAN'' in the above sentences 
is estimated to be a definite noun phrase, 
and our system judges that it refers to the first ``OJIISAN''. 

When a noun phrase is not estimated to be a definite noun phrase, 
the noun phrase can refer to a noun phrase that has already been mentioned, 
because estimating the referential property may fail. 
Therefore, 
when a noun phrase is not estimated to be a definite noun phrase, 
our system gets a possible referent of the noun phrase from topic and focus, 
and determines the referent of the noun phrase 
using the following three kinds of information. 
\begin{itemize}
\item
the plausibility 
of the estimated referential property
that is a definite noun phrase

\item
the weight 
of a possible referent in the case of topic or focus

\item
the distance 
between the estimated noun phrase and 
a possible referent
\end{itemize}

\subsection{Modifier Constraint}

It is insufficient to determine referents of noun phrases 
by only using referential property. 
When two noun phrases have different modifiers, 
they commonly do not refer to each other. 
For example, 
``HIDARI(left)-NO HOO(cheek)'' in the following sentences 
do not refer to ``MIGI(right)-NO HOO(cheek)''. 
\begin{equation}
  \begin{minipage}[h]{11.5cm}
\small
    \begin{tabular}[t]{llll@{ }ll}
KONO & OJIISAN-NO & KOBU-WA & 
\underline{MIGI-NO} & \underline{HOO}-NI & ARIMASHITA.\\
(this) & (old man) & (lump) & (right)  & (cheek) & (be on)\\
\multicolumn{6}{l}{
(This old man's lump was on \underline{his right cheek}.)}\\[0.3cm]
\end{tabular}
\begin{tabular}[t]{l@{ }l@{ }l@{ }l@{ }l}
TENGU-WA, & KOBU-WO &
\underline{HIDARI-NO} & \underline{HOO}-NI & 
TSUKETE-SHIMAIMASHITA.\\
(tengu)\footnotemark & (lump) & (left) & (cheek) & (put on)\\
\multicolumn{5}{l}{
(The "tengu" put a lump on \underline{his left cheek})}
\end{tabular}
  \end{minipage}
\label{eqn:mouhitori_ojiisan_hoho_kobu}
\end{equation}
\footnotetext{
A tengu is a kind of monster.} 
Therefore, we made the following constraint: 
When a noun phrase has a modifier, it cannot refer to 
a noun phrase that does not have the same modifier. 
When a noun phrase does not have a modifier, 
it can refer to a noun phrase that has any modifier. 

\subsection{Possessor Constraint}

When a noun phrase has a semantic marker {\sf PAR} (a part of a body)
\footnote{
In this thesis, we use Noun Semantic Marker Dictionary
\cite{imiso-in-BGH} as a semantic marker dictionary. 
}, 
our system tries to estimate 
the possessor of the object denoted by the noun phrase. 
We suppose that 
the possessor of a noun phrase is the subject or 
the noun phrase's nearest topic 
that has a semantic marker {\sf HUM} (human) or 
a semantic marker {\sf ANI} (animal). 
For example, 
the possessor of the first ``HOO (cheek)'' in the following sentences 
is estimated to be ``OJIISAN (old man)'' 
because 
``OJIISAN (old man)'' is followed by a particle ``NIWA'', 
is the topic in the sentence, 
and has a semantic marker {\sf HUM} (human). 

\vspace{0.3cm}

{\small
    \begin{tabular}[t]{l@{ }l@{}l@{ }l@{ }l@{ }l}
      OJIISAN-NIWA & \underline{[OJIISAN-NO]\footnotemark
} &
    \underline{HIDARI-NO} & \underline{HOO-NI} & KOBU-GA & ARIMASHITA.\\
    (old man) & (old man's) & (left) & (cheek) & (lump) & (be on)\\
\multicolumn{6}{l}{
  (This old man had a lump on \underline{his left cheek}.)} \\[0.3cm]
\end{tabular}
\footnotetext{
\label{foot:bracket}
The words in brackets [ ] are omitted in the sentences.} 

\begin{tabular}[t]{llll}
    SORE-WA & HITO-NO & KOBUSHI-HODOMO-ARU & KOBU-DESHITA.\\
    (it) & (person) & (fist) & (lump)\\
\multicolumn{4}{l}{
  (It is about the size of a person's fist.)}\\[0.3cm]
\end{tabular}

\begin{tabular}[t]{l@{ }l@{ }l@{ }l@{ }l}
  \underline{[OJIISAN-NO]}
 & \underline{HOO-WO} &
 HUKURAMASETE- IRUYOUNI &
    MIERUNODESHITA.\\
    (old man) & (cheek) & (puff) & (look)\\
    \multicolumn{5}{l}{
      (He looked as if he had puffed out \underline{his cheek}.)}
  \end{tabular}\\
}

\vspace{0.3cm}

\noindent
The possessor of the second ``HOO (cheek)'' 
is also estimated to be ``OJIISAN (old man)'' 
because 
``OJIISAN (old man)'' is the subject in the sentence
\footnote{
  Omitted subjects are estimated 
  by the method in Chapter \ref{chap:deno}. 
}. 

We made the following constraint by using possessors. 
When the possessor of a noun phrase is estimated, 
the noun phrase cannot refer to 
a noun phrase that does not have the same possessor. 
When the possessor of a noun phrase is not estimated, 
the noun phrase can refer to 
a noun phrase that has any possessor. 

For example, 
since the two ``HOO (cheek)'' in the above sentences 
have the same possessor ``OJIISAN (old man)'', 
our system correctly judges that the two ``HOO (cheek)'' 
have the same referent. 

\section{Anaphora Resolution System}

\subsection{Procedure}
\label{wakugumi}

Before determining referents, 
sentences are transformed into a case structure 
by the case structure analyzer\cite{csan2_ieice}. 

Referents of noun phrases are 
determined by heuristic rules which are 
made from such information as the three constraints 
mentioned in Section~\ref{sec:how_to}. 
Using these rules, our system takes possible referents 
and gives them points. It judges that 
the candidate having the maximum total score is the referent. 
This is because a number of types of information is combined 
in anaphora resolution. 
We can specify 
which rule takes priority 
by using points. 

The heuristic rules 
are given in the following form. 

\begin{center}
    \begin{minipage}[c]{8cm}
      \hspace*{0.7cm}{\sl Condition} $\Rightarrow$ \{ Proposal Proposal .. \}\\
      \hspace*{0.7cm}Proposal := ( {\sl Possible-Referent} \, {\sl Point} )
    \end{minipage}
\end{center}

\noindent
In {\sl Condition}, 
surface expressions, semantic constraints, 
referential properties, etc. are written as conditions. 
In {\sl Possible-Referent}, 
a possible referent, 
``indefinite'', 
or other things are written. 
``indefinite'' means that 
the noun phase is an indefinite noun phrase, 
and it does not refer to a previous noun phrase. 
{\sl Point} means the plausibility value of the possible referent. 

\subsection{Heuristic Rule for Estimating Referents}
\label{sec:3c_rule}

We made 8 heuristic rules for noun phrase anaphora resolution. 
All the rules are given below. 

\begin{enumerate}
\item[R1] 
  When a noun phrase is like ``IKA (the following)'', \\
  \{(Next sentences, \,$50$)\}
  \footnote{(a,b) means {\sl candidate}(a) and {\sl point}(b). }


\item[R2] 
  When a noun phrase is modified by the words ``SOREZORE-NO (each)'' and ``ONOONO-NO (each)'', \\
  \{(Indefinite, \,$25$)\}

\item[R3] 
  When a noun phrase is the word ``JIBUN (oneself)'', \\
  \{(The subject in the sentence, \,$25$)\}

\item[R4]  
When a noun phrase is estimated to be a definite noun phrase, 
and satisfies modifier constraint and possessor constraint, 
and the same noun phrase X has already appeared, 
  \\ \{(The noun phrase X, \,$30$)\}

\item[R5]  
  When a noun phrase is estimated to be a generic noun phrase, \\
  \{(Generic, \,$10$)\} 

\item[R6] 
  When a noun phrase is estimated to be an indefinite noun phrase, \\
  \{(Indefinite, \,$10$)\}

\item[R7]  
  When a noun phrase is like 
  ``ISSHO (together)'' and ``HONTOU (true)'', 
  which is used as an adverb or an adjective, 
  
  \{(No referent, \,$30$)\}

    (Ex.)
{\small
\begin{tabular}[t]{l@{ }l@{ }l@{ }l}
    TENGU-TACHI-WA & \underline{ISSHO(together)}-NI &
    WARAI & DASHIMASHITA. \\
    (tengu) & (together) & (laugh) & (begin)\\
    \multicolumn{3}{l}{
      (The tengu began laughing \underline{together}. )
    }
\end{tabular}\\
}

\item[R8]  
  When a noun phrase X is not estimated to be a definite noun phrase, \\
  \{
  (A noun phrase X which satisfies 
  modifier constraint and possessor constraint, \, $W-D+P+4$)\}\\ 

\begin{table}[t]
  \caption{The weight in the case of topic}
  \label{tab:topic}
  \begin{center}
\small
    \leavevmode
\begin{tabular}[c]{|l|l|r|}\hline
\multicolumn{1}{|c|}{Surface Expression}  & \multicolumn{1}{|c|}{Example}   & Weight \\\hline
Pronoun/Zero-Pronoun GA/WA&  (\underline{John}GA (subject))SHITA (done). &21 \\\hline
Noun WA/NIWA        &  \underline{John}WA (subject)SHITA (do).   &20 \\\hline
\end{tabular}
  \end{center}
\end{table}

\begin{table}[t]
  \caption{The weight in the case of focus}
  \label{tab:focus}
  \begin{center}
\small
    \leavevmode
\begin{tabular}[c]{|p{5cm}|l|r|}\hline
\multicolumn{1}{|p{5cm}|}{Surface Expression }  & \multicolumn{1}{|c|}{Example}   & Weight \\\hline
Pronoun/Zero-Pronoun WO(object)/NI(to) /KARA(from) & (\underline{John}NI (to))SHITA (done). & 16 \\\hline
Noun GA (subject)/MO/DA/NARA & \underline{John}GA (subject)SHITA (do).   & 15 \\\hline
Noun WO (object)/NI/, /.       & \underline{John}NI (object)SHITA (do).   & 14 \\\hline
Noun HE (to)/DE (in)/KARA (from)   & \underline{GAKKOU (school)}HE (to)IKU (go).   & 13 \\\hline
\end{tabular}
  \end{center}
\end{table}

\begin{table}[t]
  \caption{The plausibility(P) that the referential property is definite}
  \label{tab:teimeishidenai_doai}
  \begin{center}
    \leavevmode
\small
    \begin{tabular}[h]{|l|r|r|r|r|}\hline
      Difference score between definite and other referential property  &  0  & 1  &  2 & 3 $\sim$ \\\hline
      The plausibility $P$                                &  0  & $-3$ & $-6$ & $-\infty$\\\hline
    \end{tabular}
  \end{center}
\end{table}

The values $W$, $D$, $P$ are defined as follows: 
The definition and the weight ($W$) of topic and focus are given 
in Table~\ref{tab:topic} and Table~\ref{tab:focus} respectively. 
 (In this work, a topic is defined as 
a theme which is described, and 
a focus is defined as 
a word which is stressed by the speaker (or the writer). 
But we cannot detect topics and foci correctly. 
Therefore we approximated them 
by Table~\ref{tab:topic} and Table~\ref{tab:focus}.) 
When a possible referent is a topic, 
the distance ($D$) between the estimated noun phrase and 
the possible referent is the number of topics between them. 
When a possible referent is a focus, 
the distance ($D$) is the number of foci between them. 
The plausibility ($P$) that the referential property is a definite
is given in Table~\ref{tab:teimeishidenai_doai}. 
In the table 
``Difference score between definite and other referential property'' 
is determined as follows. 
When the method in Chapter \ref{chap:ref} estimates a referential property, 
it gives each category of referential property some points, 
and it outputs the score of each category. 
From these scores our system calculates 
``Difference score between definite and other referential property''. 
These values were determined by hand 
on training sentences mentioned in Section~\ref{sec:eval}.

\end{enumerate}

\subsection{Example of Estimating the Referent of a Noun Phrase}

An example of determining the referent of a noun phrase is 
shown in Figure~\ref{tab:dousarei}. 
This figure shows that 
the underlined ``HI (fire)'' in the figure was interpreted properly. 
The process is as follows:  

\begin{figure}[t]
  \begin{center}
\fbox{
\begin{minipage}[h]{13cm}
\small
\begin{tabular}[t]{l@{ }l@{ }l@{ }l@{ }l}
OJIISAN-WA &
AKICHI-NI & HI-GA & 
MOETEIRU-NONI & KIGA-TSUKIMASHITA.\\
(old man) & (open space) & (fire) & (burn) & (notice)\\
    \multicolumn{5}{l}{
      (The old man noticed that there was a big bright fire burning 
      in an open space.)}\\[0.5cm]
\end{tabular}

\begin{tabular}[t]{lllll}
AKAI & KAO-WO-SHITA & 
OTOKO-TACHI-GA, & \underline{HI}-NO & MAWARI-NI\\
(red) & (face) & (man) & (fire) & (around)\\
\end{tabular}

\begin{tabular}[t]{ll}
TATTEIRU-NOWO & MIMASHITA.\\
(stand) & (see)\\
\multicolumn{2}{l}{
(He saw some men with red faces 
standing around the \underline{fire}.)}
\end{tabular}

\vspace{0.5cm}

\begin{tabular}[h]{|l|r|r|r|}\hline
Satisfied Rule         & \multicolumn{2}{|c|}{Score}\\\hline
                       & Generic    & \multicolumn{1}{|l|}{``HI (fire)''}\\
                       &            & in the previous sentence \\\hline
Rule 5                 &   10       &                   \\
Rule 8                 &            &   12             \\\hline
Total Score            &   10       &   12             \\\hline
\end{tabular}

\vspace{0.5cm}

Estimate of referential property 

\vspace{0.2cm}

\begin{tabular}[h]{|l|r|r|r|}\hline
Referential property    &  Indefinite & Definite & Generic \\\hline
Point                   &  1          &  2       &  3       \\\hline
\end{tabular}

\vspace{0.5cm}

``HI (fire)'' in the previous sentence has the following score. 

\vspace{0.2cm}

\hspace*{1cm}
$ W - D + P = 15 - 4 - 3 + 4 = 12 $ 

\caption{Example of estimating the referent of a noun phrase}
\label{tab:dousarei}
\end{minipage}
}
\end{center}
\end{figure}

At first, our system estimated the referential property
 of the underlined ``HI (fire)''. 
The referential property was incorrectly estimated to be 
a generic noun phrase 
as shown in the table ``Estimate of referential property'' 
in the figure. 
Since the estimated referential property was a generic noun phrase, 
the rule R5 proposed a possible referent ``Generic'', 
and gave it 10 points.  
Also, the rule R8, 
which applies when 
the estimated referential property is incorrect, 
proposed a possible referent ``HI (fire)'' in the previous sentence. 
Since it does not have a modifier and a possessor, 
it satisfied modifier constraint and possessor constraint. 
It was given a value of the evaluation function $W-D+P$ in 
referential property constraint. 
The weight $W$ was given 15 by Table~\ref{tab:focus} 
because it is followed by a particle ``GA (subject)''. 
The distance $D$ was given 4 
because there are four foci ``OTOKO (man)'', ``KAO (face)'', 
``KI (notice)'' and  \verb+<+``HI (fire) in the previous sentence\verb+>+ 
between the underlined ``HI (fire)'' and 
\verb+<+``HI (fire) in the previous sentence\verb+>+. 
Since 
the difference score between definite and other referential property 
was 1 ($= 3 (\mbox{generic}) - 2 (\mbox{indefinite})$), 
the plausibility ($P$) was given $-3$ 
by Table~\ref{tab:teimeishidenai_doai}. 
Therefore, the evaluation function $W - D + P + 4$ is 12 ($= 15 - 4 - 3 + 4$). 
``HI (fire)'' in the previous sentence was 12. 
Since the value 12 of ``HI (fire)'' was higher than 
the value 10 of ``Generic'', 
our system judged that the underlined ``HI (fire)'' refers to 
the ``HI (fire)'' in the previous sentence correctly. 
As the result, the referential property of the underlined ``HI (fire)'' was 
judged to be a definite noun phrase correctly. 

\section{Experiment and Discussion}

\subsection{Experiment}
\label{sec:eval}

\begin{table}[t]
  \fbox{
  \begin{minipage}[h]{13cm}
  \begin{center}
  \caption{Result}
  \label{tab:noun_result}

\vspace{0.3cm}

    \leavevmode
    \newcommand{\mn}[1]{\begin{minipage}[t]{3cm}%
      \baselineskip0.8cm%
      #1 \smallskip \end{minipage}}
\begin{tabular}[c]{|l|r@{}c|r@{}c|}\hline
 &  \multicolumn{2}{|p{2.8cm}|}{{Recall}}& \multicolumn{2}{|p{2.8cm}|}{{Precision}}\\\hline
Training sentences      &  82\% & (130/159) & 85\% & (130/153) \\
Test sentences &  79\% & ( 89/113) &  77\% & ( 89/115)\\\hline
\end{tabular}
  \end{center}
Training sentences \{example sentences (43 sentences), a fork tale ``KOBUTORI JIISAN''\cite{kobu} (93 sentences), an essay in ``TENSEIJINGO'' (26 sentences), an editorial (26 sentences), an article in ``Scientific American (in Japanese)''(16 sentences)\}\\
Test sentences \{a fork tale ``TSURU NO ONGAESHI''\cite{kobu} (91 sentences), two essays in ``TENSEIJINGO'' (50 sentences), an editorial (30 sentences), articles in ``Scientific American(in Japanese)'' (13 sentences)\}
  \end{minipage}
  }
\end{table}

\begin{table}[t]
  \fbox{
  \begin{minipage}[h]{13cm}
  \begin{center}
  \caption{Comparison}
  \label{tab:sijisei_taishou}

\vspace{0.3cm}

    \leavevmode
    \newcommand{\mn}[1]{\begin{minipage}[t]{3cm}%
      \baselineskip0.8cm%
      #1 \smallskip \end{minipage}}
\begin{tabular}[c]{|r@{}c@{}|r@{}c@{}|r@{}c@{}|r@{}c@{}|r@{}c@{}|}\hline
 \multicolumn{2}{|c|}{Method 1} & 
 \multicolumn{2}{c|}{Method 2} & 
 \multicolumn{2}{c|}{Method 3} & 
 \multicolumn{2}{c|}{Method 4} & 
 \multicolumn{2}{c|}{Method 5}\\\hline 
\multicolumn{10}{|l|}{Training sentences}\\\hline
  92\% & (117/127) &  82\% & (130/159) &  72\% & (123/170) &  65\% & (138/213) & 52\% & (134/260) \\
  76\% & (117/153) &  85\% & (130/153) &  80\% & (123/153) &  90\% & (138/153) & 88\% & (134/153) \\\hline
\multicolumn{10}{|l|}{Test sentences}\\\hline
  92\% & ( 78/ 85) &  79\% & ( 89/113) &  69\% & ( 79/114) &  58\% & ( 92/159) & 47\% & (102/218) \\
  68\% & ( 78/115) &  77\% & ( 89/115) &  69\% & ( 79/115) &  80\% & ( 92/115) & 89\% & (102/115) \\\hline
\end{tabular}
  \end{center}
{\small

Method 1 : Only when it is estimated to be definite can it refer to another noun phrase

Method 2 : The method of this work

Method 3 : No use of referential property

Method 4 : No use of modifier constraint and possessor constraint

Method 5 : The same two nouns co-refer

}
\end{minipage}
}
\end{table}

Before estimating the referents of noun phrases, 
sentences were at first 
transformed into a case structure 
by the case structure analyzer\cite{csan2_ieice}. 
The errors made by the case analyzer were corrected by hand. 
We show the result of estimating the referents of noun phrases 
in Table \ref{tab:noun_result}. 

To verify that the three constraints
(referential proper, modifier, and possessor constraint) are effective, 
we experimented with the changed condition 
and compared them. 
The results are shown in Table~\ref{tab:sijisei_taishou}. 
The upper row and the lower row of this table 
show precision and recall respectively. 
{\it Precision}\, is 
the fraction  of noun phrases which were judged to have the antecedents. 
{\it Recall}\, is the fraction of noun phrases 
 which have the antecedents. 

In these experiments we used training sentences and test sentences. 
The training sentences were used to 
make the heuristic rules in Section~\ref{sec:3c_rule} by hand. 
The test sentences 
were used to verify the effectiveness of these rules. 

In Table \ref{tab:sijisei_taishou}, 
Method~1 ``Only when it is estimated 
to be definite can it refer to another noun phrase'' 
is a case when a noun phrase can refer to a noun phrase, 
only when the estimated referential property 
is a definite noun phrase, 
where modifier constraint and possessor constraint are used. 
Method~2 ``The method of this work'' is 
the method mentioned
in Section~\ref{sec:how_to}, 
which uses all three constraints.  
Method~3 ``No use of referential property'' 
is a method without referential property, 
which uses only such information as distance, topic-focus, 
modifier, and possessor. 
Method~4 ``No use of modifier constraint and possessor constraint'' 
is a method 
without modifier constraint and possessor constraint. 
Method~5 ``The same two nouns co-refer'' is a case 
that a noun phrase always refers to a noun phrase 
that has the same head noun. 

The table shows many results. 
In Method~2 ``The method of this work'', 
both the recall and the precision were high. 
This indicates that 
the referential property was used properly in the method that 
is described in this chapter. 
Method~2 ``The method of this work'' 
was higher than Method~3 ``No use of referential property''
 in both recall and precision. 
This indicates that the information of referential property is necessary. 
In Method~1 ``Only when it is estimated 
to be definite can it refer to another noun phrase'', 
the recall was low. 
The reason is because 
there were many noun phrases that 
are definite 
but were estimated to be indefinite/generic, and 
the system estimated that 
the noun phrases cannot refer to noun phrases. 
In Method~4 ``No use of modifier constraint and possessor constraint'', 
the precision was low. 
Since modifier constraint and possessor constraint were not used, and 
there were many pairs of two noun phrases 
that do not co-refer, such as ``HIDARI(left)-NO HOO(cheek)'' 
and ``MIGI(right)-NO HOO(cheek)'', 
these pairs were incorrectly interpreted as co-reference. 
This indicates that 
it is necessary to use modifier constraint and possessor constraint. 
In Method~5 ``The same two nouns co-refer'', 
the precision was lower than in Method~4. 
This is because 
referential properties were not used 
and the system judged that 
a noun phrase which is not a definite noun phrase 
refers to another noun phrase. 

\subsection{Examples of Errors}

We found that it was necessary to use modifiers and possessors through 
the above experiments. 
But 
since the possessor of a noun was estimated incorrectly, 
the referent was also estimated incorrectly as follows. 

%
\begin{quote}
\small
\begin{tabular}[t]{l@{ }l@{ }l@{ }l@{ }l@{ }l@{ }l@{ }l@{ }l}
OJIISAN-WA &
\underline{\underline{(OJIISAN-NO)}} & \underline{\underline{\.S\.E\.N\.A\.K\.A-KARA}} &
SHIBA-WO & {\footnotesize OROSHIMASHITA.}\\
(old man) & (old man's) & (back) & 
(firewood) & (take down)\\
\multicolumn{9}{l}{
(He took down the bundle of firewood from {\underline{\underline{his back}}}.)}\\[0.3cm]
\end{tabular}

\hspace*{0.5cm}
(an omission of a middle part)\\[0.5cm]
\begin{tabular}[t]{l@{ }l@{ }l@{ }l}
OJIISAN-WA & OTOKOTACHI-WO &
NINGEN-DATO & OMOTTEIMASHITAGA, \\
(old man) & (man) & 
(human beings) & (think)\\
\multicolumn{4}{l}{
(The old man thought they were human beings, 
)}\\[0.3cm]
\end{tabular}

\begin{tabular}[t]{l@{ }l@{ }l}
MAMONAKU & TENGU-DEARU-KOTO-GA & WAKARIMASHITA.\\
(soon) & (tengu) &(realize)\\
\multicolumn{3}{l}{
(but soon he realized that they were ``tengu,'' or supernatural beings.)}\\[0.3cm]
\end{tabular}

\begin{tabular}[t]{l@{ }l@{ }l@{ }l@{ }l}
\underline{[TENGU-NO]} & \underline{SENAKA-NIWA} & OOKINA & TSUBASA-GA &
ARUNODESU.\\
(tengu) & (back) & (large) & (wing) & (have on)\\
\multicolumn{5}{l}{
(They had large wings on \underline{their backs}.)}\\[0.3cm]
\end{tabular}
\end{quote}
%
Since the underlined ``SENAKA (back)'' in this example is a part of an animal, 
the possessor is estimated. 
Although the proper possessor is ``TENGU (tengu)'', 
the system estimated incorrectly that 
the possessor was ``OJJISAN (old man)'' 
that is a topic of the previous sentence. 
For this reason, 
our system judged that 
this ``SENAKA (back)'' refers to 
the twice underlined 
``[OJJISAN-NO] \.S\.E\.N\.A\.K\.A (the old man's back)'' incorrectly. 

Sometimes a noun can refer to a noun that has a different modifier. 
In such a case, the system made an incorrect judgment. 
\begin{quote}
\small
\begin{tabular}[t]{l@{ }l@{ }l@{ }l@{ }l@{ }l@{ }l}
OJIISAN-WA & \underline{CHIKAKU-NO} &
\underline{OOKINA} &  \underline{SUGI-NO} & \underline{KI-NO} & \underline{NEMOTO-NI}&
\underline{ARU}\\
(old man) & (near) &
(huge) & (cedar) & (tree) & (base) &
(be at)\\
\end{tabular}
\begin{tabular}[t]{l@{ }l@{ }l}
\underline{ANA}-DE & 
AMAYADORI-WO &
SURU-KOTO-NI-SHIMASHITA.\\
(hole) & 
(take shelter from the rain) & 
(decide to do)\\
\multicolumn{3}{l}{
(So, he decided to take shelter from the rain in \underline{a hole 
which is at the base of}}\\
\multicolumn{3}{l}{
\underline{a huge cedar tree nearby}.)}\\[0.3cm]
\end{tabular}

\vspace{0.3cm}

(an omission of a middle part)

\vspace{0.3cm}

\begin{tabular}[t]{l@{ }l@{ }l@{ }l@{ }l}
TSUGI-NO-HI, & KONO & OJIISAN-WA &
YAMA-HE & ITTE,\\
(next day) & (this) & (old man)& 
(mountain) & (go to)\\
\multicolumn{5}{l}{
(The next day, this man went to the mountain, 
)}\\
\end{tabular}

\vspace{0.3cm}

\begin{tabular}[t]{l@{ }l@{ }l@{ }l@{ }l}
\underline{SUGI-NO} & \underline{KI-NO} & \underline{NEMOTO-NO} & \underline{ANA}-WO & 
MITSUKEMASHITA.\\
(cedar) & (tree) & (at base) & (hole) & 
(found)\\
\multicolumn{5}{l}{
(and found \underline{the hole at the base of the cedar tree}.)}\\
\end{tabular}
\end{quote}
%
%
%
The two ``ANA (hole)'' in this sentence refer to each other. 
But our system judged that 
the two ``ANA (hole)'' in these sentences do not refer to each other 
because the modifiers of the two ``ANA (hole)'' are different. 
In order to correctly analyze this case, 
it is necessary to decide 
whether two different expressions are equal in meaning. 

\section{Summary}

This chapter described the method of how to estimate the referents of 
noun phrases using the referential properties, 
the modifiers, and the possessors. 
As a result of using this method, 
we obtained a precision rate of 82\% 
and a recall rate of 85\% in the estimation 
of referents of noun phrases that have antecedents 
on training sentences, 
and obtained a precision rate of 79\% 
and a recall rate of 77\% 
on test sentences. 
We verified that 
it is effective to use 
referential properties, modifiers, and possessors of noun phrases.

\mychapter
{\chapter{Indirect Anaphora Resolution in Noun Phrases}}
{\chapter[Indirect Anaphora Resolution]
{Indirect Anaphora Resolution in Noun Phrases}}
\label{chap:indian}

\setcounter{topnumber}{5}

\section{Introduction}
\label{sec:4c_intro}

Chapter \ref{chap:noun} described 
the case when 
a noun phrase refers to an entity that has already been mentioned. 
Chapter \ref{chap:indian} describes 
the case when 
a noun phrase refers to an entity 
that has not been mentioned yet, 
but an entity associated with an entity 
that has already been mentioned. 
For example, ``{\it I went into an old house last night. 
{The roof} was leaking badly and ...}'' 
indicates that ``{\it The roof}'' is associated with ``{\it an old house}'',
which has already been mentioned. This kind of reference 
 (indirect anaphora) has not been thoroughly 
studied in natural language processing\footnote{
\cite{Tanaka1} made the investigation of 
resolving indirect anaphora in some nouns such as ``TAISEKI (volume)'' 
in sentences on chemistry. But there is no research  
resolving indirect anaphora in all the nouns. 
},
but is important for 
coherence resolution, language understanding, and machine translation. 
We propose a method to resolve indirect anaphora in Japanese nouns 
using the relationships between two nouns. 

When we analyze indirect anaphora, 
we need a case frame dictionary for nouns 
containing an information about relations between two nouns. 
For example, in the case of the above example, 
the knowledge that 
``roof'' is a part of ``house'' is required 
to analyze the indirect anaphora. 
But no such noun case frame dictionary exists at present. 
We considered whether we can use the example-based method 
to solve this problem. 
In this case, 
the knowledge that ``roof'' is a part of ``house'' is 
analogous to ``house of roof''. 
Therefore 
we use examples of the form ``X of Y'' instead. 
In the above example, 
we use a linguistic data such as ``the roof of a house''. 
In the case of verbal nouns, we do not use 
``X of Y'' but a verb case frame dictionary. 
This is because 
a noun case frame is similar to 
a verb case frame 
and a verb case frame dictionary exists at present. 

The next section describes 
a method of resolving indirect anaphora. 

\section{How to Resolve Indirect Anaphora}

An anaphor and the antecedent in an indirect anaphora 
have a certain relation. 
For example, 
``YANE (roof)'' and ``HURUI IE (old house)'' 
are in an indirect anaphoric relation 
which is a part-of relation. 
\begin{equation}
  \begin{minipage}[h]{11.5cm}
    \begin{tabular}[t]{lllll}
   SAKUBAN & ARU  & HURUI & IE-NI & ITTA.\\
  (last night) & (a certain) & (old) & (house) & (go)\\
\multicolumn{5}{l}{
  (I went into an old house last night.)}
    \end{tabular}

\vspace{0.3cm}

  \begin{tabular}[t]{lll}
    \underline{YANE-WA} & HIDOI  & AMAMORIDE ... \\
  (roof) & (badly) & (be leaking)\\
\multicolumn{3}{l}{
  (\underline{The roof} was leaking badly and ... )}
    \end{tabular}
  \end{minipage}
  \label{eqn:ie_sakuban}
\end{equation}
When we analyze the indirect anaphora, 
we need a dictionary containing 
information about relations between anaphors and their antecedents. 

\begin{table}[t]
\small
  \leavevmode
  \begin{center}
    \caption{Example of noun case frame dictionary}
    \label{tab:noun_case_frame}
\begin{tabular}{|l@{ }|l@{ }|l@{ }|}\hline
Anaphor      & Things which can be the Antecedent& Relation\\\hline
KAZOKU (family)  & HITO (human)           & belong\\\hline
KOKUMIN (nation) & KUNI (country)         & belong\\\hline
GENSHU (the head of state)   & KUNI (country) & belong\\\hline
YANE (roof)      & TATEMONO (building)    & part of\\\hline
MOKEI (model)    & SEISANBUTSU (product) & object\\
                & (ex. HIKOUKI (air plain), HUNE (ship)) & \\\hline
GYOUJI (event)   & SOSHIKI (organization) & agent\\\hline
JINKAKU (personality)  & HITO (human)    & possessive\\\hline
KYOUIKU (education)    & HITO (human)    & agent\\
             & HITO (human)             & recipient\\
             & NOURYOKU (ability)     & object\\
             & (ex. SUUGAKU (mathematics))  & \\\hline
KENKYUU (research) & HITO (human), SOSHIKI (organization) & agent\\
             & GAKUMON BUN'YA (field of study)          & object\\\hline
\end{tabular}
\end{center}
\end{table}

We show examples of the relations between an anaphor and the antecedent 
in Table \ref{tab:noun_case_frame}. 
The form of Table \ref{tab:noun_case_frame} is 
similar to the form of a verb case frame dictionary. 
We call a dictionary containing 
the relations between two nouns {\sl a noun case frame dictionary}. 
But no noun case frame dictionary has been created so far. 
Therefore, we substitute it by examples of ``X NO Y (Y of X)'' 
and by a verb case frame dictionary. 
``X NO Y'' is a Japanese expression. 
It means ``Y of X'', ``Y in X'', ``Y for X'', etc. 

\begin{table}[t]
  \caption{Case frame of verb ``KUICHIGAU (differ)''}
  \label{tab:kuitigau_frame}
  \begin{center}
    \leavevmode
\begin{tabular}[h]{|l|l|l|}\hline
Surface Case      & Semantic Marker & Examples\\\hline
Ga-case (subject)  & abstract         & DEETA (data), IKEN (opinion)\\
To-case (object)   & abstract         & DEETA (data), MIKATA (viewpoint)\\\hline
\end{tabular}\\
  \end{center}
\end{table}

Resolution of indirect anaphora is done by the following steps. 
\begin{enumerate}
\item 
\label{enum:youso_kenshutu}
We detect some elements which will be analyzed 
in indirect anaphora resolution 
using ``X NO Y'' 
and a verb case frame dictionary. 
When a noun is a verbal noun, 
we use a verb case frame dictionary. 
Otherwise, 
we use examples: ``X NO Y''. 
For example, 
``KUICHIGAI (difference)'' is a verbal noun, and 
we use a case frame of a verb ``KUICHIGAU (differ)'' 
for the indirect anaphora resolution of ``KUICHIGAI (difference).'' 
The case frame is shown in Table \ref{tab:kuitigau_frame}. 
In this table there are two case components, 
GA-case (subject) and TO-case (object). 
These two case components are 
elements which will be analyzed in indirect anaphora resolution. 
%
\begin{equation}
  \begin{minipage}[h]{11.5cm}
    \small
    \begin{tabular}[t]{llll}
  Tom-WA & DEETA-WO & KONPYUUTA-NI & UCHIKONDE-IMASHITA.\\
  (Tom)  & (data)   & (computer)   & (store)\\
\multicolumn{4}{l}{
  (Tom was storing the data in a computer.)}
    \end{tabular}

\vspace{0.3cm}

    \begin{tabular}[t]{lll}
  YATTO & HANBUN & YARIOEMASHITA.\\
  (Finally)  & (half) & (finish)\\
\multicolumn{3}{l}{
  (Finally he was half finished. )}
    \end{tabular}

\vspace{0.3cm}

    \begin{tabular}[t]{llll}
  John-GA & HURUI & DEETA-WO & MISEMASHITA.\\
  (John)  & (old) & (data)   & (show)\\
\multicolumn{4}{l}{
  (John showed him some old data.)}
    \end{tabular}

\vspace{0.3cm}

    \begin{tabular}[t]{lll}
  IKUTSUKA-NO & \underline{KUICHIGAI}-WO & SETSUMEISHITE-KURE-MASHITA.\\
  (several)  & (difference)& (explain)\\
\multicolumn{3}{l}{
  (Tom did John a favor of explaining several \underline{differences}. )}
    \end{tabular}
  \end{minipage}
  \label{eqn:kuichigai}
\end{equation}

\item 
\label{enum:kouho_age}
We take possible antecedents 
from topics or foci in previous sentences. 
We give them some weight of topics and foci 
which means the plausibility of the antecedent 
because topics and foci have various plausibilities. 


\item 
We determine the antecedent 
by combining 
the weight of topics and foci in \ref{enum:kouho_age}, 
the weight of 
semantic similarity in ``X NO Y'' or a verb case frame dictionary, 
and the weight of the distance 
between an anaphor and its possible antecedent. 

\end{enumerate}

For example, when we want to clarify the antecedent of YANE (roof) in the sentences (\ref{eqn:ie_sakuban}), 
we gather examples of ``\verb+<+noun X\verb+>+ NO YANE (roof)'' 
 (roof of \verb+<+noun X\verb+>+), 
and select a possible noun 
which is semantically similar to \verb+<+noun X\verb+>+ as its antecedent. 
Also, when we want to have an antecedent of ``KUICHIGAI (difference)'' in the sentences (\ref{eqn:kuichigai}), 
we select a possible noun 
which satisfies the semantic marker in the case frame of 
``KUICHIGAU (differ)'' in Table \ref{tab:kuitigau_frame} 
or is semantically similar to 
examples of components in the case frame as its antecedent. 

We think that 
errors made by the substitution of a verb case frame for a noun case frame are rare, 
but 
many errors will happen 
when we substitute ``X NO Y'' for a noun case frame. 
This is because 
``X NO Y (Y of X)'' has many semantic relations, 
in particular a feature relation (ex. a man of ability), 
which cannot be an indirect anaphoric relation. 
To reduce the errors, we use the following procedure. 
\begin{enumerate}
\item 
We do not use an example of the form ``noun X NO noun Y (Y of X),'' 
when the noun X is an adjective noun (ex. HONTOU (reality)), 
a numeral, or a temporal noun. 
For example, we do not use ``HONTOU (reality) NO (of) HANNIN (criminal) 
 (a real criminal)''.  
\item 
We do not use an example of the form ``noun X NO noun Y (Y of X),'' 
when the noun Y is a noun 
that cannot be an anaphor of indirect anaphora. 
For example, we do not use 
``noun X NO TSURU (crane)'',``noun X NO NINGEN (human being).'' 
\end{enumerate}
We cannot completely avoid the errors 
by introducing the above procedure, 
but we can reduce the errors to a certain extent. 

We need some more consideration for 
nouns such as ``ICHIBU (part)'',\\
``TONARI (neighbor)'' and ``BETSU (other).'' 
When such a noun is a case component of a verb, 
we use information on semantic constraint of the verb. 
We use a verb case frame dictionary. 
\begin{equation}
  \begin{minipage}[h]{11.5cm}
    \begin{tabular}[t]{llll}
      TAKUSAN-NO & KURUMA-GA  & KOUEN-NI & TOMATTE-ITA.\\
      (many) & (car)  & (in the park) & (there were)\\
\multicolumn{4}{l}{
  (There were many cars in the park.)}\\
\end{tabular}

\vspace{0.3cm}

    \begin{tabular}[t]{lll}
      \underline{ICHIBU}-WA & KITANI  & MUKATTA\\
      (A part (of them)) & (to the north)  & (went) \\
\multicolumn{3}{l}{
  (A \underline{part} of them went to the north.)}\\
    \end{tabular}
  \end{minipage}
\label{eqn:kuruma_itibu}
\end{equation}
In this example, since ``ICHIBU (part)'' is a GA case (subject) 
of a verb ``MUKAU (go),''  
we consult the GA case (subject) of the case frame of ``MUKAU (go).''  
Some noun phrases which can be filled in the case component 
are written in the GA case (subject) of the case frame. 
In this case, 
``KARE (he)'' and ``HUNE (ship)'' are written as examples of things
which can be filled in the case component. 
This indicates that 
the antecedent is semantically similar to 
``KARE (he)'' and ``HUNE (ship).'' 
Since ``TAKUSAN NO KURUMA (many cars)'' is semantically similar to 
``HUNE (ship)'' in the meaning of vehicles, 
it is judged to be the proper antecedent. 

When such a noun as ``TONARI (neighbor or next)'' modifies a noun X 
as ``TONARI NO X'', 
we think that the antecedent is a noun 
which is similar to noun X in meaning. 
\begin{equation}
\hspace*{-2cm}
  \begin{minipage}[h]{11.5cm}
\small
    \begin{tabular}[t]{llll}
      OJIISAN-WA & OOYOROKOBI-WO-SHITE & IE-NI & KAERIMASHITA.\\
      (the old man) & (in great joy)  & (house) & (returned)\\
\multicolumn{4}{l}{
  (The old man returned home (house) in great joy,)}\\
\end{tabular}

\vspace{0.3cm}

    \begin{tabular}[t]{llll}
      OKOTTA & KOTOWO & HITOBITONI & HANASHIMASHITA\\
      (had happened to him) & (all things) & (everybody) & (told)\\
\multicolumn{4}{l}{
  (and told everybody all that had happened to him.)}\\
    \end{tabular}

\vspace{0.3cm}

    \begin{tabular}[t]{lllllll}
      \underline{TONARI}-NO & IE-NI & OJIISAN-GA & MOUHITORI & SUNDE-ORIMASHITA.\\
      (next) & (house) & (old man) & (another) & (live)\\
\multicolumn{7}{l}{
  (There lived in the \underline{next} house another old man. )}\\
    \end{tabular}
  \end{minipage}
\label{eqn:tonari_ie}
\end{equation}
For example, when ``TONARI (neighbor or next)'' 
modifies ``IE (house),'' 
we judge that the antecedent of ``TONARI (neighbor or next)'' is 
``IE (house)'' in the first sentence. 

\section{Anaphora Resolution System}

\subsection{Procedure}
\label{4c_wakugumi}

Analysis of indirect anaphora is performed 
in the same framework of Chapter \ref{chap:noun}. 
At first, 
sentences are transformed into a case structure 
by the case structure analyzer\cite{csan2_ieice}. 
Next, antecedents in indirect anaphora are 
determined by heuristic rules 
for each noun from left to right. 
Using these rules, our system takes possible referents 
and gives them points. It judges that 
the candidate having the maximum total score 
is the desired antecedent. 

The heuristic rules are given in the following form. 

\begin{center}
    \begin{minipage}[c]{10cm}
      \hspace*{0.7cm}{\sl Condition} $\Rightarrow$ \{ {\sl Proposal, Proposal,} .. \}\\
      \hspace*{0.7cm}{\sl Proposal} := ( {\sl Possible-Antecedent,} \, {\sl Point} )
    \end{minipage}
\end{center}

\noindent
Surface expressions, semantic constraints, 
referential properties, and so on, are written as conditions in {\sl Condition} part. 
A possible antecedent is written in {\sl Possible-Antecedent} part. 
{\sl Point} means the plausibility of the possible antecedent. 

\begin{table}[t]
  \caption{The weight (W) in the case of topic}
  \label{fig:shudai_omomi}
  \begin{center}
\small
    \leavevmode
\begin{tabular}[c]{|l|l|r|}\hline
\multicolumn{1}{|c|}{Surface Expression}  & \multicolumn{1}{|c|}{Example}   & Weight \\\hline
Pronoun/Zero-Pronoun GA/WA&  (\underline{John}GA (subject))SHITA (done). &21 \\\hline
Noun WA/NIWA        &  \underline{John}WA (subject)SHITA (do).   &20 \\\hline
\end{tabular}
  \end{center}
\end{table}

\begin{table}[t]
  \caption{The weight (W) in the case of focus}
  \label{fig:shouten_omomi}
  \begin{center}
\small
    \leavevmode
\begin{tabular}[c]{|p{5cm}|l|r|}\hline
\multicolumn{1}{|p{5cm}|}{Surface Expression }  & \multicolumn{1}{|c|}{Example}   & Weight \\\hline
Pronoun/Zero-Pronoun WO(object)/NI(to) /KARA(from) & (\underline{John}NI (to))SHITA (done). & 16 \\\hline
Noun GA (subject)/MO/DA/NARA & \underline{John}GA (subject)SHITA (do).   & 15 \\\hline
Noun WO (object)/NI/, /.       & \underline{John}NI (object)SHITA (do).   & 14 \\\hline
Noun HE (to)/DE (in)/KARA (from)   & \underline{GAKKOU (school)}HE (to)IKU (go).   & 13 \\\hline
\end{tabular}
  \end{center}
\end{table}

\subsection{Heuristic Rule for Estimating Antecedents}
\label{sec:4c_rule}

Resolution of indirect anaphora is performed 
by adding the rules for indirect anaphora resolution 
to the rules for direct anaphora resolution. 
We wrote 12 heuristic rules for noun phrase anaphora resolution 
in Chapter~\ref{chap:noun}. 
The rules (from R1 to R8) 
for noun phrase direct anaphora are shown in Section \ref{sec:3c_rule}. 
The rules 
for noun phrase indirect anaphora are shown as follows. \\

\begin{enumerate}

\item[R9]  
  When a noun phrase Y is not a verbal noun, $\Rightarrow$\\
  \{
  (A topic which has the weight $W$ and the distance $D$, \, $W-D+P+S$),\\ 
  (A focus which has the weight $W$ and the distance $D$, \, $W-D+P+S$),\\ 
  (A subject in a subordinate clause or a main clause of the clause, \, $23+P+S$)\}\\

  The weights $W$ of topics and foci are given 
  in Table~\ref{fig:shudai_omomi} and Table~\ref{fig:shouten_omomi}, 
  respectively, 
  and represent preference of the desired antecedent. 
  The distance $D$ is the number of the topics (foci) 
  between the anaphor and a possible antecedent 
  which is a topic (focus). 
  The value $P$ is given 
  in Table~\ref{tab:4c_teimeishidenai_doai} 
  by the score of the definiteness 
  in referential property analysis described in Chapter \ref{chap:ref}. 
  This is because it is easier for a definite noun phrase 
  to have the antecedent than for an indefinite noun phrase. 
  The value $S$ is the semantic similarity 
  between a possible antecedent and a Noun X of ``Noun X NO Noun Y''. 
  The semantic similarity is given by the similarity level 
  in ``Bunrui Goi Hyou''\cite{BGH} as Table \ref{tab:non-verbal nouns}. 
 
\begin{table}[t]
  \caption{The plausibility (P) that the referential property is a definite}
  \label{tab:4c_teimeishidenai_doai}
  \begin{center}
    \leavevmode
    \small
    \begin{tabular}[h]{|p{10cm}|r|}\hline
      The score in the estimation of the referential property & Plausibility $P$\\\hline
      When the score of the definite noun phrase 
      is the best & 5 \\\hline
      When the score of the definite noun phrase 
      is equal to the score of the indefinite noun phrase or 
      the generic noun phrase & 0\\\hline
      When the score of the definite noun phrase 
      is 1 lower than
      the score of the indefinite noun phrase or 
      the generic noun phrase & $-5$\\\hline
      When the score of the definite noun phrase 
      is 2 lower than 
      the score of the indefinite noun phrase or 
      the generic noun phrase & $-10$\\\hline
      When the score of the definite noun phrase 
      is more than 2 lower than
      the score of the indefinite noun phrase or 
      the generic noun phrase & $-\infty$\\\hline
    \end{tabular}
  \end{center}
\end{table}

\begin{table}[t]
\vspace*{-0.8cm}
  \leavevmode
    \caption{Points given to non-verbal nouns 
      by the semantic similarity}
    \label{tab:non-verbal nouns}

  \begin{center}
\begin{tabular}[c]{|l|r|r|r|r|r|r|r|r|}\hline
Similarity Level & 0   & 1  & 2  & 3 & 4 & 5 & 6 & Exact Match\\\hline
Point   & $-$10 & $-$2 & 1 & 2 & 2.5 & 3 & 3.5 & 4\\\hline
\end{tabular}
\end{center}
\end{table}

\item[R10]  
  \label{enum:sahen_meishi_jap}
  When a noun phrase is a verbal noun, $\Rightarrow$\\
  \{ (analyze in Zero Pronoun Resolution Module in Chapter \ref{chap:deno}, \, 20)\}\\
  In Zero Pronoun Resolution Module, 
  indirect anaphora is resolved 
  using the semantic constraint in a verb case frame and the distance between 
  an anaphor and an antecedent. 

\item[R11]   
  When a noun phrase is a noun such as ``ICHIBU'' and ``TONARI'',
  and it modifies a noun X, $\Rightarrow$\\
  \{ (the same noun as the noun X,  \, $30$)\}

\item[R12]   
  When a noun phrase is a noun such as ``ICHIBU'' and ``TONARI'',
  and it is a case component of a verb, $\Rightarrow$\\
  \{ (analyze in the module similar to R10,  \, $30$)\}


\end{enumerate}

\begin{figure}[t]
\fbox{
\begin{minipage}[h]{13.1cm}
\small

    \begin{tabular}[t]{lll}
  KONO DORUDAKA-WA     & KYOUCHOU-WO & GIKUSHAKU SASETEIRU.\\
  (The dollar's surge) & (cooperation)       & (is straining)\\
\multicolumn{3}{l}{
  (The dollar's surge is straining the cooperation. )}
    \end{tabular}

\hspace{0.3cm}

    \begin{tabular}[t]{l@{ }l@{ }l@{ }l@{ }l}
      {\footnotesize JIKOKUTSUUKA-WO}        & {\footnotesize MAMOROUTO} & SEIDOKU-GA   & {\footnotesize \underline{KOUTEIBUAI-WO}} & {\footnotesize AGETA}.\\
  (own currency)       & (to protect)   & (West Germany)& (official rate)& (raised)  \\
\multicolumn{5}{l}{
  (West Germany raised \underline{its official rate} to protect the Mark. )}
    \end{tabular}

\vspace{0.5cm}

\hspace*{-0.25cm}
\begin{tabular}[h]{|@{ }l@{ }l@{}|@{ }r@{ }|@{ }r@{ }|@{ }r@{ }|@{ }r@{ }|@{ }r@{ }|}\hline
\multicolumn{2}{|l|}{}        & Indefinite  & SEIDOKU    & {\footnotesize JIKOKUTSUUKA} &  KYOUCHOU & DORUDAKA \\\hline
\multicolumn{2}{|l|}{}        &             & West Germany& own currency    & cooperation      & dollar's surge \\\hline
\multicolumn{2}{|@{ }l|}{R6}      &   10        &             &           &            &           \\\hline
\multicolumn{2}{|@{ }l|}{R9}      &             &   25        &  $-23$    &  $-24$     &  $-17$    \\\hline
       &Subject             &             &   23        &           &            &           \\
      &T-F$(W)$      &             &             &    14     &    14      &    20     \\
     &Distance$(D)$          &             &             &   $-2$    &   $-3$     &   $-2$    \\
           &Definite$(P)$&             &  $-5$       &   $-5$    &   $-5$     &   $-5$    \\
            &Similarity$(S)$ &             &    7        &  $-30$    &  $-30$     &   $-30$   \\\hline
\multicolumn{2}{|@{ }l|}{Total Score} &   10    &   25        &  $-23$    &  $-24$     &   $-17$    \\\hline
\end{tabular}

\vspace{0.5cm}

Examples of ``noun X NO KOUTEIBUAI (official rate)''

\hspace{0.5cm}
``NIHON (Japan) NO KOUTEIBUAI (official rate)'',

\hspace{0.5cm}
``BEIKOKU (USA) NO KOUTEIBUAI (official rate)''

\caption{Example of indirect anaphora resolution}
\label{tab:4c_dousarei}
\end{minipage}
}\end{figure}

\subsection{Example of Analysis}

An example of resolution of indirect anaphora is shown 
in Figure \ref{tab:4c_dousarei}. 
Figure \ref{tab:4c_dousarei} shows that 
the noun ``KOUTEI BUAI (official rate)'' is analyzed well. 
This is explained as follows. 

The system estimated the referential property of 
``KOUTEI BUAI (official rate)'' to be indefinite in the method 
described in Chapter 2. 
By the rule R6 in Section \ref{sec:3c_rule} 
the system took a candidate ``Indefinite''. 
When the candidate ``Indefinite'' has the best score, 
the system does not analyze indirect anaphora. 
By the rule R9 in Section \ref{sec:4c_rule} 
the system took four possible antecedents, 
SEIDOKU (West Germany), JIKOKUTSUUKA (own currency), KYOUCHOU (cooperation), 
DORUDAKA (dollar's surge). 
The possible antecedents were given some points from 
the weight of topics and foci, the distance from the anaphor, and so on. 
The system properly judged that 
SEIDOKU (West Germany), which had the best score, 
was the desired antecedent.

\section{Experiment and Discussion}

Before determining antecedents in indirect anaphora, 
sentences were transformed into a case structure 
by the case analyzer\cite{csan2_ieice} 
as in Chapter \ref{chap:noun}. 
The errors made by the analyzer were corrected by hand. 
We used IPAL dictionary\cite{ipal} as a verb case frame dictionary. 
We used the Japanese Co-occurrence Dictionary\cite{edr_kyouki_2.1} 
as a source of examples for ``X NO Y''. 

\begin{table}[p]
\fbox{
\begin{minipage}[h]{13cm}
\small
    \caption{Result}
    \label{tab:sougoukekka}
\vspace*{0.5cm}
  \begin{center}
\begin{tabular}[c]{|r@{}c@{ }|r@{}c@{ }|r@{}c@{ }|r@{}c@{ }|r@{}c@{ }|r@{}c@{ }|}\hline
        \multicolumn{4}{|c|}{Non-verbal Noun} &\multicolumn{4}{c|}{Verbal Noun}  &\multicolumn{4}{c|}{Total}\\\cline{1-12}
        \multicolumn{2}{|c|}{Recall}
        &\multicolumn{2}{c|}{Precision}
        &\multicolumn{2}{c|}{Recall}
        &\multicolumn{2}{c|}{Precision}
        &\multicolumn{2}{c|}{Recall}
        &\multicolumn{2}{c|}{Precision}\\\hline
\multicolumn{12}{|c|}{Experiment in the case that the system does not use any semantic information}\\\hline
 85\% & (56/66) & 67\% & (56/83) & 40\% & (14/35) & 44\% & (14/32) & 69\% & (70/101)& 61\% & (70/115)\\\hline
 53\% & (20/38) & 50\% & (20/40) & 47\% & (15/32) & 42\% & (15/36) & 50\% & (35/70) & 46\% & (35/76)\\\hline
\multicolumn{12}{|c|}{Experiment using ``X NO Y'' and verb case frame}\\\hline
 91\% & (60/66) & 86\% & (60/70) & 66\% & (23/35) & 79\% & (23/29) & 82\% & (83/101)& 84\% & (83/99)\\\hline
 63\% & (24/38) & 83\% & (24/29) & 63\% & (20/32) & 56\% & (20/36) & 63\% & (44/70) & 68\% & (44/65)\\\hline
\multicolumn{12}{|c|}{Estimation for the hypothetical case when we can use noun case frame dictionary}\\\hline
 91\% & (60/66) & 88\% & (60/68) & 69\% & (24/35) & 89\% & (24/27) & 83\% & (84/101)& 88\% & (84/95)\\\hline
 79\% & (30/38) & 86\% & (30/35) & 63\% & (20/32) & 77\% & (20/26) & 71\% & (50/70) & 82\% & (50/61)\\\hline
\end{tabular}
\end{center}


The upper row and the lower row of this table 
show rates on training sentences and 
 test sentences, respectively. 

The training sentences are used to 
set by hand the values given in rules in Section~\ref{sec:4c_rule}. \\
{
Training sentences \{example sentences \cite{walker2} (43 sentences), a folk tale ``KOBUTORI JIISAN''\cite{kobu} (93 sentences), an essay in ``TENSEIJINGO'' (26 sentences), an editorial (26 sentences)\}\\
Test sentences \{a folk tale ``TSURU NO ONGAESHI''\cite{kobu} (91 sentences), two essays in ``TENSEIJINGO'' (50 sentences), an editorial (30 sentences)\}

{\it Precision}\, is 
the fraction  of the noun phrases which were judged to have the antecedents 
of indirect anaphora. 
{\it Recall}\, is the fraction of the noun phrases 
 which have the antecedents of indirect anaphora. 
We use precision and recall to evaluate 
because the system judges that 
a noun which is not an antecedent of indirect anaphora 
is an antecedent of indirect anaphora, 
and we check these errors thoroughly.}
\end{minipage}
}
\end{table}

We show the result of anaphora resolution 
using both ``X NO Y'' and a verb case frame dictionary
in Table \ref{tab:sougoukekka}. 
We obtained a recall rate of 63\% and 
a precision rate of 68\% in the estimation of indirect anaphora 
on  test sentences. 
This indicates that 
the information of ``X NO Y'' 
is useful to a certain extent 
when we cannot make use of the noun frame dictionary. 
We also tested when 
the system does not use 
any semantic information. 
The precision and the recall were lower. 
This indicates that semantic information is necessary. 
The experiment was performed by 
fixing all the semantic similarity values $S$ to 0. 

Further, we made the estimation 
for the hypothetical case when 
we can use a noun case frame dictionary. 
The estimation was made as follows. 
We looked over the errors in the experience 
using ``X NO Y'' and a verb case frame dictionary. 
We regarded 
the errors made by one of the following three reasons as right answers. 
\begin{enumerate}
\item 
Proper examples do not exist in examples of ``X NO Y'' 
or a verb case frame dictionary.  
\item 
Wrong examples exist in examples of ``X NO Y'' 
or a verb case frame dictionary. 
\item 
A noun case frame is different from a verb case frame. 
\end{enumerate}
If we will make a noun case frame dictionary by ourselves, 
the dictionary will have some errors, and 
the success ratio will be lower than 
the ratio in Table \ref{tab:sougoukekka}. 

\subsection*{Discussion of Errors}

Even if we have a noun case frame dictionary, 
there are certain pairs of nouns in indirect anaphoric relation 
that cannot be resolved by our framework. 
\begin{equation}
  \begin{minipage}[h]{11.5cm}
\small
KON'NA HIDOI HUBUKI-NO NAKA-WO ITTAI DARE-GA KITA-NO-KA-TO 
IBUKARINAGARA, OBAASAN-WA IIMASHITA.\\
(Wondering who could have come in such a heavy snowstorm, 
the old woman said:) \\
``DONATA-JANA''\\
(``Who is it?'')\\
TO-WO AKETEMIRUTO, SOKO-NIWA 
ZENSHIN YUKI-DE MASSHIRONI NATTA \underline{MUSUME}-GA 
TATTE ORIMASHITA.\\
(She opened the door, and there stood before 
her \underline{a girl} all covered with snow. )
\end{minipage}
\end{equation}
The underlined ``{MUSUME} (a daughter or a girl)'' has 
two main meanings: a daughter and a girl. 
In the above example, 
``{MUSUME}'' means girl 
and has no indirect anaphora relation. 
But the system incorrectly judged that 
it is the daughter of ``OBAASAN (the old woman)''. 
This is a problem of noun role ambiguity 
and is a very difficult problem to solve. 

The following example is also a difficult problem. 
\begin{equation}
  \begin{minipage}[h]{16cm}
\small
    \begin{tabular}[t]{l@{ }l@{ }l@{ }l@{ }l@{ }l}
      SHUSHOU-WA & \underline{TEIKOU}-NO & {\footnotesize TSUYOI} & SENKYOKU-NO & {\footnotesize KAISHOU-WO} & {\footnotesize MIOKUTTA}.\\
      (prime minister) & ({resistance}) & (very) & (electoral district) & (modification) & (give up)\\
\multicolumn{6}{p{13.5cm}}{
  (The prime minister gave up the modification of some electoral districts where \underline{the resistances} were very hard.)}
    \end{tabular}
  \end{minipage}
\label{eqn:nininku_ayamari}
\end{equation}
The underlined ``TEIKOU (resistance)'' appears 
to refer indirectly to ``SENKYO- KU (electoral district)'' 
from the surface expression. 
But actually 
``TEIKOU (resistance)'' refers to the candidates 
of ``SENKYOKU (electoral district)'' 
not to ``SENKYOKU (electoral district)'' itself. 
To arrive at this conclusion 
it is necessary to use a two step relation, 
``an electoral district $\Longrightarrow$ candidates'', 
``candidates $\Longrightarrow$ resist'' in sequence. 
However it is not easy to change our system to deal with two step relations 
because if we apply the use of two relations to nouns, 
many nouns which are not in an indirect anaphoric relation 
will be incorrectly judged as indirect anaphora. 
A new method is required to infer two relations in sequence. 

\begin{table}[p]
    \caption{Examples of arranged ``X NO Y''}
    \label{tab:noun_bgh}
\small
  \begin{center}
\begin{tabular}{|p{2cm}|p{10.5cm}|}\hline
Noun Y          & Arranged Noun X\\\hline
KOKUMIN (nation)    & $<$Human$>$  AITE (partner) \, $<$Organization$>$ KUNI (country), SENSHINKOKU (an advanced country), RYOUKOKU (the two countries), NAICHI (inland), ZENKOKU (the whole country), NIHON (Japan), SOREN (the Soviet Union),
 EIKOKU (England), AMERIKA (America), SUISU (Switzerland), DENMAAKU (Denmark), SEKAI (the world)\\\hline
GENSHU (the head of state)    & $<$Human$>$ RAIHIN (visitor) \, $<$Organization$>$ GAIKOKU (a foreign country), KAKKOKU (each country), POORANDO (Poland)\\\hline
YANE (roof)    & $<$Organization$>$ HOKKAIDO (Hokkaido), SEKAI (the world), GAKKOU (school), KOUJOU (factory), GASORINSUTANDO (gas station), SUUPAA (supermarket),
 JITAKU (one's home), HONBU (the head office) \, $<$Product$>$ KURUMA (car), JUUTAKU (housing), IE (house), SHINDEN (temple), GENKAN (entrance),
 SHINSHA (new car) $<$Phenomenon$>$ MIDORI (green) $<$Action$>$ KAWARABUKI (tile-roofed)
 $<$Mental$>$ HOUSHIKI (method) $<$Character$>$ KEISHIKI (form)\\\hline
MOKEI (model)  & $<$Animal$>$ ZOU (elephant) \,$<$Nature$>$ FUJISAN (Mt. Fuji) \,$<$Product$>$ IMONO (an article of cast metal), MANSHON (an apartment house),
 KAPUSERU (capsule), DENSHA (train), HUNE (ship), GUNKAN (warship), HIKOUKI (airplane), JETTOKI (jet plane) \,$<$Action$>$ ZOUSEN (shipbuilding)
 $<$Mental$>$ PURAN (plan) \, $<$Character$>$ UNKOU (movement)\\\hline
GYOUJI (event)  & $<$Human$>$ KOUSHITSU (the Imperial Household), OUSHITSU (a Royal family), IEMOTO (the head of a school) \, $<$Organization$>$ NOUSON (an agricultural village), KEN (prefecture), NIHON (Japan),
SOREN (the Soviet Union), TERA (temple), GAKKOU (school) \, $<$Action$>$ SHUUNIN (take up one's post), MATSURI (festival),
 IWAI (celebration), JUNREI (pilgrimage) \, $<$Mental$>$ KOUREI (an established custom), KOUSHIKI (formal) \\\hline
JINKAKU (personality) & $<$Human$>$ WATASHI (myself), NINGEN (human), SEISHOUNEN (young people), SEIJIKA (statesman) \\\hline
\end{tabular}
\end{center}
\end{table}

\mysection
{\section{Consideration of Construction of Noun Case Frame Dictionary}}
{\section[Construction of Noun Case Frame Dictionary]
{Consideration of Construction of Noun Case\\ Frame Dictionary}}

We used ``X NO Y (Y of X)'' to 
resolve indirect anaphora. 
But we will get a higher accuracy rate 
if we can utilize a good noun case frame dictionary. 
Therefore we have to consider 
how we can construct a noun case frame dictionary. 
A key is to get the detailed meaning of ``NO (of)'' in ``X NO Y''. 
If it is automatically obtainable, 
a noun case frame dictionary will be constructed automatically. 
If the semantic analysis of ``X NO Y'' is not done well, 
how do we construct the dictionary? 
We think that it is still good to construct it using ``X NO Y''. 
For example, 
we arrange ``noun X NO noun Y'' in the order of 
the meaning of ``noun Y'', 
arrange them in the order of the meaning of ``noun X'', 
delete some of them 
whose ``noun X'' are adjective nouns, 
and obtain Table \ref{tab:noun_bgh}. 
In this case, we use 
the thesaurus dictionary ``Bunrui Goi Hyou''\cite{BGH} 
to get the meanings of nouns. 
We think that it is not difficult to construct a noun case frame dictionary 
from Table \ref{tab:noun_bgh} by hand. 
We will make a noun case frame dictionary 
by removing ``AITE (partner)'' in the line of ``KOKUMIN (nation)'', 
``RAIHIN (visitor)'' in the line of ``GENSHU (the head of state)'', 
and noun phrases which mean characters and features. 
When we look over the noun phrases in a certain line 
and almost all of them mean countries, 
we will also include the feature 
that countries are easy to be filled 
by using semantic markers. 
When we make a noun case frame dictionary, 
we must remember that 
examples of ``X NO Y'' are insufficient, 
and must add examples. 
Since examples are arranged in the order of meaning in this method, 
it will not be so difficult to add examples. 

\section{Summary}

We presented how to resolve indirect anaphora in Japanese nouns. 
When we analyze indirect anaphora, 
we need a noun case frame dictionary 
containing information about noun relations. 
But no noun case frame dictionary exists at present. 
Therefore, we used examples of ``X NO Y (Y of X)'' 
and a verb case frame dictionary. 
We experimented with the estimation of indirect anaphora 
by using this information, 
and obtained a recall rate of 63\% and 
a precision rate of 68\% on  test sentences. 
This indicates that 
the information of ``X NO Y'' 
is useful 
when we cannot make use of a noun case frame dictionary. 
We made an estimation in 
the case that we can use a noun case frame dictionary, 
and obtained results with the recall and the precision rates of 
71\% and 82\%, respectively. 
Finally we proposed how to construct a noun case frame dictionary 
from examples of ``X NO Y''. 

\newenvironment{indention}[1]{\par\begingroup\addtolength{\leftskip}{#1}}{\par\endgroup}
\chapter{An Estimate of Referents of Pronouns}
\label{chap:deno}

\section{Overview}
\label{sec:deno_overview}

We described in Chapter \ref{chap:noun} and Chapter \ref{chap:indian} 
how to estimate the referents of noun phrases. 
This chapter 
describes how to resolve the referents of pronouns: 
demonstrative pronouns, personal pronouns, and zero pronouns. 
Pronoun resolution is especially important 
for machine translation. 
For example, if the system cannot resolve zero pronouns 
\footnote{Ellipses of noun phrases are called {\it zero pronouns}.}, 
the system cannot translate sentences with them from Japanese into English. 
When the word order of sentences is changed and 
the pronominalized words are changed in translating into English, 
the system must detect the referents of the pronouns. 

There has been much work done 
in pronoun resolution \, 
\cite{Tanaka1} \,  
\cite{kameyama1} \,\,
\cite{yamamura92_ieice} \,\, \cite{walker2} \,\, \cite{takada1} \, 
\cite{nakaiwa}. 
Major distinguishing features of our work are as follows: 

\begin{itemize}
\item 
In conventional pronoun resolution methods, 
semantic markers have been used 
for semantic constraints. 
On the other hand, we use examples for semantic constraints 
and show in our experiments that 
examples are as useful as semantic markers. 
The result is important 
because 
the cost of constructing the case frame 
using semantic markers is generally higher than 
the cost of constructing the case frame 
using examples. 

\item 
We use examples in the form ``X of Y'' for 
estimating referents of demonstrative adjectives. 

\item 
We deal with the case 
when a demonstrative refers to 
elements which appear later. 

\item 
We resolve a personal pronoun in quotation 
by estimating the speaker and the hearer. 
\end{itemize}

In this work, 
we used almost all the potentials of conventional methods 
and proposed new method. 

In Section \ref{5c_wakugumi}, 
we explain how the system estimates the referent of a pronoun. 
Next, we explain the rules 
for demonstratives, personal pronouns, and 
zero pronouns in Sections \ref{sec:sijisi_ana}, 
\ref{sec:pro_ana}, 
and \ref{sec:zero_ana}, respectively. 
In Section \ref{sec:jikken}, 
we report the results of experiments 
using these rules. 
In Section \ref{sec:owari}, 
we conclude this chapter. 

\section{The Framework for Estimating the Referent}
\label{5c_wakugumi}

Pronoun resolution is performed 
in the framework similar to that in 
Chapter \ref{chap:noun} and Chapter \ref{chap:indian}. 
The antecedents of pronouns are 
determined by heuristic rules 
from left to right. 
Using these rules, 
our system gives possible antecedents points, and it judges that 
the possible antecedent having the maximum total score 
is the desired antecedent. 

Heuristic rules are classified into 
two kinds of rules: 
{\it Candidate enumerating rule}s 
and {\it Candidate judging rule}s. 
{\it Candidate enumerating rule}s 
are used in enumerating  candidate antecedents and 
giving them points (which mean plausibility of the proper antecedent). 
{\it Candidate judging rule}s 
are used in 
giving the candidate antecedents taken by {\it Candidate enumerating rule}s
points. 
These rules are shown in Figure \ref{fig:kouho_rekkyo} 
and Figure \ref{fig:kouho_hantei}. 
Surface expressions, semantic constraints, 
referential properties, etc., are written as conditions in {\sl Condition} part. 
A possible antecedent is written in {\sl Possible-Antecedent} part. 
{\sl Point} means the plausibility of the possible antecedent. 

\begin{figure}[t]
  \leavevmode
  \begin{center}
\fbox{
    \begin{minipage}[c]{10cm}
      \hspace*{0.7cm}Condition $\Rightarrow$ \{Proposal Proposal ..\}\\[-0.1cm]
      \hspace*{0.7cm}Proposal := ( Possible-Antecedent \, Points )

    \caption{Form of {\it Candidate enumerating rule}}
    \label{fig:kouho_rekkyo}
    \end{minipage}
}
  \end{center}
\end{figure}

\begin{figure}[t]
  \leavevmode
  \begin{center}
\fbox{
    \begin{minipage}[c]{10cm}
      \hspace*{1.5cm}Condition $\Rightarrow$ ( Points )
    \caption{Form of {\it Candidate judging rule}}
    \label{fig:kouho_hantei}
    \end{minipage}
}
  \end{center}
\end{figure}

An estimation of the referent is performed 
by using the total scores of possible antecedents given 
by {\it Candidate enumerating rule}s and {\it Candidate judging rule}s. 
First, the system applies 
all {\it Candidate enumerating rule}s to the anaphor 
and enumerates candidate antecedents having the points. 
Next, 
the system applies 
all {\it Candidate judging rule}s to all the candidate antecedents 
and sums up the score of each candidate antecedent. 
Consequently, 
the system judges the candidate antecedent 
having the best score is the proper antecedent. 
If the candidate referents having the best score are  
plural, 
the candidate referent taken in the first order 
\footnote{
The order is based on the order applying rules. 
}
is judged as 
the proper antecedent. 

We made 50 {\it Candidate enumerating rule}s and 
10 {\it Candidate judging rule}s for analyzing demonstratives, 
4 {\it Candidate enumerating rule}s and 
6 {\it Candidate judging rule}s for analyzing personal pronouns, 
and 19 {\it Candidate enumerating rule}s and 
4 {\it Candidate judging rule}s for analyzing zero pronouns. 
All of the rules are described in Appendix \ref{c:ap:pronoun}. 
Some of the rules are described in the following sections. 

\section{Heuristic Rule for Demonstrative}

\label{sec:sijisi_ana}

We made 
heuristic rules for demonstratives 
by consulting the papers 
of \cite{seiho1} 
\cite{hyasi2}\cite{sijisi_nihongogaku}\cite{sijisi} 
and examining Japanese sentences by hand. 
Demonstratives 
have three categories: 
demonstrative pronouns,  
demonstrative adjectives,  and 
demonstrative adverbs. 
In the following sections, we explain the rules for 
analyzing demonstratives. 

\begin{table}[t]
  \caption{The weight in the case of topic}
  \label{tab:5c_topic}
  \begin{center}
\small
    \leavevmode
\begin{tabular}[c]{|l|l|r|}\hline
\multicolumn{1}{|c|}{Surface Expression}  & \multicolumn{1}{|c|}{Example}   & Weight \\\hline
Pronoun/Zero-Pronoun GA/WA&  (\underline{John}GA (subject))SHITA (done). &21 \\\hline
Noun WA/NIWA        &  \underline{John}WA (subject)SHITA (do).   &20 \\\hline
\end{tabular}
  \end{center}
\end{table}

\begin{table}[t]
  \caption{The weight in the case of focus}
  \label{tab:5c_focus}
  \begin{center}
\small
    \leavevmode
\begin{tabular}[c]{|p{5cm}|l|r|}\hline
\multicolumn{1}{|p{5cm}|}{Surface Expression }  & \multicolumn{1}{|c|}{Example}   & Weight \\\hline
Pronoun/Zero-Pronoun WO(object)/NI(to) /KARA(from) & (\underline{John}NI (to))SHITA (done). & 16 \\\hline
Noun GA (subject)/MO/DA/NARA & \underline{John}GA (subject)SHITA (do).   & 15 \\\hline
Noun WO (object)/NI/, /.       & \underline{John}NI (object)SHITA (do).   & 14 \\\hline
Noun HE (to)/DE (in)/KARA (from)   & \underline{GAKKOU (school)}HE (to)IKU (go).   & 13 \\\hline
\end{tabular}
  \end{center}
\end{table}

\subsection{Rule for Demonstrative Pronoun}
\label{sec:meishi_siji}

\subsection*{\underline{Rule in the Case when
the Referent is a Noun Phrase}}

\noindent
{\bf {\it Candidate enumerating rule}1}
\begin{indention}{0.8cm}\noindent
  When a pronoun is a demonstrative pronoun or 
  ``SONO (of it) / KONO (of this) / ANO (of that)'',\\
  \{(A topic which has the weight $W$ and the distance $D$, \, $W-D-2$)\\
  (A focus which has the weight $W$ and the distance $D$, \, $W-D+4$)\}\\
  This bracket expression represents 
  the lists of proposals in Figure \ref{fig:kouho_rekkyo}. 
  The definition and the weight $W$ of 
  topic and focus are shown in 
  Table \ref{tab:5c_topic} and Table \ref{tab:5c_focus}. 
  The distance ($D$) is the number of topics and foci 
  between the demonstrative and the possible referent. 
  Since a demonstrative more often refer to 
  foci than a zero pronoun, 
  we add the coefficient $-2$, $+4$ as compared with 
  the heuristic rules in zero pronoun resolution. 
\end{indention}
\vspace{0.5cm}

The score (in other words, 
the certification value) of a candidate referent
depends on 
the weight of topics/foci and the geographical distance 
between the demonstrative and the candidate referent. 

\subsection*{\underline{Rule when 
the Referent is a Verb Phrase}}

\noindent
{\bf {\it Candidate enumerating rule}2}
\begin{indention}{0.8cm}\noindent
  When a pronoun is  ``SORE/ARE/KORE'' or 
  a demonstrative adjective,\\
  \label{kore_mae}
  \{(
  The previous sentence (or the verb phrase 
  which is a conditional form containing a conjunctive particle such as
  ``GA (but)'', `` DAGA (but)'', and ``KEREDO (but)''
  if the verb phrase is in the same sentence), 
   \,$15$)\}
\end{indention}
\vspace{0.5cm}

The following is an example 
of a pronoun referring to the verb phrase of the previous sentence. 
\begin{equation}
  \begin{minipage}[h]{11.5cm}
\small
  \begin{tabular}[t]{lll}
TENGU-TACHI-WA & MAMONAKU & YATTEKITE\\
 (The tengus) & (presently) & (came) \\
\multicolumn{3}{l}{
(Presently, they came)}\\
\end{tabular}

\vspace{0.3cm}

  \begin{tabular}[t]{l@{ }l@{ }l@{ }l}
MAENOBAN-NO-YOUNI & UTATTARI & ODOTTARI & SHI-HAJIMEMASHITA.\\
 (the previous night) & (sing) & (dance) & (begin to do) \\
\multicolumn{4}{l}{
(and began singing  
and dancing just as they had done the previous night.)}\\
\end{tabular}

\vspace{0.3cm}

  \begin{tabular}[t]{l@{ }l@{ }l@{ }l@{ }l}
OJIISAN-WA & \underline{SORE}-WO & MITE, & KON'NAHUUNI & UTAI-HAJIMEMASHITA.\\
 (the old man) & (it) & (see) & (as follows) & (begin to sing)\\
\multicolumn{5}{l}{
(When the old man saw \underline{this}, he began to sing as follows. )}\\
\end{tabular}
  \end{minipage}
\label{eqn:sore_mite_utau}
\end{equation}
In these sentences, 
a demonstrative pronoun ``SORE (it)'' refers to 
the event 
``TENGUTACHI-GA UTATTARI ODOTTARI SHI-HAJIMEMASHITA
(tengu began singing  
and dancing just as they had done the previous night.)''.

The following is an example 
of a pronoun referring to a verb phrase
(the event) containing 
a conjunctive particle such as
``GA'', ``DAGA'', and ``KEREDO''
 in the same sentence. 
\begin{equation}
  \begin{minipage}[h]{11.5cm}
    \small
  \begin{tabular}[t]{l@{ }l@{ }l@{ }l@{ }l}
OJIISAN-WA & ISSHOUKENMEINI & UTAI & SOSHITE & ODORIMASHITAGA,\\
 (the old man) & (one's best) & (sing) & (and) & (dance)\\
\multicolumn{5}{l}{
(The man did his best singing and dancing,)}\\
\end{tabular}

\vspace{0.3cm}

  \begin{tabular}[t]{lll}
\underline{SORE}-WA & KOTOBADE-IIARAWASENAIHODO & HETAKUSODESHITA.\\
 (they) & (unspeakably) & (poor) \\
\multicolumn{3}{l}{
( but \underline{they} were unspeakably poor.)}\\
\end{tabular}
  \end{minipage}
\label{eqn:sore_mite_heta}
\end{equation}

\begin{table}[t]
\vspace{-0.8cm}
  \leavevmode
    \caption{Points given in the case of demonstrative pronouns}
    \label{tab:hininshoudaimeisi_ruijido}

  \begin{center}
\begin{tabular}[c]{|l|@{\hspace{0.12cm}}r@{\hspace{0.12cm}}|@{\hspace{0.12cm}}r@{\hspace{0.12cm}}|@{\hspace{0.12cm}}r@{\hspace{0.12cm}}|@{\hspace{0.12cm}}r@{\hspace{0.12cm}}|@{\hspace{0.12cm}}r@{\hspace{0.12cm}}|@{\hspace{0.12cm}}r@{\hspace{0.12cm}}|@{\hspace{0.12cm}}r@{\hspace{0.12cm}}|@{\hspace{0.12cm}}r@{\hspace{0.12cm}}|}\hline
Similarity Level & 0 & 1 & 2 & 3 & 4 & 5 & 6 & Exact Match\\\hline
Point   & 0 & 0 & $-$10 & $-$10 & $-$10 & $-$10& $-$10& $-$10\\\hline
\end{tabular}
\end{center}
\end{table}

\subsection*{\underline{Rule Using 
the Feature that 
Demonstrative Pronouns usually}\\ \underline{do not Refer to People}}

\noindent
{\bf {\it Candidate judging rule}1}
\begin{indention}{0.8cm}\noindent
  When a pronoun is a demonstrative pronoun and 
  a candidate referent has a semantic marker {\sf HUM} (human), 
  it is given $-10$. 
  We use 
  Noun Semantic Marker Dictionary\cite{imiso-in-BGH} 
  as a semantic marker dictionary. 
\end{indention}

\begin{table}[p]
\vspace*{-0.8cm}
  \leavevmode
    \caption{Modification of category number of ``BUNRUI GOI HYOU''}
    \label{tab:bunrui_code_change}
  \begin{center}
\begin{tabular}[c]{|@{\hspace{0.15cm}}l@{\hspace{0.15cm}}|@{\hspace{0.15cm}}l@{\hspace{0.15cm}}|@{\hspace{0.15cm}}l@{\hspace{0.15cm}}|}\hline
Semantic Marker   & Original       & Modified\\
                  &   code            &     code\\\hline
ANI(animal)         &  156                & 511\\[0cm]
HUM(human)         &  12[0-4]            & 52[0-4]\\[0cm]
ORG(organization)   &  125,126,127,128    & 535,536,537,538\\[0cm]
PLA(plant)         &  155                & 611\\[0cm]
PAR(part of living thing)   &  157                & 621\\[0cm]
NAT(natural)       &  152                & 631\\[0cm]
PRO(products) &  14[0-9]            & 64[0-9]\\[0cm]
LOC(location)   &  117,125,126        & 651,652,653\\[0cm]
PHE(phenomenon)     &  150,151            & 711,712\\[0cm]
ACT(action)   &  13[3-8]            & 81[3-8]\\[0cm]
MEN(mental)         &  130                & 821\\[0cm]
CHA(character)         &  11[2-58],158       & 83[2-58],839\\[0cm]
REL(relation)         &  111                & 841\\[0cm]
LIN(linguistic products)     &  131,132            & 851,852\\[0cm]
The others            &  110                & 861\\[0cm]
TIM(time)         &  116                & a11\\[0cm]
QUA(quantity)         &  119                & b11\\[0cm]\hline
\end{tabular}

``125'' and ``126'' are given two category number. 
\end{center}
\end{table}

\vspace{0.5cm}
\noindent
{\bf {\it Candidate judging rule}2}
\begin{indention}{0.8cm}\noindent
  When a pronoun is a demonstrative pronoun, 
  a candidate referent is given the points 
  in Table \ref{tab:hininshoudaimeisi_ruijido} 
  by using the highest semantic similarity 
  between the candidate referent and 
  the codes 
  \{5200003010 5201002060 5202001020 5202006115 5241002150 5244002100\}
  in ``Bunrui Goi Hyou (BGH)'' \cite{BGH} which signify human beings. 
  When we calculate the semantic similarity, 
  we use the modified code table 
  in Table \ref{tab:bunrui_code_change}.
  The reason for this modification is that 
  some codes in BGH \cite{BGH} are incorrect. 
\end{indention}
\vspace{0.5cm}

These rules use the feature that 
a demonstrative pronoun rarely refer to people, 
and reduce candidates of the referent. 
For example, 
we find 
``SORE (it)'' in the following sentences refers to 
``KONPYUUTA (computer)'', 
because 
``SORE (it)'' refers to only a thing which is not human 
and the noun which is near ``SORE (it)'' 
and which is not human is only ``KONPYUUTA (computer)''. 
\begin{equation}
  \begin{minipage}[h]{11.5cm}
  \begin{tabular}[t]{llll}
TAROO-WA & SAISHIN-NO & KONPYUUTA-WO & KAIMASHITA.\\
 (Taroo) & (new) & (computer) & (buy)\\
\multicolumn{4}{l}{
(Taroo bought a new computer.)}\\
\end{tabular}

\vspace{0.3cm}

  \begin{tabular}[t]{llll}
JON-NI & SASSOKU & \underline{SORE}-WO & MISEMASHITA.\\
 (John) & (at once) & (it) & (show)\\
\multicolumn{4}{l}{
([Taroo] showed \underline{it} at once to John. )}
\end{tabular}
  \end{minipage}
\label{eqn:5c_sore_new_computer}
\end{equation}

\begin{table}[t]
\vspace*{-0.8cm}
  \leavevmode
    \caption{Points given demonstrative pronouns 
      which refer to places}
    \label{tab:bashomeisi_ruijido}

  \begin{center}
\begin{tabular}[c]{|l|r|r|r|r|r|r|r|r|}\hline
Similarity Level & 0 & 1 & 2 & 3 & 4 & 5 & 6 & Exact Match\\\hline
Point   & $-$10 & $-$5 & 0 & 5 & 10 & 10& 10& 10\\\hline
\end{tabular}
\end{center}
\end{table}

\subsection*{\underline{
Rule with Feature that 
``KOKO'' and ``SOKO'' Often Refer}\\ \underline{to Locations}}

\noindent
{\bf {\it Candidate judging rule}3}
\begin{indention}{0.8cm}\noindent
  When a pronoun is ``KOKO (here) / SOKO (there) / ASOKO (over there)'' 
  and a candidate referent has 
  a semantic marker {\sf LOC} (location), 
  the candidate referent is given $10$ points. 
\end{indention}

\vspace{0.5cm}
\noindent
{\bf {\it Candidate judging rule}4}
\begin{indention}{0.8cm}\noindent
  When a pronoun is ``KOKO/SOKO/ASOKO'', 
  a candidate referent is given the points in 
  Table \ref{tab:bashomeisi_ruijido} 
  by using the semantic similarity 
  between the candidate referent and 
  the codes 
  \{6563006010 6559005020 9113301090 9113302010 6471001030 6314020130\}
  which signify locations in BGH \cite{BGH}. 
\end{indention}
\vspace{0.5cm}

``SOKO (there)'' commonly refers to location. 
For example, 
``SOKO'' in the following sentences refers to 
``BAITEN (shop)'' which signifies location. 
\begin{equation}
  \begin{minipage}[h]{11.5cm}
  \begin{tabular}[t]{llll}
TAROO-GA & KOUEN-DE & HON-WO & YONDE-IMASHITA.\\
 (Taroo) & (in the park) & (book) & (be reading)\\
\multicolumn{4}{l}{
(Taroo was reading a book in the park.)}\\
\end{tabular}

\vspace{0.3cm}

  \begin{tabular}[t]{llll}
KOORA-WO & KAINI & BAITEN-NI & HAIRIMASHITA.\\
 (cola) & (buy) & (shop) & (enter)\\
\multicolumn{4}{l}{
(Taroo entered a shop to buy a cola.)}\\
\end{tabular}

\vspace{0.3cm}

  \begin{tabular}[t]{llll}
JIROO-WA & \underline{SOKO-DE} & GUUZEN & DEKUWASHIMASHITA.\\
 (Jiroo) & (there) & (by chance) & (meet)\\
\multicolumn{4}{l}{
(Jiroo met Taroo \underline{there} by chance. )}\\
\end{tabular}
  \end{minipage}
\label{eqn:soko_dekuwasu}
\end{equation}

\subsection*{\underline{Rule when
``KOKODE'' or ``SOKODE'' is Used as a Conjunction}}

\noindent
{\bf {\it Candidate enumerating rule}3}
\begin{indention}{0.8cm}\noindent
  When a pronoun is ``KOKODE'' or ``SOKODE'', \\
  \{(the pronoun is used as conjunctions, \,$11$)\}
\end{indention}
\vspace{0.5cm}

This rule is for 
when ``KOKODE (here or then)'' or ``SOKODE (there or then)'' 
is used as conjunctions. 
If a word which signifies location 
is not found near ``KOKODE'' or ``SOKODE'', 
the candidate which is listed by this rule has the highest score, and 
``KOKODE'' or ``SOKODE'' is judged as a conjunction. 
By using this rule, 
``SOKODE''  in the following sentences is judged 
to be a conjunction. 
\begin{equation}
  \begin{minipage}[h]{11.5cm}
    \small
  \begin{tabular}[t]{lll}
OJIISAN-WA & TENGU-GA & KOWAKUNAKUNATTE-IMASHITA.\\
(old man) & (tengu) & (lose all fear of)\\
\multicolumn{3}{l}{
(The old man lost all fear of the ``tengu.'')}\\
\end{tabular}

\vspace{0.3cm}

  \begin{tabular}[t]{l@{ }l@{ }l@{ }l@{ }l}
\underline{SOKODE} & OJIISAN-WA & KAKURETEITA & ANA-KARA & DETEKIMASHITA.\\
 (so) & (old man) & (be hiding) & (hole) & (leave)\\
\multicolumn{5}{l}{
(\underline{So}, he left the hole where he had been hiding.)}\\
\end{tabular}
  \end{minipage}
\label{eqn:soko_ojiisan_kakureru}
\end{equation}
This rule is necessary 
when the system translates ``SOKODE'' into English, 
judges whether it is used as a demonstrative or as a conjunction, 
and translates it into ``there'' or ``then.'' 

\subsection*{\underline{Rule in the Case of Cataphora}}

Demonstrative pronouns can be intersentential cataphoric
\footnote{
Cataphora is the phenomenon 
that an anaphor refers to elements 
which appear later. }. 
In this case, 
we analyze a demonstrative pronoun by using rules 
based on Matsuoka's method \cite{matsuoka_nl}. 
This work \cite{matsuoka_nl} 
also deals with cases in which 
demonstrative pronouns refer to the next sentences. 
But these cases rarely happen. 
When we do not use this rule, 
the precision increases. 
For this reason we do not use this rule. 

\subsection*{\underline{The Other Rules}}

\vspace{0.5cm}
\noindent
{\bf {\it Candidate enumerating rule}4}
\begin{indention}{0.8cm}\noindent
  When a pronoun is ``SORE/ARE/KORE'' or 
  a demonstrative adjective and 
  the previous bunsetsu contains 
  the expression of 
  the predicative form of a verb 
  or the expression of 
  enumerating examples such as ``{TOKA} (and so on),''
  \{(the expression, \,$40$)\}
\end{indention}

\vspace{0.5cm}
\noindent
{\bf {\it Candidate enumerating rule}5}
\begin{indention}{0.8cm}\noindent
  When a pronoun is a demonstrative pronoun, 
  a demonstrative adverb, or a demonstrative adjective,\\
  \{(Introduce an individual, \,$10$)\}\\

This rule is used 
when there is no referent of a pronoun in the sentences. 
This rule makes the system 
introduce a certain individual. 
\end{indention}

\subsection{Rule for Demonstrative Adjective}
\label{sec:rentai}

Demonstrative pronouns 
such as ``KONO (this)'', 
``SONO (the)'', 
``ANO (that)'', 
``KON'NA (like this)'', 
and ``SON'NA (like it)'' 
are classified into two reference categories: 
{\it gentei}-reference and {\it daikou}-reference. 

In a {\it Gentei}-reference 
although a demonstrative adjective does not refer to an entity by itself, 
the phrase of ``demonstrative adjective $+$ noun phrase'' 
refers to the antecedent. 
For example 
``KONO OJIISAN (this old man)'' in the following sentences: 
\begin{equation}
  \begin{minipage}[h]{11.5cm}
    \small
  \begin{tabular}[t]{l@{ }l@{ }l@{ }l}
{\footnotesize OJIISAN-WA} & {\footnotesize TENGUTACHI-NO-MAENI} & {\footnotesize DETEITTE} & {\footnotesize ODORI-HAJIMEMASHITA}\\
 (old man) & (before the ``tengu'') & (appear) & (begin to dance)\\
\multicolumn{4}{l}{
(He appeared before the ``tengu,'' 
and began to dance.)}\\
\end{tabular}

\vspace{0.3cm}

  \begin{tabular}[t]{l@{ }l@{ }l@{ }l@{ }l}
{\footnotesize KEREDOMO} & {\footnotesize \underline{KONO OJIISAN-WA}} & {\footnotesize UTA-MO} & {\footnotesize ODORI-MO} & {\footnotesize HETAKUSO-DESHITA}\\
 (but) & (this old man) & (sing) & (dance) & (poor)\\
\multicolumn{5}{l}{
(But \underline{the old man} was a poor singer, 
and his dancing was no better. )}\\
\end{tabular}
  \end{minipage}
\label{eqn:kono_ojiisan_heta}
\end{equation}
In this example, 
although the demonstrative 
``KONO (this)'' does not refer to ``OJIISAN (old man)'' in the first sentence, 
the noun phrase 
``KONO OJIISAN (this old man)'' refers to ``OJIISAN (old man)'' 
in the first sentence. 

{\it Daikou}-reference is 
a demonstrative adjective that refers to an entity. 
In this case, we can analyze 
``{SONO} (the)'' as well as ``{SORE-NO} (of it)''. 
In the following sentences, 
``{SONO}'' refers to ``{TENGU}''. 
It is the case of {\it daikou}-reference. 
\begin{equation}
  \begin{minipage}[h]{11.5cm}
    \small
  \begin{tabular}[t]{lllll}
{\footnotesize MATA} & {\footnotesize KARASU-NO-YOUNA} & {\footnotesize KAO-WO-SHITA} & {\footnotesize TENGU-MO} & {\footnotesize IMASHITA}\\
 (also) & (like crows) & (with face) & (``tengu'') & (exist)\\
\multicolumn{5}{l}{
(There were also some ``tengu'' with faces like those of crows. )}\\
\end{tabular}

\vspace{0.3cm}

  \begin{tabular}[t]{l@{ }l@{ }l}
{\footnotesize \underline{SONO KUCHI}-WA} & {\footnotesize TORINO-KUCHIBASHI-NOYOUNI} & {\footnotesize TOGATTE-IMASHITA}\\
 (their mouths) & (like the beaks of birds) & (be pointed) \\
\multicolumn{3}{l}{
(\underline{Their mouths} were pointed like the beaks of birds. )}\\
\end{tabular}
  \end{minipage}
\label{eqn:sono_kuti}
\end{equation}

Rules for {\it gentei}-reference and {\it daikou}-reference are as follows:  

\subsection*{\underline{Rule for {\it Gentei}-Reference}}

\noindent
{\bf {\it Candidate enumerating rule}6}
\begin{indention}{0.8cm}\noindent
  When a pronoun is ``{\it so}-series demonstrative adjective + noun 
  $\alpha$,''\\
  \{
  (the noun phrase containing a noun $\alpha$, \,$45$)\\
  (the topic which is a subordinate of the noun $\alpha$ 
  and which has the weight $W$ and the distance $D$, \,$W-D*2+10$)\\
  (the focus which is a subordinate of the noun $\alpha$ 
  and which has the weight $W$ and the distance $D$, \,$W-D*2+10$)\}

  The definition and the weight ($W$) of 
  topic and focus are shown in 
  Table \ref{tab:5c_topic} and Table \ref{tab:5c_focus}. 

  When a possible referent is a topic, 
  the distance ($D$) between the estimated noun phrase and 
  the possible referent is the number of topics between them. 
  When a possible referent is a focus, 
  the distance ($D$) is the number of foci between them. 

  The relations between a super-ordinate word and a subordinate word 
  is detected by 
  the last word in the definition of the word $\alpha$ in EDR Japanese word dictionary\cite{edr_tango_2.1} 
  is judged to be the super-ordinate of the word $\alpha$ \cite{tsurumaru91}. 
  
  Since a {\it so}-series demonstrative refers to 
  noun phrases nearer than a {\it ko}-series demonstrative, 
  we give the coefficient $2$ in the second term. 
\end{indention}

\vspace{0.5cm}
\noindent
{\bf {\it Candidate enumerating rule}7}
\begin{indention}{0.8cm}\noindent
  When a pronoun is ``{\it ko}-series demonstrative adjective + noun 
  $\alpha$,''\\
  \{
  (the noun phrase containing a noun $\alpha$, \,$45$)\\
  (the topic which is a subordinate of the noun $\alpha$ 
  and which has the weight $W$ and the distance $D$, \,$W-D+30$)\\
  (the focus which is a subordinate of the noun $\alpha$ 
  and which has the weight $W$ and the distance $D$, \,$W-D+30$)\}
\end{indention}

\vspace{0.5cm}
\noindent
{\bf {\it Candidate enumerating rule}8}
\begin{indention}{0.8cm}\noindent
  When a pronoun is ``{\it a}-series demonstrative adjective + noun 
  $\alpha$,''\\
  \{
  (the noun phrase containing a noun $\alpha$, \,$45$)\\
  (the topic which is a subordinate of the noun $\alpha$ 
  and which has the weight $W$ and the distance $D$, \,$W-D*0.4+30$)\\
  (the focus which is a subordinate of the noun $\alpha$ 
  and which has the weight $W$ and the distance $D$, \,$W-D*0.4+30$)\}
\end{indention}
\vspace{0.5cm}

Because of the above three rules, 
when 
a pronoun is ``demonstrative adjective $+$ noun phrase $\alpha$'' 
and there is the same noun phrase $\alpha$ near it, 
it is judged to be ``{\it gentei}-reference'' 
and is selected as a candidate of the referent. 
When there is a subordinate of a noun phrase $\alpha$ near it, 
it is also selected as a candidate of the referent. 
These rules give higher points 
to a candidate referent than in the other rules. 
The following is an 
example of 
the ``demonstrative adjective $+$ noun phrase $\alpha$'' 
referring to the subordinate of the noun phrase $\alpha$. 
\begin{equation}
  \begin{minipage}[h]{11.5cm}
\small
  \begin{tabular}[t]{l@{ }l@{ }l@{ }l@{ }l}
OJIISAN-WA & TOONOITEIKU & TSURU-NO & SUGATA-WO & MIOKURIMASHITA.\\
 (old man) & (recede) & (crane) & (figure) & (watch)\\
\multicolumn{5}{l}{
(The old man watched the receding figure of the crane. )}\\
\end{tabular}

\vspace{0.3cm}

  \begin{tabular}[t]{llll}
``\underline{ANO TORI}-WO & TASUKETE & YOKATTA'' TO & IIMASHITA.\\
 (that bird) & (save) & (glad) & (say)\\
\multicolumn{4}{l}{
(``I'm glad I saved \underline{that bird},'' said the old man to himself. )}\\
\end{tabular}
  \end{minipage}
\label{eqn:ano_tori}
\end{equation}
In this example, 
the underlined ``{ANO TORI} (that bird)'' refers to 
a subordinate ``{TSURU} (crane)'' in the previous sentence. 

\begin{table}[t]
\vspace*{-0.8cm}
  \leavevmode
    \caption{Points given to so-series demonstrative adjective}
    \label{tab:sokei_meishi_anob_ruijido}

  \begin{center}
\begin{tabular}[c]{|l|r|r|r|r|r|r|r|r|}\hline
Similarity Level & 0   & 1  & 2  & 3 & 4 & 5 & 6 & Exact Match\\\hline
Point   & $-$10 & $-$2 & $-$1 & 0 & 1 & 2 & 3 & 4\\\hline
\end{tabular}
\end{center}
\end{table}

\subsection*{\underline{Rules for 
{\it Daikou}-Reference of So-Series Demonstrative Adjective}}

\noindent
{\bf {\it Candidate judging rule}5}
\begin{indention}{0.8cm}\noindent
  When a pronoun is a {\it so}-series demonstrative adjective, 
  the system consults examples of 
  the form ``noun X {NO} noun Y'' 
  whose noun Y is modified by the pronoun, 
  and gives a candidate referent 
  the point
  in Table~\ref{tab:sokei_meishi_anob_ruijido}
  by the similarity between 
  the candidate referent and noun X 
  in ``Bunrui Goi Hyou''\cite{BGH}. 
  The Japanese Co-occurrence Dictionary\cite{edr_kyouki_2.1} is used 
  as a source of examples of ``X NO Y''. 
\end{indention}
\vspace{0.5cm}

This rule is for checking 
the semantic constraint 
(For a {\it daikou}-reference, candidates of the referent 
are selected by {\it Candidate enumerating rule}1 in Section \ref{sec:meishi_siji}.).

We explain how to use the rule 
in the underlined ``{SONO} (the)'' 
in the sentences (\ref{eqn:sono_kuti}). 
First, the system gathers 
examples of the form ``Noun X {NO KUCHI} ( mouth of Noun X )''. 
Table \ref{fig:meishi_A_kuti} shows 
some examples of ``Noun X {NO KUCHI} ( mouth of Noun X )'' in the Japanese Co-occurrence Dictionary\cite{edr_kyouki_2.1}. 
Next, the system checks the semantic similarity 
between candidate referents and Noun X, and 
judges that the candidate referent which has a higher similarity 
is a better candidate referent. 
In this example, 
``{TENGU}'' is semantically similar to Noun X 
in that they are living things. 
At last, the system selects 
``{TENGU}'' as the proper referent. 

\begin{table}[t]
  \caption{Examples of the form ``the mouth of Noun X'' }
  \label{fig:meishi_A_kuti}

\begin{center}
\begin{tabular}[c]{|p{11.5cm}|}\hline
Examples of Noun X \\\hline
{HUKURO} (sack),  {RUPORAIT\=A}(documentary writer) 
{IIN}(member), {AKACHAN}(baby), {KARE}(he)\\\hline
\end{tabular}
\end{center}
\end{table}

\begin{table}[t]
\vspace*{-0.8cm}
  \leavevmode
  \caption{Points given in the case of non-{\it so}-series demonstrative adjective}
    \label{tab:akei_meishi_anob_ruijido}

  \begin{center}
\begin{tabular}[c]{|l|@{\hspace{0.12cm}}r@{\hspace{0.12cm}}|@{\hspace{0.12cm}}r@{\hspace{0.12cm}}|@{\hspace{0.12cm}}r@{\hspace{0.12cm}}|@{\hspace{0.12cm}}r@{\hspace{0.12cm}}|@{\hspace{0.12cm}}r@{\hspace{0.12cm}}|@{\hspace{0.12cm}}r@{\hspace{0.12cm}}|@{\hspace{0.12cm}}r@{\hspace{0.12cm}}|@{\hspace{0.12cm}}r|}\hline

Similarity Level & 0   & 1   & 2   & 3   & 4   & 5 & 6 & Exact match\\\hline
Point   & $-$30 & $-$30 & $-$30 & $-$30 & $-$10 & $-$5& $-$2& 0\\\hline
\end{tabular}
\end{center}
\end{table}

\subsection*{\underline{Rules when 
Non-{\it So}-Series Demonstrative has 
{\it Daikou}-Reference}}

\noindent
{\bf {\it Candidate judging rule}6}
\begin{indention}{0.8cm}\noindent
  When a pronoun is a non-{\it so}-series demonstrative adjective, 
  the system consults examples of 
  the form ``Noun X {NO(of)} Noun Y (Y of X)'' 
  whose Noun Y is modified by the pronoun, 
  and gives candidate referents 
  the point
  in Table \ref{tab:akei_meishi_anob_ruijido}
  by the similarity between 
  the candidate referent and noun X 
  in ``Bunrui Goi Hyou''\cite{BGH}. 
  Since a non-{\it so}-series demonstrative adjective 
  rarely is a {\it daikou} reference 
  \cite{seiho1}
  \cite{yamamura92_ieice}, 
  the point is lower than that in the case of {\it so}-series. 
\end{indention}

\subsection*{\underline{Rule when a Pronoun Refers to a Verb Phrase}}

As in a demonstrative pronoun, 
a demonstrative adjective can refer to the meaning 
of the verb phrase in the previous sentence 
\footnote{
It is necessary 
to distinguish between 
{\it daikou}-reference and {\it gentei}-reference  
even in the case when a pronoun refers to a verb phrase. 
But, in this thesis, 
we do not distinguish them because of the difficulty of the problem. 
}. 
\begin{quote}
\small
TSUMARI, NINGEN-NO NOU-YORI YUUSHUUNA PATAAN NINSHIKI PUROGURAMU-GA 
TSUKURENAI DANKAI-DEWA, HIJOUNI HUKUZATSUDE OMOSHIROSOUNA JISHOU-NITSUITEWA, 
MAZU SONO GAZOU WO TSUKUTTE, SONO DEETA-WO BUTSURIGAKUSHA-NI 
GINMI-SASERU HITSUYOU-GA-ARU. \\
(Until scientists invent a pattern recognition program that works better than the human brain, it will be necessary to produce images of the most complicated and interesting events so that physicists can scrutinize the data.)\\
1980 NEN DAI-NO SHOTOU-NI LEP JIKKEN SOUCHI-NO SEKKEI-GA HAJIMATTA-TOKI, 
\underline{KONO SENRYAKU}-GA SAIYOU SARETANODATTA.\\
(\underline{This strategy} was adopted by workers when they began to design the LEP detectors in the early 1980s.)
\end{quote}
%
The referent of ``{KONO SENRYAKU} (this strategy)'' is 
the meaning of the previous sentence. 
The resolution in this case 
is performed as follows: 
When there are no noun phrases 
which are suitable for 
the referent of ``{KONO} (this)'' or 
the referent of ``{KONO SENRYAKU} (this strategy)'' 
near the demonstrative, 
the system judges that 
the meaning of the previous sentence 
is the proper referent, 
provided that, as in a demonstrative pronoun 
when the verb phrase 
containing a conjunctive particle such as
  ``GA'', ``DAGA'', and ``KEREDO'' 
or  a conditional form
exists in the same sentence, 
the verb phrase is judged to be the proper referent. 
The above procedure is done by {\it Candidate enumerating rule}2 
in Section \ref{sec:meishi_siji}. 

\subsection*{\underline{Rule for ``KON'NA $+$ Noun (noun like this)''}}

``KON'NA Noun'' can also refer to the next sentences 
in addition to a noun phrase and the previous sentences. 
\begin{equation}
  \begin{minipage}[h]{11.5cm}
  \begin{tabular}[t]{llll}
OJIISAN-WA & ODORINAGARA & \underline{KON'NA UTA}-WO & UTAIMASHITA.\\
 (old man) & (dance) & (song like this) & (sing)\\
\multicolumn{4}{l}{
(As he danced, he sang \underline{the following song}: )}\\
\end{tabular}

\vspace{0.3cm}

  \begin{tabular}[t]{llll}
``TENGU & TENGU & HACHI TENGU.\\
 (tengu) & (tengu) & (eight tengu)\\
\multicolumn{4}{l}{
(```Tengu,' `tengu,' Eight `tengu.''')}\\
\end{tabular}
  \end{minipage}
\label{eqn:konana_kouhou}
\end{equation}
In the above example, 
``KON'NA UTA (song like this)'' refers to the next sentence 
``TENGU, TENGU, HACHI TENGU.''

\begin{table}[t]
  \caption{The result of the investigation 
whether ``KON'NA $+$ noun (noun like this)'' refers to the previous sentences 
or the next sentences}
  \label{fig:konna_meishi_joshi_tyosha}
\small
\begin{center}
\begin{tabular}[c]{|l|r|r|}\hline
  \multicolumn{1}{|l|}{Postpositional particle}  & \multicolumn{1}{|l|}{the previous sentence} & \multicolumn{1}{|l|}{the next sentence}\\\hline
WA (topic) & 9 & 0\\\hline
WA-NAI  & 5 & 0\\\hline
NI (indirect object) & 17 & 0\\\hline
NI-MO & 1 & 0\\\hline
NI-WA & 2 & 0\\\hline
DE (place)   & 15 & 0\\\hline
DE-WA & 5 & 0\\\hline
NO (possessive)   & 9 & 0\\\hline
SURA   & 2 & 0\\\hline
GA (subject)   & 27 & 22\\\hline
WO (object)   & 43 & 26\\\hline
MO (also)  & 2 & 4\\\hline
DE-WA-NAI & 0 &1\\\hline
Total & 137& 53\\\hline
\end{tabular}
\end{center}
\end{table}

But 
we cannot decide whether ``KON'NA $+$ noun (noun like this)'' 
refers to the previous sentences or the next sentences 
only by the expression of 
``KON'NA $+$ noun (noun like this)'' itself. 
To make the decision, 
we gather 317 sentences containing ``KON'NA (like this)'' 
from about 60,000 sentences in 
TENSEIJINGO and editorials (1986 and 1987), 
and count the total frequency that 
``KON'NA''  refers to the previous sentences 
or to the next sentences. 
The result is shown in Table \ref{fig:konna_meishi_joshi_tyosha}. 
This table indicates that 
``KON'NA $+$ noun'' followed by the other particles of the particles 
``{GA}'' and ``{WO},'' 
which are used when representing new information, 
very often refers to the previous sentence. 
Therefore, the system judges that 
the desired antecedent is the previous sentence. 
When ``KON'NA $+$ noun'' followed by the particles 
``{GA}'' and ``{WO},'' 
the proper referent is determined 
by the expression of quotation marks (``,'') 
as well as Matsuoka's method \cite{matsuoka_nl}. 

\subsection{Rule for Demonstrative Adverb}

\subsection*{\underline{Rule when 
{\it So}-Series Demonstrative Adverb Refers to}\\ 
\underline{the Previous Sentences}}

\noindent
{\bf {\it Candidate enumerating rule}9}
\begin{indention}{0.8cm}\noindent
  When an anaphor is 
  a {\it so}-series demonstrative adverb such as ``{SOU (so)},''\\ 
  \{(the previous sentences, \,$30$)\}
\end{indention}

The example is as follows.
\begin{equation}
  \begin{minipage}[h]{11.5cm}
\small
  \begin{tabular}[t]{llll}
``{\footnotesize TENGU} & {\footnotesize TENGU} & {\footnotesize HACHI TENGU}.\\
 (tengu) & (tengu) & (eight tengu)\\
\multicolumn{4}{l}{
(```Tengu,' `tengu,' Eight `tengu.''')}\\
\end{tabular}

\vspace{0.3cm}

  \begin{tabular}[t]{l@{ }l@{ }l@{ }l@{ }l}
{\footnotesize \underline{SOU} UTATTA-NOWA} & {\footnotesize SOKONI} & {\footnotesize HACHIHIKI-NO} & {\footnotesize TENGU-GA} & {\footnotesize ITAKARA-DESU}.\\
 (sing so) & (there) & (eight) & (tengu) & (exist)\\
\multicolumn{5}{l}{
(He sang \underline{so} because he counted eight of them there. )}\\
\end{tabular}
  \end{minipage}
\label{eqn:sono_utau}
\end{equation}
``{SOU} (so)'' refers to the previous sentence 
``TENGU TENGU HACHI TENGU''.

\subsection*{\underline{Rule when 
{\it So}-Series Demonstrative Adverb Cataphorically}\\ 
\underline{Refers to 
the Verb Phrase in the Same Sentence}}

\noindent
{\bf {\it Candidate enumerating rule}10}
\begin{indention}{0.8cm}\noindent
  When an anaphor is ``SOU/SOUSHITE/SONOYOUNI'' 
  and
  is in the subordinate clause 
  which has a conjunctive particle such as
  ``GA'', `` DAGA'', and `` KEREDO'' 
  or an adjective conjunction such as ``YOUNI'',\\
  \{(the main clause, \,$45$)\}\\
  This rule is based on Matsuoka's method \cite{matsuoka_nl}. 
\end{indention}

\subsection*{\underline{Rule when
{\it Ko}-Series Demonstrative Adverb Refers to}\\ 
\underline{the Previous Sentences}}

\noindent
{\bf {\it Candidate enumerating rule}11}
\begin{indention}{0.8cm}\noindent
  When an anaphor is a {\it ko}-series demonstrative adverb 
  such as ``KOU (in this way)'',\\
  \{(the previous sentences, \,$25$)\}
\end{indention}
\vspace{0.5cm}

\subsection*{\underline{Rule when 
{\it Ko}-Series Demonstrative Adverb Refers to the Next}\\ 
\underline{Sentences}}

\noindent
{\bf {\it Candidate enumerating rule}12}
\begin{indention}{0.8cm}\noindent
  When an anaphor is a {\it ko}-series demonstrative adverb, \\
  \{(the next sentences, \,$26$)\}
\end{indention}
\vspace{0.5cm}

A {\it ko}-series demonstrative adverb can also refer to the next sentences 
in addition to the previous sentences. 
\begin{equation}
  \begin{minipage}[h]{11.5cm}
  \begin{tabular}[t]{llll}
TENGU-TACHI-WA & TOUTOU & \underline{KOU} & IIMASHITA.\\
 (tengu) & (finally) & (like this) &(say)\\
\multicolumn{4}{l}{
(The ``tengu''  finally said \underline{as follows}:)}\\
\end{tabular}

\vspace{0.3cm}

  \begin{tabular}[t]{llll}
KYOU-NO & OMAE-WA & DAME-DANA. ...\\
(today) & (you) & (no good)\\
\multicolumn{3}{l}{
(``You're no good today. ...'')}\\
\end{tabular}
  \end{minipage}
\label{eqn:kou_iu}
\end{equation}
In the example, 
``KOU (in this way)'' refers to the next sentences. 
When ``{KOU} (in this way)'' is a part of 
the typical form such as 
``{KOU SHITE}'' and ``{KOU SUREBA},'' 
it often refers to the previous sentences. 
Therefore 
if ``{KOU} (in this way)'' is a part of 
this typical form, 
the system judges that 
the desired antecedent is the previous sentence. 
Otherwise, 
the system judges that 
the desired antecedent is the next sentence. 
To implement this procedure, we made the following rules. 

\vspace{0.5cm}
\noindent
{\bf {\it Candidate enumerating rule}13}
\begin{indention}{0.8cm}\noindent
When an anaphor is a part of 
  ``KOU/KON'NAHUUNI'' + conditional form or ``KOU SHITE'' 
  and is not the last word in the sentence,\\
  \{(the previous sentence, \,$7$)\}
\end{indention}

\section{Heuristic Rule for Personal Pronoun}
\label{sec:pro_ana}

\noindent
{\bf {\it Candidate enumerating rule}1}
\begin{indention}{0.8cm}\noindent
  When an anaphor is a first personal pronoun, \\
  \{(the first person (the speaker) in the context, \,$25$)\}
\end{indention}

\vspace{0.5cm}
\noindent
{\bf {\it Candidate enumerating rule}2}
\begin{indention}{0.8cm}\noindent
  When an anaphor is a second personal pronoun, \\
  \{(the second person (the hearer) in the context, \,$25$)\}
\end{indention}
\vspace{0.5cm}

A first or second personal pronoun is often presented 
in quotation, 
and can be resolved 
by estimating 
the first person (speaker) or the second person (hearer) 
in advance. 
The estimation of 
the first person and the second person is performed 
by regarding {\it ga}-case component and {\it ni}-case component 
of the verb phase which represents the speaking action of the quotation 
as the first person and the second person, respectively. 
The detection of the verb phase representing the speaking action 
is performed as follows. 
If the quotation is followed by a speaking action verb phrase 
such as ``{TO ITTA} (was said),'' 
the verb phrase is regarded as the verb phase 
representing the speaking action. 
Otherwise, the last verb phrase in the previous sentence is 
regarded as the verb phase 
representing the speaking action
\footnote{
There are some errors in the detection 
of the verb phrase representing the speaking action in this method. 
But in the sample texts used in the experiment of this thesis, 
all detection could be performed properly in this method. 
}. 
For example, 
the second personal pronoun ``{OMAESAN} (you)'' in the following sentences 
refers to the second person ``{OJIISAN} (the old man)'' 
in this quotation. 
\begin{equation}
  \begin{minipage}[h]{11.5cm}
    \small
  \begin{tabular}[t]{lll}
``ASU, & MATA & MAIRIMASUYO.'' TO,\\
 (tomorrow) & (again) & (come) \\
\multicolumn{3}{l}{
(``I'll come again tomorrow,'')}\\
\end{tabular}

\vspace{0.3cm}

  \begin{tabular}[t]{lll}
OJIISAN-WA & YAKUSOKU-SHIMASHITA.\\
 (old man) & (promise) \\
\multicolumn{3}{l}{
(promised the old man.)}\\
\end{tabular}

\vspace{0.3cm}

  \begin{tabular}[t]{lll}
``MOCHIRON & \underline{OMAESAN}-WO & UTAGAUWAKEDEWANAINODAGA,''\\
 (of course) & (you) & (don't mean to doubt) \\
\multicolumn{3}{l}{
(``Of course, we don't mean to doubt \underline{you},'')}\\
\end{tabular}

\vspace{0.3cm}

  \begin{tabular}[t]{lll}
TENGU-GA & \underline{OJIISAN-NI} & IIMASHITA.\\
 (tengu) & (old man) & (said) \\
\multicolumn{3}{l}{
(said one of the ``tengu'' to \underline{the old man}.)}\\
\end{tabular}
  \end{minipage}
\label{eqn:ojiisan_mairu_omae}
\end{equation}
The fact that 
the second person in the quotation is ``{OJIISAN}'' is 
estimated by 
the fact that 
{\it ni}-case component of the verb phrase ``{IIMASHITA} (said)'' 
representing the speaking action of the quotation is 
``{OJIISAN}''. 

\vspace{0.5cm}
\noindent
{\bf {\it Candidate enumerating rule}3}
\begin{indention}{0.8cm}\noindent
  When an anaphor is a third personal pronoun, \\
  \{(a first person, \,$-10$) (a second person, \,$-10$)\}
\end{indention}
\vspace{0.5cm}

Personal pronouns are generally analyzed by the following three rules: 
The system lists candidate referents with the scores 
(the certification value) considering 
topic/focus and the distance between the anaphor and the candidate referents 
by {\it Candidate enumerating rule}4, 
and increases the score of the candidate referents which signify human beings 
by {\it Candidate judging rule}1 and {\it Candidate judging rule}2. 

\begin{table}[t]
\vspace*{-0.8cm}
  \leavevmode
    \caption{Points given in the case of personal pronoun}
    \label{tab:ninshoudaimeisi_ruijido}

  \begin{center}
\begin{tabular}[c]{|l|r|r|r|r|r|r|r|r|}\hline
Similarity Level & 0 & 1 & 2 & 3 & 4 & 5 & 6 & Exact Match\\\hline
Point   & 0 & 0 & 3 & 7 & 10& 10& 10& 10\\\hline
\end{tabular}
\end{center}
\end{table}

\vspace{0.5cm}
\noindent
{\bf {\it Candidate enumerating rule}4}
\begin{indention}{0.8cm}\noindent
  When an anaphor is a personal pronoun,\\ 
  \{(A topic which has the weight $W$ and the distance $D$,
\,$W-D-2$)\\
  (A focus which has the weight $W$ and the distance $D$, \,$W-D+4$)\}
\end{indention}

\vspace{0.5cm}
\noindent
{\bf {\it Candidate judging rule}1}
\begin{indention}{0.8cm}\noindent
  When an anaphor is a personal pronoun 
  and a candidate referent has 
  a semantic marker {\sf HUM}, 
  the candidate referent is given $10$ points. 
\end{indention}

\vspace{0.5cm}
\noindent
{\bf {\it Candidate judging rule}2}
\begin{indention}{0.8cm}\noindent
  When an anaphor is a personal pronoun, 
  a candidate referent is given the points 
  in Table \ref{tab:ninshoudaimeisi_ruijido}
  by using the highest semantic similarity 
  between the candidate referent and 
  the code 
  \{5200003010 5201002060 5202001020 5202006115 5241002150 5244002100\}
  which signifies human being in BGH\cite{BGH}. 
\end{indention}

\section{Heuristic Rule for Zero Pronoun}
\label{sec:zero_ana}

\subsection*{\underline{Rule Proposing Candidate Referents 
of General Zero Pronoun}}

\noindent
{\bf {\it Candidate enumerating rule}1}
\begin{indention}{0.8cm}\noindent
  When a zero pronoun is a {\it ga}-case component, \\
  \{(A topic which has the weight $W$ and the distance $D$, \,$W-D*2+1$)\\
  (A focus which has the weight $W$ and the distance $D$, \,$W-D+1$)\\
  (A subject of a clause coordinately connected to the clause containing the anaphor, \,25)\\
  (A subject of a clause subordinately connected to the clause containing the anaphor, \,23)\\
  (A subject of a main clause whose 
  embedded clause contains the anaphor, \,22)\}\\
\end{indention}

\vspace{0.5cm}
\noindent
{\bf {\it Candidate enumerating rule}2}
\begin{indention}{0.8cm}\noindent
  When a zero pronoun is not a {\it ga}-case component, \\
  \{(A topic which has the weight $W$ and the distance $D$, \,$W-D*2-3$)\\
  (A focus which has the weight $W$ and the distance $D$, \,$W-D*2+1$)\}
\end{indention}
\vspace{0.5cm}

\subsection*{\underline{Rule for Analyzing Complex Sentences}}

\noindent
{\bf {\it Candidate enumerating rule}3}
\begin{indention}{0.8cm}\noindent
  When a zero pronoun is {\it ga}-case of the main (or subordinate) clause 
  in a complex sentence, 
  the complex sentence is connected 
  by the conjunctive particle indicating 
  the disagreement of the subjects in a complex sentence 
  such as ``{NODE} (because)'' and ``{NARABA} (if)'' 
  and the subject of the subordinate (or main) clause is not omitted 
  and is followed by the particle ``{GA},''\\
  \{(the subject of the subordinate (or main) clause, \,$-30$)\}
\end{indention}
\vspace{0.5cm}

For a {\it ga}-case zero pronoun 
of the main (or subordinate) clause in a complex sentence, 
if there is a {\it ga}-case noun phrase 
in the subordinate (or main) clause, 
the system commonly judges that 
the {\it ga}-case noun phrase 
is the antecedent of the {\it ga}-case zero pronoun. 
But it is known that 
there are conjunctive particles 
which produce 
disagreement of subjects in a complex sentence 
\cite{minami} \cite{yoshimoto} \cite{hirai} \cite{nakaiwa}. 
When a complex sentence is connected by these conjunctive particles, 
the system does not judge that 
the noun phrase of the subordinate (or main) clause 
is the desired antecedent. 
{\it Candidate enumerating rule}3 is for this procedure. 

\subsection*{\underline{Rule Using Semantic Relation 
to Verb Phrase}}

\noindent
{\bf {\it Candidate judging rule}1}
\begin{indention}{0.8cm}\noindent
  When a candidate referent of a case component (a zero pronoun) 
  does not satisfy the semantic marker of the case component 
  in the case frame, 
  it is given $-5$. 
\end{indention}

\vspace{0.5cm}
\noindent
{\bf {\it Candidate judging rule}2}
\begin{indention}{0.8cm}\noindent
  A candidate referent of a case component ( a zero pronoun ) is 
  given points in Table \ref{tab:yourei_ruijido}
  by using the highest semantic similarity 
  between the candidate referent and examples 
  of the case component in the case frame. 
\end{indention}
\vspace{0.5cm}

\begin{table}[t]
\vspace*{-0.8cm}
  \leavevmode
    \caption{Points given from a verb-noun relationship}
    \label{tab:yourei_ruijido}

  \begin{center}
\begin{tabular}[c]{|l|r|r|r|r|r|r|r|r|}\hline
Similarity Level & 0 & 1 & 2 & 3 & 4 & 5 & 6 & Exact Match\\\hline
Point   & $-$10 & $-$2 & 1 & 2 & 2.5& 3 & 3.5 & 4\\\hline
\end{tabular}
\end{center}
\end{table}

These two rules are 
for checking 
the semantic constraint 
between the candidate referent and 
the verb phrase which has the candidate referent in its case component. 
{\it Candidate judging rule}1 
checks semantic constraints by using semantic markers. 
{\it Candidate judging rule}2 
checks semantic constraints by using examples. 
We explain how to check 
semantic constraints in the example sentences 
in Figure \ref{fig:datousei_hantei_rei}. 
\begin{figure}[t]
  \leavevmode
  \begin{center}
\fbox{
    \begin{minipage}[c]{13cm}
    \begin{tabular}[t]{lll}
\underline{OJIISAN}-WA & JIMEN-NI & KOSHI-WO-OROSHIMASHITA.\\
(old man) & (ground) & (sit down)\\
\multicolumn{3}{l}{
(\underline{The old man} sat down on the ground.)}\\[0.3cm]
\end{tabular}
\begin{tabular}[t]{lll}
YAGATE & (\underline{OJIISAN}-WA) & NEMUTTE-SHIMAIMASHITA.\\
(soon) & (old man) & (fall asleep)\\
\multicolumn{3}{l}{
(\underline{He} soon fell asleep.)}\\
\end{tabular}\\

\begin{tabular}[c]{l@{ }l@{ }l}
{\bf Semantic Marker} & HUM/ANI GA(agent) &NEMURU (sleep)\\
{\bf Example} &    KARE (he)/ INU (dog) GA(agent) &NEMURU (sleep)\\
\end{tabular}
    \caption{Example of how to check semantic constraint}
    \label{fig:datousei_hantei_rei}
    \end{minipage}
}
  \end{center}
\end{figure}

In the method using semantic markers, 
a candidate referent is the proper referent 
if one of the semantic markers which the candidate referent has 
is equal or subordinate to the semantic marker of the case component. 
For example, 
with respect to the zero pronoun in Figure \ref{fig:datousei_hantei_rei}, 
since the {\it ga}-case component 
in the verb ``{NEMURU} (sleep)'' 
has the semantic markers {\sf HUM} (human being) and {\sf ANI} (animal)
\footnote{
{\sf HUM} and {\sf ANI} are the semantic markers which indicate 
human being (\underline{HUM}AN) 
and animal (\underline{ANI}MAL), respectively. 
}, 
and ``OJIISAN (old man)'' has the semantic marker {\sf HUM}, 
``OJIISAN'' is judged to be the proper referent. 

In the example-based method, 
the validity of a candidate referent 
is decided by 
the semantic similarity 
between the candidate referent and 
the examples of the case component in the verb case frame. 
The higher the semantic similarity is, 
the higher the validity is. 
For example, 
with respect to a zero pronoun in Figure \ref{fig:datousei_hantei_rei}, 
since the examples of {\it ga}-case are 
``{KARE} (he)'' and ``{INU} (dog)''  
and ``OJIISAN (old man)'' is semantically similar to ``{KARE} (he)'', 
``OJIISAN (old man)'' is the proper referent. 

These rules, which use semantic relations to verbs,  
are also used in the estimation of the referent of 
demonstratives and  personal pronouns. 

\subsection*{\underline{
Rule Using the Feature that 
it is Difficult for a Noun Phrase 
to}\\ 
\underline{be Filled 
in Plural Case Components of the Same Verb}}

\noindent
{\bf {\it Candidate enumerating rule}4}
\begin{indention}{0.8cm}\noindent
  When there is ``Noun X'' in another case component 
  of the verb which has the analyzed case component 
  (the analyzed zero pronoun), 
  \{(Noun X, \,$-20$)\}
\end{indention}

\subsection*{\underline{Rule Using Empathy}}

\noindent
{\bf {\it Candidate enumerating rule}5}
\begin{indention}{0.8cm}\noindent
  When an anaphor is a {\it ga}-case zero pronoun 
  whose verb is followed by the auxiliary verbs 
  such as ``{KURERU}'' and ``{KUDASARU}'' 
  and there is a {\it ni}-case zero pronoun in the verb, 
  the {\it ni}-case zero pronoun is analyzed first. 
  With respect to the {\it ga}-case zero pronoun,  
  \{(do not fill a zero pronoun, \,$-5$)\}\\
  This rule is based on empathy theory\cite{kameyama1}. 

\end{indention}

  When an anaphor is a {\it ga}-case zero pronoun 
  whose verb is followed by the auxiliary verbs 
  such as ``{KURERU}'' and ``{KUDASARU},'' 
  the {\it ni}-case zero pronoun is analyzed first, 
  and 
  it is filled with 
  the noun phrase which has high empathy such as topic, 
  and 
  a {\it ga}-case zero pronoun is filled with 
  the other noun phrase. 

\subsection*{\underline{Rule for Zero Pronoun 
in the Quotation}}

\noindent
{\bf {\it Candidate enumerating rule}6}
\begin{indention}{0.8cm}\noindent
In the quotation, when an anaphor is a {\it ga}-case zero pronoun 
which is easily filled with a first person, 
whose verb is such as ``{YARU} (give)'', ``{SHITAI} (want)'', and 
``{IKU (go)},'' 
\{(the first person, \,$5$)\}
\end{indention}

\vspace{0.5cm}
\noindent
{\bf {\it Candidate enumerating rule}7}
\begin{indention}{0.8cm}\noindent
In the quotation, when an anaphor is a {\it ga}-case zero pronoun 
which is easily filled with a second person, 
whose verb is such as ``{KURERU} (give)'', ``{NASARU} (do)'', and 
``{KURU} (come)'', 
or whose verb is in an imperative or interrogative form, 
\{(the first person, \,$-30$)(the second person, \,$25$)\}
\end{indention}

\vspace{0.5cm}
\noindent
{\bf {\it Candidate enumerating rule}8}
\begin{indention}{0.8cm}\noindent
In the quotation, when an anaphor is a {ga}-case zero pronoun,  \\
\{(the first person, \,$15$)\}
\end{indention}
\vspace{0.5cm}

A zero pronoun in a quotation can often be resolved 
by the surface expression of the last words in the sentence. 
A zero pronoun can be resolved 
by estimating 
the first person (speaker) or the second person (hearer) 
as in a personal pronoun 
\footnote{
\cite{kudou93_ieice} estimates 
the person of a zero pronoun in a conversational corpus. 
But in this work, 
quotations in the novel are dealt with, 
and it is necessary to estimate 
the speaker and the hearer of the quotation. 
}. 
For example, 
in the next quotation, 
we find that 
the first person is ``TENGU TACHI (tengu)'' 
and that the second person is ``OJIISAN (old man)'' 
by checking {\it ga}-case component and {\it ni}-case component 
of the verb ``{IU} (say),'' 
\begin{equation}
  \begin{minipage}[h]{11.5cm}
\small
  \begin{tabular}[t]{llll}
TENGU-TACHI-WA & TOUTOU & KOU & IIMASHITA.\\
 (tengu) & (finally) & (like this) &(say)\\
\multicolumn{4}{l}{
(The ``tengu''  finally said:)}\\
\end{tabular}

\vspace{0.3cm}

  \begin{tabular}[t]{lll}
``KYOU-NO & OMAE-WA & DAME-DANA.\\
(today) & (you) & (no good)\\
\multicolumn{3}{l}{
(``You're no good today.)}\\
\end{tabular}

\vspace{0.3cm}

  \begin{tabular}[t]{l@{ }l@{ }l@{ }l}
KORE-WO & [TENGU-TACHI-GA] & [OJIISAN-NI] & KAESHITE-YARU-KARA\\
(this) & (tengu) & (old man) & (give back to)\\
\multicolumn{4}{l}{
(``[We]'ll give this back [to you].)}\\
\end{tabular}

\vspace{0.3cm}

  \begin{tabular}[t]{ll}
[OJIISAN-GA] & KAETTE-SHIMAE.\\
(old man)   & (go home)\\
\multicolumn{2}{l}{
([You should] Now go home.'')}\\
\end{tabular}
  \end{minipage}
\label{eqn:ojiisan_omae_dame}
\end{equation}
The referent of the {\it ga}-case zero pronoun of the verb 
``KAESHITE YARU'' is the first person ``TENGU TACHI (`tengu's)'' 
because 
``{KAESHITE YARU}'' contains ``{YARU}.'' 
The referent of the {\it ni}-case zero pronoun of the verb 
``KAESHITE YARU'' is the second person ``OJIISAN (old man)'' 
because 
``{KAESHITE YARU}'' contains ``{YARU}.'' 
The referent of the {\it ga}-case zero pronoun of the verb 
``KAETTE SHIMAE'' is the second person ``OJIISAN (old man)'' 
because 
``{KAETTE SHIMAE}'' is the imperative sentence. 

\subsection*{\underline{The Other Rules}}

\noindent
{\bf {\it Candidate enumerating rule}9}
\begin{indention}{0.8cm}\noindent
  When an anaphor is a {\it ga}-case zero pronoun 
  of ``Y {DA} (is Y)'' in the expression 
  of ``X WO Y DA TO MINASU (consider X as Y)'', 
  \{(Noun X, \,$50$)\}
\end{indention}

\begin{figure}[t]
  \begin{center}
\fbox{
\begin{minipage}[h]{13cm}
\small

\hspace*{-0.3cm}
\begin{tabular}[t]{l@{ }l@{ }l@{ }l}
DORU SOUBA-WA & KITAI-KARA & 130-YEN-DAI-NI & {\footnotesize JOUSHOUSHITA}.\\
(dollar)   & (the expectations) & (130 yen) & (surge)\\
\multicolumn{4}{l}{
(The dollar has since rebounded to
about 130 yen because of the expectations. )}\\
\end{tabular}

\vspace{0.3cm}

\hspace*{-0.3cm}
\begin{tabular}[t]{l@{ }l@{ }l@{ }l}
\underline{KONO DORU-DAKA-WA} & {\footnotesize OUSHUU-TONO} & {\footnotesize KANKEI-WO} & {\footnotesize GIKUSHAKU-SASETEIRU.}\\
(the dollar's surge) & (Europe) & (relation) & (strain)\\
\multicolumn{4}{l}{
(\underline{The dollar's surge} is straining the relations with Europe. 
)}\\
\end{tabular}

\vspace{0.5cm}

\begin{tabular}[h]{|@{}l@{}|@{}c@{}|@{}c@{}|@{}c@{}|@{}c@{}|@{}c@{}|}\hline
Rule                   & \multicolumn{5}{c|}{The score of each candidate(points)}\\\hline
                       & {\footnotesize the previous} & new & 130 YEN & KITAI & {\footnotesize DORUSOUBA}\\[-0.1cm]
                       & sentence &  {\footnotesize individual} & (130 yen) & ({\footnotesize expectations}) & (dollar)\\\hline
{\footnotesize \it Candidate enumerating rule}2       & 15   &            &         &        &         \\[-0.1cm]
{\footnotesize \it Candidate enumerating rule}5    &      &    10      &         &        &         \\[-0.1cm]
{\footnotesize \it Candidate enumerating rule}1       &      &            &   17    &  15    &  15     \\[-0.1cm]
{\footnotesize \it Candidate judging rule}6      &      &            &  $-30$  &  $-30$ &  $-30$  \\\hline
Total Score             & 15   &    10      &  $-13$  &  $-15$ &  $-15$  \\\hline
\end{tabular}

\caption{Example of resolving demonstrative ``KONO (this)''}
\label{tab:5c_dousarei}
\end{minipage}
}
\end{center}
\end{figure}


\section{Experiment and Discussion}
\label{sec:jikken}

\subsection{Experiment}

Before pronoun resolution, 
sentences were transformed into a case structure 
by the case structure analyzer\cite{csan2_ieice} 
as in the experiments of the other chapters. 
The errors made by the structure analyzer were corrected by hand. 
We used IPAL dictionary\cite{ipal} as a verb case frame dictionary. 
We put together 
the case frames of the verb phrases which were not contained in this dictionary 
by consulting a large amount of linguistic data.

An example of resolution of the demonstrative ``KONO (this)'' 
is shown in Figure \ref{tab:5c_dousarei}. 
Figure \ref{tab:5c_dousarei} shows that 
the referent of the noun phrase ``{KONO DORUDAKA} (this dollar's surge)'' 
was properly judged to be the previous sentence. 

By {\it Candidate enumerating rule}2 in Section \ref{sec:sijisi_ana}, 
the system took a candidate ``The previous sentence'' 
and gave it 15 points. 
By {\it Candidate enumerating rule}5 in Section \ref{sec:sijisi_ana}, 
the system took a candidate 
``New individual'' and gave it 10 points. 
By {\it Candidate enumerating rule}1 in Section \ref{sec:sijisi_ana}, 
the system took three candidates,  
``130 YEN (130 yen)'', ``KITAI (expectations)'', 
and ``DORUSOUBA (dollar)'', 
and gave them 17, 15, and 15 points, respectively. 
The system applied {\it Candidate judging rule}6 to them. 
{\it Candidate judging rule}6 uses examples of ``X NO Y''. 
In this case, 
{\it Candidate judging rule}6 used examples of ``X NO DORUDAKA (the dollar's surge of X)''. 
The noun phrase X of this form ``X NO DORUDAKA'' was 
only ``SAIKIN (recently)'' in EDR occurrence dictionary. 
All three candidates, 
``130 YEN (130 yen)'', ``KITAI (expectations)'', 
and ``DORUSOUBA (dollar)'', 
were low in similarity to ``SAIKIN (recently)'' in ``Bun Rui Goihyou'', 
and were given $-30$ points by Table \ref{tab:akei_meishi_anob_ruijido}. 
Two candidate, ``The previous sentence'' and 
``New individual'' ,
are not noun phrases, 
and were not given points by {\it Candidate judging rule}6. 
As a result, 
``the previous sentence'' had the highest score and 
was judged to be the proper referent. 

\begin{table*}[t]

\fbox{
  \begin{minipage}[h]{13cm}
\small
    \caption{Result}
    \label{tab:5c_sougoukekka}

\vspace{0.4cm}

  \begin{center}
\begin{tabular}[c]{|@{ }l@{ }|r@{}c@{ }|r@{}c@{ }|r@{}c@{ }|r@{}c@{ }|}\hline
\multicolumn{1}{|p{2cm}|}{text}&
\multicolumn{2}{c@{ }|}{demonstrative}&
\multicolumn{2}{c@{ }|}{personal pronoun}&
\multicolumn{2}{c@{ }|}{zero pronoun}&
\multicolumn{2}{c@{ }|}{total score}\\\hline
Training    &  87\% &  ( 41/ 47)  & 100\% &   ( 9/ 9)   &  86\% & (177/205) &  87\% & (227/261) \\\hline
Test  &  86\% &  ( 42/ 49)  &  82\% &   ( 9/11)   &  76\% & (159/208) &  78\% & (210/268) \\\hline
\end{tabular}
\end{center}
{
The point given in each rule 
is manually adjusted by using the training sentences. \\
Training sentences \{example sentences (43 sentences), a folk tale ``KOBUTORI JIISAN''\cite{kobu} (93 sentences), an essay in ``TENSEIJINGO'' (26 sentences), an editorial (26 sentences), an article in ``Scientific American (in Japanese)''(16 sentences)\}\\
Test sentences \{a folk tale ``TSURU NO ONGAESHI''\cite{kobu} (91 sentences), two essays in ``TENSEIJINGO'' (50 sentences), an editorial (30 sentences), articles in ``Scientific American(in Japanese)'' (13 sentences)\}
}
  \end{minipage}
}
\end{table*}

\begin{table*}[t]
  \begin{center}
\small
    \caption{The detailed result of demonstrative}
    \label{tab:sijisi_kekka}

\vspace{0.4cm}

\begin{tabular}[c]{|l@{ }|r@{}c@{ }|r@{}c@{ }|r@{}c@{ }|r@{}c@{ }|}\hline
\multicolumn{1}{|p{2cm}|}{text}&
\multicolumn{2}{c|}{demonstrative }&
\multicolumn{2}{c|}{demonstrative }&
\multicolumn{2}{c|}{demonstrative }&
\multicolumn{2}{c|}{total score}\\[-0.2cm]
\multicolumn{1}{|p{2cm}|}{}&
\multicolumn{2}{c|}{pronoun}&
\multicolumn{2}{c|}{adjective}&
\multicolumn{2}{c|}{adverb}&
\multicolumn{2}{c|}{}\\\hline
Training   &  83\% &  ( 15/ 18)  &  86\% &   ( 19/ 22)   & 100\% & ( 7/ 7) &  87\% & ( 41/ 47) \\\hline
Test  &  82\% &  ( 14/ 17)  &  88\% &   ( 23/ 26)   &  83\% & ( 5/ 6) &  86\% & ( 42/ 49) \\\hline
\end{tabular}
\end{center}
\end{table*}

We show 
the result of our resolution 
of demonstratives, personal pronouns, and zero pronouns 
in Table \ref{tab:5c_sougoukekka}. 
The detailed result of demonstrative 
is shown in Table \ref{tab:sijisi_kekka}. 
When a demonstrative refers to some sentences, 
even if the scope of the referent cannot be estimated and 
a demonstrative 
can be correctly judged to be anaphoric or cataphoric, 
it is regarded as correct. 
This is because we think that 
the estimation of the scope of the referent should be analyzed 
after the analysis of the relation of the sentences 
such as {\it cause--effect} and {\it exemplification}. 
The precision rate of zero pronouns is in the case when 
the system knows whether the zero pronoun has the referent or not
in advance. 

\subsection{Discussion}

With respect to demonstratives, 
the precision rate was over 80\% even in the test sentences. 
It indicates that the rule used in this system is effective. 
But since Japanese demonstratives are classified into many kinds, 
the precision may increase 
by making more detailed rules. 
In this work 
we used the feature that 
``{KONO} (this)'' rarely functions as a {\it daikou}-reference. 
There were four cases analyzed correctly because of this rule. 

With respect to personal pronouns, 
since only first personal pronouns and second personal pronouns appeared  
in texts used by the experiment, 
almost all of the personal pronouns were resolved correctly 
by estimating the first persons and the second persons in the quotation. 
The main reason for the errors in the personal pronoun resolution 
is that 
{\it ni}-case zero pronoun was resolved incorrectly 
and the second person was estimated incorrectly. 

Reasons for the errors of the zero pronoun resolution are 
that there are errors in ``Bunrui goi hyou'', Noun Semantic Marker Dictionary, 
and Case Frame Dictionary, 
and that rules are insufficient 
although they can be improved by making new rules 
using syntax structures and auxiliary expressions. 

An example of errors necessary for 
understanding and reasoning is as follows: 
%
\begin{quote}
\small
{\footnotesize SONNA JOUKYOU NANONI, WASHINTON-DE HIRAKARERU 
SHUYOU-SENSHIN-7-KAKOKU-NO 
ZOUSHOU CHUUOU GINKOU SOUSAI KAIGI (G7) NI TSUITE 
KAKKOKU-NO TSUUKA TOUKYOKU-WA 
``OOKINA MONDAI-WA NAI-NODE KYOUDOU KOMINYUKE-WA DANAI. 
KAOAWASE CHUUSHIN-NO KAIGOU-DA''-TO, 
MARUDE KAIGI-NO IGI-WO USUMEYOU-TO-SHITEIRUYOUNA IIKATA-DA.} \\
(Despite these problems that plague the global economy, the
monetary authorities of the Group of Seven nations seem to be
trying to downplay the upcoming G-7 meeting in Washington. The
participants regard the meeting as just a "get-acquainted
session" and have decided against issuing a joint communique.)\\[0.3cm]
(...)\\[0.3cm]
(omission)\\[0.3cm]
(...)\\[0.3cm]
{\footnotesize BEI-SHINSEIKEN-WA CHIKAKU, 
ZAISEI AKAJI SAKUGEN-NO GUTAITEKI-KOKUSOU-WO GIKAI-NI
SHIMESU-YOTEI-DEARU.}\\
(The administration will shortly indicate its
specific deficit-cutting plans to Congress. )\\[0.3cm]
\underline{\footnotesize [TSUUKA TOUKYOKU GA]} 
{\footnotesize KYOUDOU KOMINYUKE-NO HAPPYOU-WO\\
 HIKAERUNOWA, 
KAWASE SHIJHO-NI KADAINA KITAI-WO ATAETAKU-NAI-TAME-DAROU.}\\
(The reason for [\underline{the monetary authorities'}] 
doing away with a joint communique this time
seems to be to avoid arousing any false hopes in the foreign
exchange market. )
\end{quote}
The {\it ga}-case of ``{HIKAERU} (do away with)'' in this example 
refers to ``KAKKOKU NO TSUUKA TOUKYOKU (the monetary authorities)''. 
But the system incorrectly judged that 
the referent was ``BEI-SHINSEIKEN (administration)''. 
To correct result, 
it is necessary to 
understand that 
the thing which does away with a joint communique 
is the monetary authorities. 

\subsection{Comparison Experiment}
\label{sec:taishojikken}

\begin{table}[t]
  \fbox{
  \begin{minipage}[h]{13cm}
  \begin{center}
    \caption{Result of comparison between semantic marker and example-base}
    \label{tab:yourei_taishou}

\vspace{0.4cm}

\begin{tabular}[c]{|@{ }r@{}c@{}|@{ }r@{}c@{}|@{ }r@{}c@{}|@{ }r@{}c@{}|@{ }r@{}c@{}|}\hline
 \multicolumn{2}{|c|}{Method 1} & 
 \multicolumn{2}{c|}{Method 2} & 
 \multicolumn{2}{c|}{Method 3} & 
 \multicolumn{2}{c|}{Method 4} & 
 \multicolumn{2}{c|}{Method 5}\\\hline 
\multicolumn{10}{|c|}{Demonstrative}\\\hline
87\% &  (41/47)  &  83\% &  (39/47)  &  87\% &  (41/47)  &  83\% &  (39/47)  &  79\% &  (37/47)  \\\hline
86\% &  (42/49)  &  88\% &  (43/49)  &  88\% &  (43/49)  &  84\% &  (41/49)  &  86\% &  (42/49)  \\\hline
\multicolumn{10}{|c|}{Personal pronoun}\\\hline
100\% &   (9/ 9)   & 100\% &   (9/ 9)  & 100\% &   (9/ 9)  & 100\% &   (9/ 9)  &  89\% &   (8/ 9)   \\\hline
 82\% &   (9/11)   &  64\% &   (7/11)  &  82\% &   (9/11)  &  55\% &   (6/11)  &  64\% &   (7/11)  \\\hline
\multicolumn{10}{|c|}{Zero pronoun}\\\hline
 86\% & (177/205)  &  83\% & (171/205) &  86\% & (176/205) &  82\% & (169/205) &  66\% & (135/205) \\\hline
 76\% & (159/208)  &  76\% & (158/208) &  79\% & (164/208) &  75\% & (155/208) &  63\% & (131/208) \\\hline
\end{tabular}
  \end{center}
{\small
Method 1 : Using both Semantic Marker and Example

Method 2 : Using Semantic Marker

Method 3 : Using Example (using modified codes of BUNRUI GOI HYOU)

Method 4 : Using Example (using original codes of BUNRUI GOI HYOU)

Method 5 : Using neither Semantic Marker nor Example

}
\end{minipage}
}
\end{table}

As we mentioned before, 
we use both the example rule 
and the semantic marker rule 
as judging rules. 
To check which rule is more effective, 
we made a comparison between 
the example method and 
the semantic marker method. 
The result is shown in Table \ref{tab:yourei_taishou}. 
The upper and lower row of this table 
show the accuracy rates for training sentences 
and test sentences, respectively. 
The rules using examples are
{\it Candidate judging rule}2,4 for demonstratives, 
{\it Candidate judging rule}2 for personal pronouns, and 
{\it Candidate judging rule}2 for zero pronouns. 
The rules using semantic markers are
{\it Candidate judging rule}1,3 for demonstratives, 
{\it Candidate judging rule}1 for personal pronouns, and 
{\it Candidate judging rule}1 for zero pronouns. 
We used the example rules of ``X NO(of) Y (Y of X)'' 
on all of these comparison experiments, 
because there are no rules using semantic markers 
which correspond to rules of ``X NO(of) Y''. 
The precision of the method using examples 
was equivalent or superior 
to the precision in the method using semantic markers 
as Table \ref{tab:yourei_taishou}. 
This indicates that 
we can use examples as well as semantic markers. 
Since some codes in BGH are incorrect, 
we modified the codes. 
Since the precision using modified codes was higher than 
using original codes, 
this indicates that 
the modification of codes is valid. 

There were some cases when 
the example method is still effective 
in the expression somewhat semantically far from those 
written in a case frame. 
For example, 
since the {\it ni}-case in the case frame of ``{IU} (say)'' 
is given only the semantic marker {\sf HUM} (human), 
the system cannot fill ``TSURU (crane bird)'' 
in the {\it ni}-case of the following example sentences 
by the semantic marker method. 
\begin{equation}
  \begin{minipage}[h]{11.5cm}
    \small
  \begin{tabular}[t]{l@{ }l@{ }l@{ }l@{ }l}
OJIISAN-WA & TSURU-WO & NIGASHI-NAGARA & [TSURU-NI] & IIMASHITA.\\
 (old man) & (crane) & (let loose) & (to crane) & (say)\\
\multicolumn{5}{l}{
(The old man let the crane loose, and said [to crane]. )}\\
\end{tabular}
  \end{minipage}
\label{eqn:turu_hanasu}
\end{equation}
But by the example method 
the system can fill ``TSURU (crane bird)'' 
in the {\it ni}-case 
because the similarity level between human beings and animals is 1 
and the subtraction of the score is low. 

\subsection{Examining Which Rules are Important}
We used many rules in this work. 
We examined the importance of various rules. 

In zero pronoun resolution, 
the information of the semantic relation between verbs and 
case components is important 
because there are few key surface expressions. 

On the contrary, 
in demonstrative resolution, 
the information of the semantic relation 
between verbs and case components is not so important 
because there are many surface expressions 
and referents limited to things 
which are not human. 
In demonstrative resolution, 
all the rules are important, 
because Japanese demonstratives are classified into many kinds 
and we must make many detailed rules. 

In first and second personal pronoun resolution, 
the rules using first persons and second persons 
were very effective. 

\section{Summary}
\label{sec:owari}

In this chapter, we presented a method of estimating 
referents of demonstrative pronouns, personal pronouns, 
and zero pronouns in Japanese sentences
using examples, surface expressions, topics and foci.
In conventional work, 
semantic markers have been used 
for semantic constraints. 
In contrast, we used examples for semantic constraints 
and showed in our experiments that 
examples are as useful as semantic markers. 
We also proposed many new methods for estimating referents of pronouns. 
For example, we use the form ``X of Y'' for 
estimating referents of demonstrative adjectives. 
In addition to our new methods, 
we used many conventional methods. 
As a result, experiments using these methods 
obtained a precision rate of 87\% 
in the estimation 
of referent of demonstrative pronouns, personal pronouns, 
and zero pronouns 
on training sentences, 
and obtained a precision rate of 78\% 
on test sentences. 


\chapter{Verb Phrase Ellipsis Resolution}
\label{chap:0verb}

\section{Introduction}
\label{sec:6c_intro}

In the previous chapters, 
we have discussed anaphora resolution 
in Japanese noun phrases and pronouns. 
The remaining problem is 
anaphora resolution in Japanese verb phrases. 
Verb phrase anaphora is classified 
into two categories: 
(i) anaphora in pro-verbs such as ``SOU SURU (do so)'' 
and (ii) the ellipsis of a verb phrase. 
In this thesis, (i) anaphora by pro-verbs is 
handled already in Chapter \ref{chap:deno} 
as demonstrative adverbs such as ``SOU (so)'' and 
``KOU (like this)''. 
This chapter 
describes (ii) 
how to resolve 
the verb phrase ellipsis. 

Verb phrases are sometimes omitted in Japanese sentences. 
It is necessary to resolve verb phrase ellipses   
for purposes of language understanding, 
machine translation, and dialogue processing. 
This chapter 
describes  a practical method to resolve omitted verb phrases 
by using surface expressions and examples. 
In short, 
(1) when the referent of a verb phrase ellipsis appears in the sentences, 
we use surface expressions (clue words); 
(2) when the referent does not appear in the sentences, 
we use examples (linguistic data). 
We define the verb phrase to which a verb phrase ellipsis refers 
as {\it the complemented verb phrase}. 
For example, 
``[KOWASHITA]\footnote{
\label{foot:6c_bracket}
A phrase in brackets ``['',``]'' represents an omitted verb phrase.}
 (broke)'' in the second sentence 
of the following example 
is a verb phrase ellipsis. 
``KOWASHITA (broke)'' in the first sentence 
is a complemented verb phrase. 
\begin{equation}
  \begin{minipage}[h]{10cm}
  \begin{tabular}[t]{lll}
KARE-WA & IRONNA MONO-WO & KOWASHITA.\\
(he)& (several things) & (broke) \\
\multicolumn{3}{l}{
(He broke several things.)}\\
\end{tabular}
  
\vspace{0.3cm}

  \begin{tabular}[t]{lll}
KORE-MO & ARE-MO & [KOWASHITA].\\
(this)  & (that) & (broke) \\
\multicolumn{2}{l}{
([He broke] this and that.)}\\
\end{tabular}
\end{minipage}
\end{equation}

(1) When a complemented verb phrase exists in the sentences, 
we use surface expressions (clue words). 
This is because 
an elliptical sentence in the case (1) is 
in one of several typical patterns and 
has some clue words. 
For example, 
when the end of an elliptical sentence is the clue word ``MO (also)'', 
the system judges that 
the sentence is a repetition of the previous sentence 
and the complemented verb ellipsis is 
the verb phrase of the previous sentence. 

\begin{figure}[t]
  \begin{center}
  \begin{tabular}[t]{lll}
 & {\bf The matching part} & {\bf The latter part}\\[0.5cm]
KON'NANI & \underline{UMAKU IKUTOWA} & OMOENAI. \\
(like this)  & (it succeeds) & (I don't think)\\
\multicolumn{3}{l}{
(I don't think that it succeeded like this)}\\[0.5cm]
ITUMO & \underline{UMAKU IKUTOWA} & KAGIRANAI.\\
(every time)  & (it succeeds) & (cannot expect to)\\
\multicolumn{3}{l}{
(You cannot expect to succeed every time.)}\\[0.5cm]
KANZENNI & \underline{UMAKU IKUTOWA} & IENAI.\\
(completely)  & (it succeeds) & (it cannot be said)\\
\multicolumn{3}{l}{
(It cannot be said that it succeeds completely)}\\
\end{tabular}
  \end{center}
\caption{Sentences containing ``UMAKU IKUTOWA (it succeeds)'' in a corpus (examples)}
\label{tab:how_to_use_corpus}
\end{figure}

(2) When a complemented verb phrase does not appear in the sentences, 
we use examples. 
The reason is that 
omitted verb phrases in this case (2) 
are diverse and 
we use examples to construct the omitted verb phrases. 
The following is an example of 
a complemented verb phrase that does not appear in the sentences. 
\begin{equation}
  \begin{minipage}[h]{10cm}
    \begin{tabular}[t]{lll}
      SOU & UMAKU IKUTOWA & [OMOENAI] .\\
      (so) & (succeed so well) & (I don't think)\\
\multicolumn{3}{l}{
  ([I don't think] it succeeds so well. )}
    \end{tabular}
  \end{minipage}
\label{eqn:6c_souumaku}
\end{equation}
When we want to resolve the verb phrase ellipsis 
in this sentence ``SOU UMAKU IKUTO WA [OMOENAI]'', 
the system gathers sentences 
containing the expression 
``SOU UMAKU IKUTOWA (it succeeds so well. )'' from corpus 
as shown in Figure~\ref{tab:how_to_use_corpus}, 
and judges that 
the latter part in the obtained sentence 
(in this case, ``OMOENAI (I don't think)'' etc.) 
is the desired complemented verb phrase. 

\section{Categories of Verb Phrase Ellipsis}

\begin{figure}[t]
      \begin{center}
      \epsfile{file=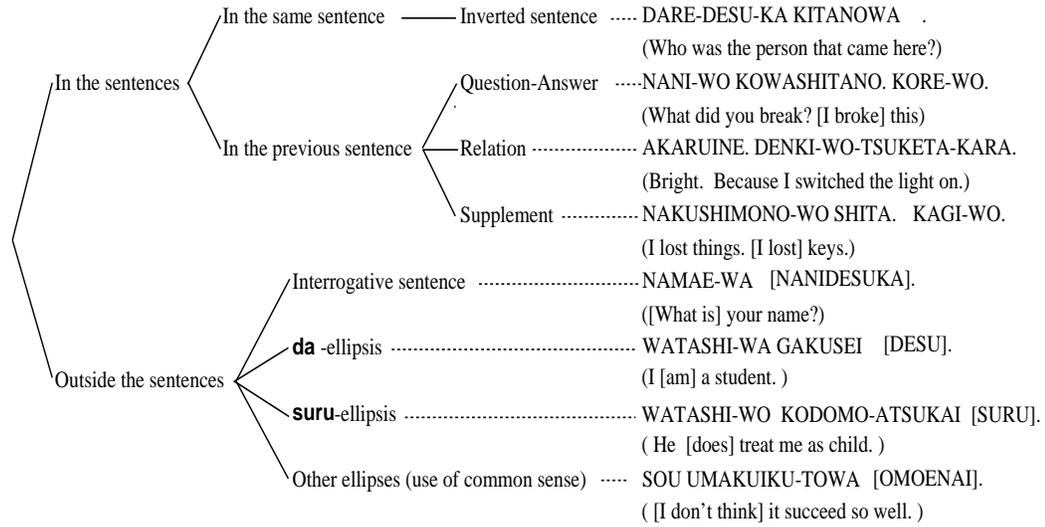,height=7cm,width=14cm} 
      \end{center}
    \caption{Categories of verb phrase ellipsis}
    \label{fig:shouryaku_bunrui}
\end{figure}

We handle only ellipses in the ends of sentences. 
Although there are some ellipses in the inner part of sentences, 
we think that they should be solved as problem of syntax 
and we do not deal with them. 

We classified verb phrase ellipses 
from the view point of machine processing. 
The classification is shown in Figure \ref{fig:shouryaku_bunrui}. 
First, we classified 
verb phrase ellipses 
by checking 
whether there is a complemented verb phrase in the sentences or not. 
If 
there is a complemented verb phrase in the sentences, 
we classified 
verb phrase ellipses 
by checking 
whether the complemented verb phrase is in the same sentence or 
in the previous sentence. 
Finally, 
we classified 
verb phrase ellipses 
by meaning. 
``In the sentences'', ``Outside the sentences'', 
``In the sentence'', and ``In the previous sentence'' 
in Figure \ref{fig:shouryaku_bunrui} 
represent where the complemented verb phrase exists, respectively. 
Although 
the above classification is not perfect 
and needs modification, 
we think that it is useful 
to understand the outline of verb phrase ellipses in machine processing. 

The feature and the analysis 
of each category of verb phrase ellipsis are described 
in the following sections. 

\subsection{When a Complemented Verb Phrase Ellipsis Appears 
in the Sentences}

\subsection*{Inverted Sentence}

Inverted sentences have expressions 
which are normally at the end of a sentence 
in the inner part of the sentence. 
For example, 
the following sentence has 
the words ``DARE DESUKA (Who is)'', 
an inverted expression 
normally at the end of a sentence. 
\begin{equation}
  \begin{minipage}[h]{11.5cm}
  \begin{tabular}[t]{lll}
DARE & DESUKA, & KITA-NO-WA\\
(who)& (is)   & (the person that came here)\\
\multicolumn{3}{l}{
(Who was the person that came here?)}\\
\end{tabular}
\end{minipage}
\end{equation}

Therefore, we analyze 
inverted sentences as followed. 
When a sentence has 
an expression which is normally at the end of a sentence 
and followed by a comma, 
the system judges the sentence to be an inverted sentence. 

\subsection*{Question--Answer}

In question--answer sentences  
verbs in answer sentences are often omitted, 
when answer sentences use the same verb 
as question sentences. 
For example, 
the verb of ``KORE WO (this)'' is omitted 
and is ``KOWASHITA (break)'' in the question sentence. 
\begin{equation}
  \begin{minipage}[h]{11.5cm}
  \begin{tabular}[t]{ll}
NANI-WO & KOWASHITANO\\
(what)& (break)  \\
\multicolumn{2}{l}{
(What did you break?)}\\
\end{tabular}
  
\vspace{0.3cm}

  \begin{tabular}[t]{ll}
KORE-WO & [KOWASHITA].\\
(this)  & (break)\\
\multicolumn{2}{l}{
([I broke] this.)}\\
\end{tabular}
\end{minipage}
\end{equation}

The system judges whether 
the sentences are question--answer sentences or not 
by using surface expressions such as ``NANI (what)'', 
and, if so, 
it judges that the complemented verb phrase 
is the verb phrase of the question sentence. 

\subsection*{Relation}

In verb phrase ellipsis, 
there is a phenomenon 
that 
an elliptical sentence whose end is a conjunctive particle 
relates causatively, contrastingly or conditionally
to the previous sentence, 
and they make inverted sentence across two sentences. 
For example, 
``DENKI-WO TSUKETA-KARA (Because I switched the light on.)''
is the reason 
for the previous sentence ``AKARUINE (bright)''. 
The omitted element of 
``DENKI-WO TSUKETA-KARA'' is ``AKARUINE (bright)''. 

\begin{equation}
  \begin{minipage}[h]{11.5cm}
  \begin{tabular}[t]{l}
AKARUI.\\
(bright) \\
\multicolumn{1}{l}{
(Bright.)}\\
\end{tabular}

\vspace{0.3cm}

  \begin{tabular}[t]{ll}
DENKI-WO & TSUKETA-KARA.\\
(the light)  & (switch on)\\
\multicolumn{2}{l}{
(Because I switched the light on.)}\\
\end{tabular}
\end{minipage}
\end{equation}

When a sentence has a conjunctive particle at the end, 
the system normally judges that 
the complemented verb phrase is the verb at the end of the previous sentence. 
But, 
there are some cases 
that a conjunctive particle is used for 
indicating hesitation, 
the sentence is not in contrast 
to the previous sentence. 
\begin{equation}
  \begin{minipage}[h]{11.5cm}
  \begin{tabular}[t]{lll}
OKIKI-SHITE & IINOKA & WAKARIMASENGA.\\
(ask)  & (whether it is all right) & (do not know)\\
\multicolumn{3}{l}{
(Although I don't know 
whether you mind I ask you, ...)}
\end{tabular}
\end{minipage}
\end{equation}
Therefore, 
in the case of ``NONI (but)'' 
which is easy to relate to the previous sentence, 
the system judges that 
the complemented verb phrase is the previous sentence. 
In the case of the other particles 
if the previous sentence is an interrogative sentence, 
the system judges 
that the sentence contrasts to the previous sentence, 
and otherwise, 
the system judges 
that the sentence does not contrast to the previous sentence 
and indicates a kind of feeling. 

\subsection*{Supplement}

In sentences which play a supplementary role to 
the previous sentence, 
verb phrases are sometimes omitted. 
For example, 
the second sentence 
is supplementary, explaining 
that ``the things I lost'' is ``keys''. 
\begin{equation}
  \begin{minipage}[h]{11.5cm}
  \begin{tabular}[t]{ll}
MONO-WO & NAKUSHITA.\\
(things)  & (lost)\\
\multicolumn{2}{l}{
(I lost things.)}
\end{tabular}

\vspace{0.3cm}

  \begin{tabular}[t]{ll}
KAGI-WO & [NAKUSHITA.]\\
(keys)  & (lost)\\
\multicolumn{2}{l}{
([I lost] keys. )}
\end{tabular}
\end{minipage}
\end{equation}

To solve this, we present the following two methods 
using word meanings. 
The first method is 
when the word at the end of the elliptical sentence
is semantically similar to the word 
of the same case element in the previous sentence, 
they correspond, 
and 
the omitted verb is judged to be 
the verb of the word of the same case element in the previous sentence. 
In this case, 
since ``MONO (thing)'' and ``KAGI (key)'' are semantically similar 
in the sense that they are both objects, 
the system judges they correspond,
and the verb of ``KAGI (key)'' is ``NAKUSHITA (lost)''. 

The second method is for 
when 
the same case element in the previous sentence is 
omitted. 
\begin{equation}
  \begin{minipage}[h]{11.5cm}
  \begin{tabular}[t]{ll}
NAKUSHITA.\\
(lost)\\
\multicolumn{2}{l}{
(I lost.)}
\end{tabular}

\vspace{0.3cm}

  \begin{tabular}[t]{ll}
KAGI-WO & [NAKUSHITA.]\\
(keys)  & (lost)\\
\multicolumn{2}{l}{
([I lost] keys. )}
\end{tabular}
\end{minipage}
\end{equation}
In this case, 
the system 
checks the semantic distance 
between ``KAGI (key)'' and 
the words which are easily filled in the WO case (object) 
of the ``NAKUSU (lose)'' 
by using the case frame of the verb ``NAKUSU (lose)''
\footnote{
IPAL case frame dictionary\cite{ipal} 
has the information 
of what kind of words can be filled in each case frame. 
In this work, we use this information. 
}
. 
If they are semantically similar, 
the system judges that 
the omitted verb phrase is ``NAKUSU (lose)''. 

In addition to these methods, 
we use methods using surface expressions. 
For example, 
when a sentence has clue words such as the particle ``MO'' 
(which indicates repetition), 
the sentence is judged to be the supplement of the previous sentence. 

There are many cases
when an elliptical sentence is the supplement of the previous sentence. 
In this work, 
if there is no clue, 
the system judges that 
an elliptical sentence is the supplement of the previous sentence. 

\subsection{When a Complemented Verb Phrase does not Appear 
in the Sentences}

\subsection*{Interrogative Sentence}

Sometimes, in interrogative sentences, 
the particle ``WA'' is at the end of the sentence
and the verb phrase is omitted. 
For example, 
the following sentence is an interrogative sentence 
and the verb phrase is omitted. 
\begin{equation}
  \begin{minipage}[h]{11.5cm}
  \begin{tabular}[t]{ll}
NAMAE-WA & [NANI-DESUKA.]\\
(name)  & (what?)\\
\multicolumn{2}{l}{
([What is] your name?)}
\end{tabular}
\end{minipage}
\end{equation}

If the end is of the form of ``Noun $+$ WA'', 
the sentence is probably an interrogative sentence, 
and thus 
the system judges it to be an interrogative sentence
\footnote{
Since this work is verb phrase ellipsis resolution, 
the system must complement a verb phrase 
such as ``NANI-DESUKA (what?)''. 
But the expression of the verb phrase 
changes according to the content of the interrogative sentence 
and we do not deal with this problem in this work. }.

\subsection*{{\it da}-Ellipsis}

When the end of the previous sentence is a noun phrase, 
the copula ``DA (be)'' is often omitted. 
\begin{equation}
  \begin{minipage}[h]{11.5cm}
  \begin{tabular}[t]{lll}
WATASHI-WA & GAKUSEI & [DESU].\\
(I)  & (student) & (be)\\
\multicolumn{3}{l}{
(I [am] a student.)}
\end{tabular}
\end{minipage}
\end{equation}
In this example, 
the copula ``DA (be)'' is omitted 
from the sentence 
``WATASHI-WA GAKUSEI DESU (I am a student.). 

The analysis of this case is performed by 
checking 
whether the end of the sentence is a noun phrase 
and by using syntactic structures 
such that there is a subject. 

\subsection*{{\it suru}-Ellipsis}

When the end of the previous sentence is a noun phrase, 
the basic verb ``SURU (do)'' is often omitted. 
\begin{equation}
  \begin{minipage}[h]{11.5cm}
  \begin{tabular}[t]{lll}
WATASHI-WO & KODOMO-ATSUKAI & [SURU].\\
(I)  & (to treat as child) & (do)\\
\multicolumn{3}{l}{
(He [does] treat me as child.)}
\end{tabular}
\end{minipage}
\end{equation}
In this example, 
the verb ``SURU (do)'' is omitted 
from the sentence 
``WATASHI-WO KODOMO-ATSUKAI SURU. (He treats me as child.)''. 

The analysis of this problem is done by 
checking 
whether the end of the sentence is a verbal noun 
and 
whether the {\it rentai}-form modifier modifies the verbal noun
\footnote{
A modifier is in the {\it rentai}-form, 
when it modifies a nominal phrase. }. 

\subsection*{Other Ellipses (Resolved Using Common Sense)}

In the case of ``Outside the sentences'' 
the following example exists 
besides ``Interrogative sentence'', 
``{\it da}-ellipsis'', and ``{\it suru}-ellipsis''. 
\begin{equation}
  \begin{minipage}[h]{11.5cm}
  \begin{tabular}[t]{llll}
JITSU-WA & CHOTTO & ONEGAIGA & [ARU-NO-DESUGA].\\
(the truth)  & (a little) & (request) & (I have)\\
\multicolumn{4}{l}{
(To tell you the truth, [I have] a request.)}
\end{tabular}
\end{minipage}
\end{equation}
This kind of ellipsis does not have 
the complemented expression in sentences. 
The form of the complemented expression has various types. 
This problem is difficult to analyze. 

To solve this problem, 
we estimate a complemented content 
by using a large amount of linguistic data. 

When Japanese people read the above sentence, 
they naturally recognize the omitted verb is ``ARIMASU (I have)''. 
This is because 
they empirically have the sentence 
``JITSU-WA CHOTTO ONEGAIGA ARU-NO-DESUGA.
(To tell the truth, I have my request.)'' 
in their mind. 
When we perform the same interpretation using 
a large amount of linguistic data, 
we detect the sentence 
containing an expression 
which is semantically similar to 
``JITSU-WA CHOTTO ONEGAIGA.
(To tell you the truth, (I have) a request.)'', 
and 
the latter part of ``JITSU-WA CHOTTO ONEGAIGA'' 
is judged to be 
the content of the ellipsis. 
In this work, 
we solve this problem by using the above method. 

\section{Verb Phrase Ellipsis Resolution System}

\subsection{Procedure}
\label{6c_wakugumi}

In this work, verb phrase ellipses are resolved 
in the same framework as Chapter \ref{chap:noun}. 
Before the verb phrase ellipsis resolution process, 
sentences are transformed into a case structure 
by the case structure analyzer\cite{csan2_ieice}. 
Verb phrase ellipses are resolved 
by heuristic rules for each sentence from left to right. 
Using these rules, 
our system gives possible complemented verb phrases some points, 
and it judges that the possible complemented verb phrase 
having the maximum point total is the desired complemented verb phrase. 

The heuristic rules are given in the following form. 

\begin{center}
    \begin{minipage}[c]{12cm}
      \hspace*{0.7cm}{\sl Condition} $\Rightarrow$ \{ {\sl Proposal, Proposal,} .. \}\\
      \hspace*{0.7cm}{\sl Proposal} := ( {\sl Possible complemented verb phrase,} \, {\sl Point} )
    \end{minipage}
\end{center}

\noindent
Surface expressions, semantic constraints, 
referential properties, etc., are written as conditions in the {\sl Condition} section. 
A possible complemented verb phrase 
is written in the {\sl Possible complemented verb phrase} section. 
{\sl Point} means the plausibility of the possible complemented verb phrase. 

\subsection{Heuristic Rule}
\label{rule}
\label{sec:6c_ref_pro}

We made 22 heuristic rules for verb phrase ellipsis resolution. 
We show all the rules in Table~\ref{tab:doushi_shouryaku_bunrui}. 
These rules are made by examining 
training sentences in Section \ref{sec:6c_jikken} by hand. 
When 
the system analyzes verb phrase ellipsis, 
it also analyzes anaphora in noun phrases and pronouns. 
The rules for this resolution are shown in 
Chapter \ref{chap:noun}, Chapter\ref{chap:indian}, 
and Chapter \ref{chap:deno}. 

For these rules
a semantic marker dictionary \cite{imiso-in-BGH} 
is used 
to determine 
whether 
a word means a human, time, etc. 


The value $s$ in Rule 12 and Rule 13 is 
given from the semantic similarity between 
``Noun $X$ and Noun $Y$'' in EDR concept dictionary \cite{edr_gainen_2.1}. 
This similarity is given 
($nz$ + $nz$)/($nx$ + $ny$), 
let $nx$ stand for 
the number of links between 
the top node and the node of Noun X, 
let $ny$ stand for 
the number of links between 
the top node and the node of Noun Y, 
let node $Z$ stand for 
the intersection node from Noun X and Noun Y to top node, 
and let $nz$ stand for 
the number of the links between 
the top node and the node of Noun Z\cite{nlp}. 

The corpus (linguistic data) used in Rule 22 is 
a set of newspapers (one year, about 70,000,000 characters). 
The method detecting a similar sentence is performed 
by sorting the corpus in advance 
and using a binary search. 

\begin{table}[p]
  \footnotesize
  \caption{Rule for verb phrase ellipsis resolution}
    \label{tab:doushi_shouryaku_bunrui}
  \begin{center}
\begin{tabular}[c]{|r|p{4cm}|p{2.5cm}|p{0.5cm}|p{4cm}|}\hline
  & Condition & Candidate & \multicolumn{1}{|@{}c@{}|}{Point} & Example sentence\\\hline
\multicolumn{5}{|c|}{Rule in the case that 
a verb ellipsis does not exist}\\\hline
  1 &
  When the end of the sentence is 
  a formal form of a verb 
  or terminal postpositional particles such as 
  ``YO'' and ``NE'', &
  the system judges that a verb phrase ellipsis does not exist. &
  30 & 
  SONO MIZUUMI WA, KITANO KUNINI ATTA.
  (The lake was in a northern country.)
  \\
  2&
  When the end of the sentence is 
  a person's name or 
  a word signifying a human being,&
  a verb phrase ellipsis does not exist. &
  30 &
  ``HAI, SENSEI.'' (``Yes, sir.'')\\
  3&
  When the end is 
  an imperative form of a verb,&
  the sentence is an imperative sentence 
  and 
  a verb phrase ellipsis does not exist. &
  30&
  ``SAA, MEWO TSUBUTTE'' (Here, close your eyes.)\\
  4&
  When the end is 
  the conjunctive particle ``GA'',  &
  a verb phrase ellipsis does not exist. 
  &
  $5$&
  ``CHOTTO SHITSUMON-GA ARUNO DESUGA'' 
  (Well, I have some questions.)\\\hline
\multicolumn{5}{|c|}{Rule in the case of ``Inverted sentence''}\\\hline
5&
  When 
  the sentence has an expression 
  normally at the end of a sentence 
  in the inner part, &
  it is judged to be an inverted sentence.&
  $10$&
  ``DARE DESUKA, KITA-NO-WA'' 
  (``Who was the person that came here?'' )\\\hline
\multicolumn{5}{|c|}{Rule in the case of ``Question--Answer''}\\\hline
6&
  When the sentence has an expression 
  which indicates a reply 
  and the previous sentence has 
  an expression 
  which indicates an interrogative sentence
  such as ``KA (?)'', &
  the verb phrase at the end of the interrogative sentence&
  $5$&
  ``CHIKAYOTTE KANSATSU SHITEMO IIDESHOUKA.'' 
  ``DOUZO, GOJIYUUNI...'' 
  (``Can I approach and look at this?'' 
  ``Yes, please.'')
  \\\hline
\end{tabular}
\end{center}
\end{table}

\begin{table}[p]
\center{Table \ref{tab:doushi_shouryaku_bunrui}: Rule for verb phrase ellipsis resolution (cont.)}
  \footnotesize
  \begin{center}
\begin{tabular}[c]{|r|p{4cm}|p{2.5cm}|p{0.5cm}|p{4cm}|}\hline
  & Condition & Candidate & \multicolumn{1}{|@{}c@{}|}{Point} & Example sentence\\\hline
\multicolumn{5}{|c|}{Rule in the case of ``Question--Answer''}\\\hline
  7&
  When the previous sentence 
  has an interrogative pronoun such as 
  ``DARE (who)'' and ``NANI (what)'', &
  the verb modified by the interrogative pronoun &
  $5$&
  ``DARE-WO KOROSHITANDA'' 
  ``WATASHI-GA KATTE-ITA SARU-WO [KOROSHITA]'' 
  (``Who did you kill?''
  ``[I killed] my monkey'')\\\hline
\multicolumn{5}{|c|}{Rule in the case of ``Relation''}\\\hline
  8&
  When the end is 
  postpositional particles which indicates cause
  such as ``NODE'' and ``KARA'', &
  the sentence is interpreted to be 
  the reason for the previous sentence&
  5 &
  ``TOCHI-WO AGERU-WAKE-NIWA-IKANAI. 
  SOKONI, YASHIRO-WO TATE-NAKUTEWA-NARANAI-NODAKARA'' 
  (``We can't give you the lot. 
  Because we must build a shrine there.'')\\
  9&
  When the end is 
  a postpositional particle such as 
  ``NONI''  and ``KEREDOMO'', & 
  the sentence is interpreted to 
  contrast with the previous sentence. &
  $5$&
  ``KORE-GA AKUMA-TOWA-NEE. 
  MOU-SUKOSHI DOUDOU-TO SHITA MONO-KA-TO OMOTTE-ITA-NONI'' 
  (``This is a devil. 
  Although I thought 
  it was majestic.'' )
  \\
  10&
  When the end is 
  a conditional form of a verb or 
  postpositional particles indicating conditions, &
  the sentence is interpreted to be 
  the condition of the previous sentence. &
  $5$&
  ``SORENARA, IIJANAIKA. 
  NANIMO, KOUBAN-NI-MADE KONAKUTEMO.''
  (It is good.
  Unless you came to the police office.)\\\hline
\end{tabular}
\end{center}
\end{table}

\begin{table}[p]
\center{Table \ref{tab:doushi_shouryaku_bunrui}: Rule for verb phrase ellipsis resolution (cont.)}
  \footnotesize
  \begin{center}
\begin{tabular}[c]{|r|p{4cm}|p{2.5cm}|p{0.6cm}|p{4cm}|}\hline
  & Condition & Candidate & \multicolumn{1}{|@{}c@{}|}{Point} & Example sentence\\\hline
\multicolumn{5}{|c|}{Rule in the case of ``Supplement''}\\\hline
  11&
  When the end is 
  an infinitive form of a verb, &
  the sentence is interpreted to be 
  the supplement of the previous sentence
  and the verb phrase at the end 
  of the previous sentence is judged to be 
  the complemented verb phrase & 
  5&
  MESHITSUKAI-WA HEYA-NI HAIRI, 
  ESA-WO TORIKAETA. 
  SHUUKURIIMU-MO KUWAETE [TORIKAETA]. 
  (A servant came into the room and 
  changed the pet food. 
  [He changed it] with a cream puff. )\\
  12&
  When the end is Noun X followed by a case postpositional particle, 
  there is a Noun Y followed by the same case postpositional particle 
  in the previous sentence, 
  and the semantic similarity between Noun X and Noun Y is 
  a value $s$, &
  the verb phrase modified by Noun Y &
  $s*20$ $-2$&
  SUBETENO AKU-GA NAKUNATTEIRU. 
  GOUTOU-DA-TOKA SAGI-DA-TOKA, ARAYURU HANZAI-GA [NAKUNATTEIRU]. 
  (All the evils have disappeared. 
  All the crimes such as robbery and fraud [have disappeared]. )
  \\
  13&
  When the end is Noun X followed by a case postpositional particle, 
  there is a zero pronoun of a verb phrase Y in the same case element 
  in the previous sentence, 
  and the semantic similarity between Noun X and 
  the words which is easy to be filled in the zero pronoun, 
  described in the case frame &
  the verb phrase Y&
  $s*20$ $-2$&
  WATASHI-WA [JUUTAKU-WO] DOURYOU-NI YUBISASHITE MISETA. 
  OOKINA NIRE-NO-KI NO SHITA-NI ARU KOHUUNA TSUKURI-NO JUUTAKU-WO. 
  (I pointed my colleague [to the house]. 
  An old-fashioned house under the big elm.)
  \\\hline
\end{tabular}
\end{center}
\end{table}

\begin{table}[p]
\center{Table \ref{tab:doushi_shouryaku_bunrui}: Rule for verb phrase ellipsis resolution (cont.)}
  \footnotesize
  \begin{center}
\begin{tabular}[c]{|r|p{4cm}|p{2.5cm}|p{0.6cm}|p{4cm}|}\hline
  & Condition & Candidate & \multicolumn{1}{|@{}c@{}|}{Point} & Example sentence\\\hline
\multicolumn{5}{|c|}{Rule in the case of ``Supplement''}\\\hline
  14&
  When the end is the postpositional particle ``MO'' 
  or 
  there is an expression 
  which indicates repetition such as ``MOTTOMO'', 
  the repetition of the same speaker's previous sentence is interpreted, &
  the verb phrase at the end 
  of the same speaker's previous sentence is judged to be 
  a complemented verb phrase & 
  $5$&
  ``OTONATTE WARUI KOTO BAKARI SHITEIRUNDAYO. 
  YOKU WAKARANAIKEREDO, WAIRO NANTE KOTO-MO [SHITEIRUNDAYO].''
  (``Adults do only bad things. 
  I don't know, but [they do] bribe.'')\\
  15&
  When the previous sentence is an interrogative sentence, &
  the verb phrase in the end of the previous sentence&
  $1$&
  \\
  16&
  In all cases,&
  the previous sentence&
  $0$&\\\hline
\multicolumn{5}{|c|}{Rule in the case of ``Interrogative sentence''}\\\hline
  17&
  When the end is a noun followed by postpositional particle ``WA'', &
    the sentence is interpreted to be an interrogative sentence.&
  $3$&
  ``NAMAE-WA [NANI-DESUKA]''
  (``[What is] your name?'')
  \\\hline
\multicolumn{5}{|c|}{Rule in the case of ``{\it da}-ellipsis''}\\\hline
  18&
  When the end is a noun or 
  a postpositional particle such as ``BAKARI (only)'', ``DAKE (only)'',
  and 
  there is a noun phrase 
  followed by a postpositional particle 
  ``WA (topic)'', ``MO (subject)'', and ``GA (subject)'' 
  which corresponds to the subject in the sentence,& 
  the system judges it as {\it da}-ellipsis &
  $2$&
  ``KORE-WA WATASHI-NO KANCHIGAI [DESU]'' 
  (``This [is] my mistake.'')\\
  19&
  When the end is a noun which signifies time,&
  the system judges it as {\it da}-ellipsis &
  $5$&
  SONO TSUGI-NO NATSU [NO-KOTO-DESU].
  ([It is] the next summer.)\\\hline
\end{tabular}
\end{center}
\end{table}

\begin{table}[p]
\center{Table \ref{tab:doushi_shouryaku_bunrui}: Rule for verb phrase ellipsis resolution (cont.)}
  \footnotesize
  \begin{center}
\begin{tabular}[c]{|r|p{4cm}|p{2.5cm}|p{0.6cm}|p{4cm}|}\hline
  & Condition & Candidate & \multicolumn{1}{|@{}c@{}|}{Point} & Example sentence\\\hline
\multicolumn{5}{|c|}{Rule in the case of ``{\it da}-ellipsis''}\\\hline
  20&
  When the end is a noun or 
  a postpositional particle such as ``BAKARI (only)'', ``DAKE (only)'',&
  the system judges it as {\it da}-ellipsis &
  $1$&
  ATO-WA KOUGEKI-WO MATSUBAKARI [DESU].
  (What I do [is] only 
  wait for the attack. )
 \\\hline
\multicolumn{5}{|c|}{Rule in the case of ``{\it suru}-ellipsis''}\\\hline
  21&
  When the end is a verbal noun 
  which is not modified the {\it rentai} modifier, &
  the system judges it as {\it suru}-ellipsis &
  $2$&
  WATASHI-WO KODOMO-ATSUKAI [SURU]. 
  (He [does] treat me like a child.)\\\hline
\multicolumn{5}{|c|}{Rule in the case of use of common sense}\\\hline
  22&
  When the system detects 
  a sentence 
  containing the longest expression at the end of the sentence 
  from corpus, 
  (If the highest frequency is much 
  higher than the second highest frequency, 
  the expression is given 9 points, 
  otherwise it is given 1 point. )
  &
  the expression of the highest frequency 
  in the latter part of the detected sentences 
  &
  1 or 9&
  SOU UMAKU IKUTOWA [OMOENAI]. 
  ([I don't think] it will succeed.)
  \\\hline
\end{tabular}
\end{center}
\end{table}

\begin{figure}[t]
\fbox{
\begin{minipage}[h]{13cm}
  \begin{tabular}[t]{l}
MURI-MO-ARIMASENWA. \\
\multicolumn{1}{l}{
(You may well do so.)}
\end{tabular}
  \begin{tabular}[t]{ll}
HAJIMETE & OAISURU-NO-DESUKARA. \\
(for the first time)  & (I meet you)\\
\multicolumn{2}{l}{
(I meet you for the first time)}
\end{tabular}
  \begin{tabular}[t]{llll}
JITSU-WA & CHOTTO & ONEGAIGA & (ARU-NO-DESUGA).\\
(the truth)  & (a little) & (request) & (I have)\\
\multicolumn{4}{l}{
(To tell you the truth, [I have] a request.)}
\end{tabular}

\vspace{0.5cm}

\begin{tabular}[h]{|l|r|r|r|}\hline
Candidate  &  the end of the previous sentence &  ``ARIMASU (I have)''\\\hline
Rule 16     &   0 point        &           \\\hline
Rule 22     &              &   1 point     \\\hline
Total score    &   0 point        &   1 point     \\\hline
\end{tabular}

\vspace{0.5cm}

\begin{tabular}[h]{|l|l|}\hline
the latter part of the sentence 
containing ``ONEGAI GA'' & Frequency\\\hline
ARIMASU (I have)  & 5\\
ARU     (I have)  & 3\\\hline
\end{tabular}

\caption{Example of verb phrase ellipsis resolution}
\label{tab:6c_dousarei}
\end{minipage}
}\end{figure}

\subsection{Example of Verb Phrase Ellipsis Resolution}

We show an example of a verb phrase ellipsis resolution 
in Figure \ref{tab:6c_dousarei}. 
Figure \ref{tab:6c_dousarei} shows that 
the verb phrase ellipsis 
in ``ONEGAI (request)'' 
was analyzed well. 

Since the end of the sentence is not 
an expression which can normally be at the end of a sentence, 
Rule 1 was not satisfied and 
the system judged that 
a verb phrase ellipsis exists. 
By Rule 16 the system took 
the candidate ``the end of the previous sentence''. 
Next, by Rule 22 using corpus, 
the system took the candidate ``ARIMASU (I have)''. 
Although 
there are ``ARU (I have)'' and ``ARIMASU (I have)'', 
the frequency of ``ARIMASU (I have)'' is more than the others 
and it was selected as a candidate. 
The candidate ``ARIMASU (I have)'' 
having the best score was properly 
judged to be the desired complemented verb phrase. 

\section{Experiment and Discussion}

\label{sec:6c_jikken}


We ran the experiment on the novel 
``BOKKOCHAN''\cite{bokko}. 
This is because 
novels contain various verb ellipses. 
In the experiment, we divided 
the text into 
training sentences and 
test sentences. 
We made heuristic rules by examining training sentences. 
We tested our rules by using test sentences. 
We show the results of verb phrase ellipsis resolution 
in Table \ref{tab:0verb_result}. 

\begin{table}[p]
\fbox{
\begin{minipage}[h]{13cm}
\small
    \caption{Result of resolution of verb phrase ellipsis}
    \label{tab:0verb_result}
    \label{tab:6c_sougoukekka}
\vspace*{0.5cm}
  \begin{center}
\begin{tabular}[c]{|@{ }lll@{}|@{ }r@{}c@{}|@{ }r@{}c@{}|@{ }r@{}c@{}|@{ }r@{}c@{}|}\hline
&&      &\multicolumn{4}{c|}{Training sentences}
        &\multicolumn{4}{c|}{Test sentences}\\\cline{4-11}
&&      &\multicolumn{2}{c|}{Recall}
        &\multicolumn{2}{c|}{Precision}
        &\multicolumn{2}{c|}{Recall}
        &\multicolumn{2}{c|}{Precision}\\\hline
\multicolumn{3}{|l|}{Total score}        &  92\% &(129/140)&  90\% &(129/144)&  84\% &(125/148)&  82\% &(125/152)\\\hline
 &\multicolumn{2}{l|}{In the sentences}  & 100\% & (57/57) &  85\% & (57/67) &  94\% & (64/68) &  81\% & (64/79) \\\hline
& &Inverted sentence   & 100\% & (13/13) & 100\% & (13/14) & 100\% & ( 8/ 8) &  80\% & ( 8/10) \\
& &Question--Answer    & 100\% & ( 3/ 3) & 100\% & ( 3/ 3) & ---\% & ( 0/ 0) & ---\% & ( 0/ 0) \\
& &Relation            & 100\% & (24/24) &  89\% & (24/27) & 100\% & (33/33) &  85\% & (33/39) \\
& &Supplement          & 100\% & (17/17) &  74\% & (17/23) &  85\% & (23/27) &  77\% & (23/30) \\\hline
 &\multicolumn{2}{l|}{Outside the sentences}  &  87\% & (72/83) &  94\% & (72/77) &  76\% & (61/80) &  84\% & (61/73) \\\hline
& &Interrogative sentence & 100\% & ( 3/ 3) &  75\% & ( 3/ 4) & ---\% & ( 0/ 0) &   0\% & ( 0/ 3) \\
& &{\it da}-ellipsis   & 100\% & (54/54) & 100\% & (54/54) & 100\% & (51/51) &  96\% & (51/53) \\
& &{\it suru}-ellipsis & 100\% & ( 2/ 2) & 100\% & ( 2/ 2) & ---\% & ( 0/ 0) & ---\% & ( 0/ 0) \\
& &Other ellipses &  72\% & (13/18) &  76\% & (13/17) &  56\% & (10/18) &  59\% & (10/17) \\
& &Impossible  &   0\% & ( 0/ 6) & ---\% & ( 0/ 0) &   0\% & ( 0/11) & ---\% & ( 0/ 0) \\\hline
\end{tabular}
\end{center}

\vspace{0.5cm}

The training sentences are used to 
make the set of rules in Section~\ref{sec:6c_ref_pro}. \\
{
Training sentences \{the first half of a collection of short stories ``BOKKO CHAN'' \cite{bokko} (2614 sentences, 23 stories)\}\\
Test sentences \{the latter half of novels ``BOKKO CHAN'' \cite{bokko} (2757 sentences, 25 stories)\}

{\it Precision}\, is 
the fraction  of the ends of the sentences 
which were judged to have verb phrase ellipses. 
{\it Recall}\, is the fraction of the ends of the sentences 
which have the verb phrase ellipses. 
The reason why we use precision and recall to evaluate is 
that the system judges that 
the ends of the sentences which do not have 
the verb phrase ellipses 
have the verb phrase ellipses 
and we check these errors properly.

We made a new category ``Impossible'' 
which is not in Figure \ref{fig:shouryaku_bunrui}. 
This category represents 
when 
the utterance is interrupted in the middle of the sentence, 
or 
the reader cannot recognize the omitted content. 
Since they are difficult to be resolved 
and we want to properly evaluate 
the method of ``use of common sense'', 
we separated the category from 
``Other ellipses (use of common sense)''. 
}
\end{minipage}
}
\end{table}

To judge whether 
the result is correct or not, 
we used the following evaluation criteria. 
When the complemented verb phrase is correct, 
even if the tense, aspect, etc. are incorrect, 
we regard it as correct. 
For ellipses in interrogative sentences, 
if the system estimates that 
the sentence is an interrogative sentence, 
we judge it to be correct. 
When the desired complemented verb phrase appears in the sentences 
and the complemented verb phrase 
chosen by the rule using corpus 
is nearly equal to the correct verb phrase, 
we judge that it is correct. 

\subsection{Discussion}

As in Table \ref{tab:6c_sougoukekka} 
we obtained a recall rate of 84\% and 
a precision rate of 82\% in the estimation of indirect anaphora 
on test sentences. 
This indicates that 
our method 
is effective. 

The recall rate of ``In the sentences'' is 
higher than that of ``Outside the sentences''. 
For ``In the sentences'' 
the system only specifies the location of 
the complemented verb phrase. 
But 
in the case of ``Outside the sentences'' 
the system judges that 
the complemented verb phrase does not exist in the sentences 
and gathers the complemented verb phrase from other information. 
Therefore ``Outside the sentences'' is very difficult to analyze. 

The accuracy rate of ``Other ellipses (use of common sense)'' 
was not so high. 
But, 
since the analysis of the case of ``Other ellipses (use of common sense)'' 
is very difficult, 
we think that it is valuable 
to obtain a recall rate of 56\% 
and a precision rate 59\%. 
In both training sentences and test sentences, 
about half of all the error cases occurred 
because 
the solution proposed by the rule using corpus 
is correct  and 
the point is lower than that of the other rule 
or 
because 
the correct answer does not have the highest frequency 
but  
the second or third highest. 
This indicates that 
there is room for improving the method by using corpus. 
We think that 
when the size of corpus becomes larger, 
this method becomes very important. 
Although we calculate the similarity between 
the input sentence and the example sentence in the corpus 
only by using simple character matching, 
we think that 
we must use the information of semantics and the parts of speech 
when calculating the similarity. 
Moreover 
we must detect the desired sentence 
by using only examples of the type 
(whether it is an interrogative sentence or not) whose previous sentence 
is the same as the previous sentence of the input sentence. 

Although the accuracy rate of the category using surface expressions 
is already high, 
there are some incorrect cases 
which can be corrected 
by refining the use of surface expressions in each rule. 
There is also a case 
which requires a new kind of rule 
in the experiment on test sentences. 
\begin{equation}
  \begin{minipage}[h]{11.5cm}
\small
  \begin{tabular}[t]{llll}
SONOTOTAN & WATASHI-WA & OOKINA HIMEI-WO & KIITA.\\
(at the moment)& (I)   & (a scream) & (hear)\\
\multicolumn{4}{l}{
(At the moment, I heard a scream?)}\\
\end{tabular}

\vspace{0.3cm}

  \begin{tabular}[t]{l@{ }l@{ }l@{ }l}
NANIKA-NI & OSHITSUBUSARERU-YOUNA & OSOROSHII & KOE-NO.\\
(something)& (be crushed)   & (fearful) & (voice)\\
\multicolumn{4}{l}{
(of a fearful voice such that 
he was crushed by something)}\\
\end{tabular}
\end{minipage}
\end{equation}
In these sentences, 
``OSOROSHII KOE-NO (of a fearful voice)'' 
is the supplement of ``OOKINA HIMEI (a scream)'' in the previous sentence. 
To solve this ellipsis, 
we need the following rule. 
\begin{equation}
  \begin{minipage}[h]{10.5cm} 
    When the end is the form of ``noun X $+$ NO(of)'' and 
    there is a noun Z which is semantically similar to 
    noun Y in the examples of ``noun X $+$ NO(of) $+$ noun Y'', 
    the system judges that 
    the sentence is the supplement of noun Z. 
\end{minipage}
\end{equation}

\begin{table}[t]
    \caption{The number of ellipses in essays in ``TENSEI JINGO''}
    \label{tab:6c_tenseijingo_shutsugen}
  \begin{center}
\begin{tabular}[c]{|@{ }lll@{}|r|r|r|}\hline
&&      &\multicolumn{1}{c|}{In quotations}
        &\multicolumn{1}{c|}{Outside quotations}
        &\multicolumn{1}{c|}{Total}\\\hline
\multicolumn{3}{|l|}{Total}        & 5 & 34 & 39\\\hline
 &\multicolumn{2}{l|}{In the sentences}  & 1 & 1 & 2\\\hline
& &Inverted sentence   &  0 & 0 & 0\\
& &Question--Answer    &  0 & 0 & 0\\
& &Relation            &  1 & 0 & 1\\
& &Supplement          &  0 & 1 & 1\\\hline
 &\multicolumn{2}{l|}{Outside the sentences}  & 4 & 33 & 37\\\hline
& &Interrogative sentence & 0 & 0 & 0\\
& &{\it da}-ellipsis   &   0 & 28 & 28\\
& &{\it suru}-ellipsis &   0 & 0 & 0\\
& &Other ellipses &    4 & 5 & 9\\\hline
\end{tabular}
\end{center}
\end{table}

We experimented on novels 
in order to detect various ellipses. 
To check what kind of phenomena exist in other texts, 
we counted the number of ellipses in 
essays ``TENSEI JINGO'' (79 stories, 1871 sentences). 
The results are shown in Table~\ref{tab:6c_tenseijingo_shutsugen}. 
We find that 
the number of ellipses is small in essays 
where there are few conversational sentences. 
Although there are five cases 
in ``Other ellipses'' 
outside conversational sentences, 
they are all 
in the form of ``TO $+$ human being'' 
such as `` `... TAISHO-SURU' TO SHUSHOU [GA-ITTA]. 
(`I will take ...', [said] the prime minister)''. 
There are not many different kinds of elliptical phenomena 
in essays. 

\section{Summary}

This chapter 
described a practical way to resolve omitted verb phrases 
by using surface expressions and examples. 
We obtained a recall rate of 84\% and 
a precision rate of 82\% in the resolution of 
verb phrase ellipsis 
on test sentences. 
The accuracy rate of 
the case of 
complemented verb phrase appearing in the sentences 
was high. 
The accuracy rate of the case of using corpus (examples) was not so high. 
Since the analysis of this phenomena is very difficult, 
we think that it is valuable 
to have proposed 
a way of solving the problem to a certain extent. 
We think that 
when the size of corpus becomes larger and 
the machine performance becomes greater, 
the method of using corpus will become effective.

\chapter{Conclusion}
\label{chap:conc}

Anaphora resolution is important for 
language understanding, 
machine translation, and dialogue processing. 
We resolved varieties of anaphora by using 
surface expressions and examples. 
We experimented on several kinds of texts to test our methods. 
The results of these experiments indicate that 
our methods are effective. 


\section{Summary}


Chapter \ref{chap:ref} described a method of 
determining the referential property and number of noun phrases 
in Japanese sentences using surface expressions. 
The referential property of a noun phrase is 
how the noun phrase denotes the referent. 
The referential property is classified into three types: 
generic, definite and indefinite. 
A definite noun phrase refers to a given object. 
An indefinite noun phrase refers to a new object. 
In English, they correspond to 
a noun phrase with a definite article 
and a noun phrase with an indefinite article, respectively. 
A generic noun phrase refers to all objects 
which the noun phrase denotes. 
The number of a noun phrase is 
the number of the referent denoted by the noun phrase. 
The number is classified into three types: 
singular, plural, and uncountable. 
The referential property and the number of a noun phrase 
are basic factors in anaphora resolution. 
The system can grasp the outline of the referent of the noun phrase 
by using the referential property and the number of a noun phrase. 
The referential property and the number are also useful 
when the system generates the article 
in translating Japanese nouns into English. 
Many rules for the estimation of the referential property 
and the number of a noun phrase were written
in forms similar to rewriting rules in expert systems with scores.
We obtained the correct recognition scores of 85.5\% and 89.0\% 
in the estimation of referential property and number 
respectively for the sentences
which were used for the construction of our rules.
We tested these rules for some other texts,
and obtained the scores of 68.9\% and 85.6\%, respectively. 

Chapter \ref{chap:noun} gave a method for 
estimating the referent of a noun phrase in Japanese sentences 
using referential properties, modifiers, and possessors of noun phrases.
Since there are no articles in the Japanese language, 
it is difficult to decide whether two noun phrases 
have the same referent in Japanese. 
But we researched referential properties of noun phrases 
that correspond to articles using words in the sentences 
as in Chapter \ref{chap:ref}. 
We estimated referents of noun phrases using these referential properties.
For example if the referential property of a noun phrase is definite, 
the noun phrase can refer to a noun phrase that appears previously, and 
if the referential property of a noun phrase is indefinite, 
the noun phrase cannot refer to a noun phrase that appears previously. 
Furthermore we estimated referents of noun phrases 
using modifiers and possessors of noun phrases more precisely. 
As a result, we obtained a precision rate of 82\% 
and a recall rate of 85\% in the estimation 
of referent of noun phrases that have antecedents 
on training sentences, 
and obtained a precision rate of 79\% 
and a recall rate of 77\% 
on test sentences. 
We verified that it is effective to use 
referential properties, modifiers, and possessors of noun phrases 
through experiments.

Chapter \ref{chap:indian} described
how to resolve indirect anaphora resolution. 
A noun phrase can indirectly refer to an entity that has 
already been mentioned. For example, ``There is a house. 
The roof is white.'' indicates that ``the roof'' 
is associated with ``a house'', 
which was mentioned in the previous sentence. This kind of reference 
(indirect anaphora) has not been 
studied well in natural language processing, but is important for 
coherence resolution, language understanding, and machine translation. 
When we analyze indirect anaphora, 
we need a case frame dictionary for nouns 
containing an information about relationships between two nouns. 
But no noun case frame dictionary exists at present. 
Therefore, we used examples of ``X of Y'' 
and a verb case frame dictionary instead. 
We estimated indirect anaphora 
by using this information, 
and obtained a recall rate of 63\% and 
a precision rate of 68\% on test sentences. 
This indicates that the information of ``X of Y'' 
is useful 
when we cannot make use of a noun case frame dictionary. 
We made a hypothetical estimation 
that we can use a good noun case frame dictionary, 
and obtained the result with the recall and the precision rates of 
71\% and 82\%, respectively. 
Finally we proposed how to construct a noun case frame dictionary 
from examples of ``X of Y''.

Chapter \ref{chap:deno} described
how to estimate the referent of a pronoun in Japanese sentences. 
It is necessary to clarify referents of pronouns 
in machine translation and dialogue processing. 
We presented a method of estimating 
referents of demonstrative pronouns, personal pronouns, 
and zero pronouns in Japanese sentences
using examples, surface expressions, topics and foci.
In conventional work, 
semantic markers have been used 
for semantic constraints. 
On the other hand, we used examples for semantic constraints 
and showed in our experiments that 
examples are as useful as semantic markers. 
We also proposed many new methods for estimating referents of pronouns. 
For example, we used examples of the form ``X of Y'' for 
estimating referents of demonstrative adjectives. 
We used many useful conventional methods 
in addition to our new methods. 
When we experimented using these methods, 
we obtained a precision rate of 87\% 
in the estimation 
of referent of demonstrative pronouns, personal pronouns, 
and zero pronouns 
on training sentences, 
and obtained a precision rate of 78\% 
on test sentences. 

Chapter \ref{chap:0verb} described
the method of 
resolving verb phrase ellipsis using 
surface expressions and examples. 
When a complemented verb phrase appears in the sentences, 
the structure of the elliptical sentence is commonly 
in a typical form and 
the resolution is done by using surface expressions. 
When a complemented verb phrase does not appear in the sentences, 
the system resolved the elliptical sentence using examples. 
The analysis using examples is performed by 
gathering sentences containing the expression of 
the end of the elliptical sentence from linguistic data 
and judging the latter part of the matching expression 
in the gathered sentences to be the desired complemented verb phrase. 
As a result, 
we obtained a recall rate of 84\% and 
a precision rate of 82\% in the resolution of 
verb phrase ellipsis 
on test sentences. 

\section{Future Work in Anaphora Resolution}

\begin{itemize}


\item 
  {\bf Refinement of heuristic rules 
    using large collection of sentences}

  It is necessary to 
  refine heuristic rules in this work. 
  Although the points (certainty value) given by heuristic rules 
  are set in the training sentences, 
  it is necessary to 
  set them automatically 
  by using a computational learning algorithm. 
  At this time, we require large scale linguistic data for 
  refinement of heuristic rules 
  and learning the parameters of the points. 
  The construction of the linguistic data 
  need a syntactic structure analysis and 
  a case structure analysis. 
  But since a syntactic structure analysis and 
  a case structure analysis cannot be done 
  with high accuracy at present, 
  we cannot collect large amounts of linguistic data. 
  We must improve 
  a syntactic structure analyzer and 
  a case structure analyzer before refining heuristic rules. 

\item 
  {\bf Anaphora resolution using knowledge and reasoning}

  In this work, we resolved anaphora 
  by using only information 
  which is available at present. 
  But, there are problems 
  which require knowledge and reasoning 
  as in the following example \cite{Tanaka1}. 
\begin{equation}
  \begin{minipage}[h]{10cm}
    \begin{tabular}[t]{llll}
      KARE-WA & MIZU-TO & SHOKUEN-WO & MAZETA.\\
      (he) & (water) & (salt) & (mixed) \\
\multicolumn{3}{l}{
  (He mixed water and salt. )}
    \end{tabular}

\vspace{0.3cm}

    \begin{tabular}[t]{llll}
      KORE-WO & RUTSUBO-NI & SOSOIDA.\\
      (this) & (melting pot) & (advice)  & (poured)\\
\multicolumn{3}{l}{
  (He poured this into the melting pot. )}
    \end{tabular}
  \end{minipage}
\label{eqn:7c_suiron}
\end{equation}
What ``KORE (this)'' refers to is 
salty water 
which comes from mixing water and salt. 
To solve this problem, 
we need 
the knowledge that 
if we mix water and salt, 
salty water results. 
Solving this kind of problem requires many complicated analyses. 
Although this problem is very difficult, 
we must solve it 
for anaphora resolution to improve. 

\end{itemize}



\appendix

\mychapter
{\chapter{Rule for Referential Property and Number of Noun Phrase}}
{\chapter[Rule for Referential Property/Number]
{Rule for Referential Property and Number of Noun Phrase}}
\label{app:ref_num}

We have written 86 heuristic rules for the referential property and 48 
heuristic rules for the number.  
All the rules are given 
in Table~\ref{app:tab:rule_ref} and Table~\ref{app:tab:rule_num}. 

\begin{table}[p]
  \footnotesize
  \caption{Rule for referential property}
  \label{app:tab:rule_ref}
  \begin{center}
\begin{tabular}[c]{|r|p{3cm}|r|r|r|r|r|r|p{5cm}|}\hline
  & Condition & \multicolumn{2}{c|}{Indef} & \multicolumn{2}{c|}{Def} & \multicolumn{2}{c|}{Gener} & Example\\\cline{3-8}
& & 
P$^{*}$ & V & P & V& P & V & \\\hline
1 & When a noun is a personal pronoun, &
0 & 0 & 1 & 2 & 0 & 0 &
\underline{KARE}-WA SONO BENGOSHI-NO MUSUKO-NO HITORI-DESU. (\underline{He} is a son of that lawyer.)\\
2 & When a noun is an unique entity 
which does not have a modifier 
such as ``CHIKYU (the earth)'', &
0 & 0 & 1 & 2 & 0 & 0 & 
OOKU-NO HITOBITO-NO MOKUHYOU-WA \underline{CHIKYUU}-NO HEIWA-DESU. (The goal of many groups is peace on \underline{earth}.)\\
3 & When a noun is a proper noun 
which does not have a modifier, &
0 & 0 & 1 & 2 & 0 & 0 & 
\\
4 & When a noun is modified by a noun 
which signifies time, &
1 & 0 & 1 & 2 & 1 & 0 & 
KYOU-NO GOGO-NO \underline{YOTEI}-WA DOU-DESUKA. (What is your \underline{plan} in the afternoon today?)\\
5 & When a noun is ``HOU (on the part)'', &
0 & 0 & 1 & 0 & 1 & 0 & 
\\
6 & When a noun is followed by a particle ``WA'' 
which does not have a modifier, &
1 & 0 & 1 & 1 & 1 & 1 & 
\underline{SEKIYU-JIGYO}-WA WATASHI-GA TE-WO DASHITAKU NAI JIGYO-NO HITOTSU-DESU. (\underline{The oil business} is one business that I don't wish to get involved with.)\\
7 & When a noun is accompanied by a particle (WA), and the 
predicate is in the past tense, &
1 & 0 & 1 & 3 & 1 & 1 & 
\underline{IINKAI}-WA ZEN'IN SONO MONDAI-WO KAIKETSU SURUTAME-NI SHIGOTO-WO SHIMASHITA. (Everyone on \underline{the committee} worked to solve that problem.)\\\hline
\end{tabular}
\end{center}

$^{*}$P: possibility, V: value
\end{table}

\begin{table}[p]
\center{Table \ref{app:tab:rule_ref}: Rule for referential property (cont.)}
  \footnotesize
  \begin{center}
\begin{tabular}[c]{|r|p{3cm}|r|r|r|r|r|r|p{5cm}|}\hline
  & Condition & \multicolumn{2}{c|}{Indef} & \multicolumn{2}{c|}{Def} & \multicolumn{2}{c|}{Gener} & Example\\\hline
8 &
When a noun is accompanied by a particle (WA), and the 
predicate is not in the past tense, &
1 & 0 & 
1 & 2 & 
1 & 3 & 
\underline{DAIGAKU}-WA KOUDO-NO KYOIKU-WO UKERU TOKORO-DESU. (\underline{A college} is an institution of higher learning.) \\
9 &
When a noun is followed by ``NIWA (topic)'' or ``DEWA (topic)'', &
1 & 0 & 
1 & 2 & 
1 & 2 & 
MAINICHI CHUUSHOKU-NO TOKI-NIWA \underline{BIJINESUKAI}-NIWA NAGOYAKANA HITOTOKI-GA ARIMASU. (There is a bit of the piece of \underline{ the business world} every day at lunch time.) \\
10 &
When a noun is followed by ``GA (subject)'', &
1 & 2 & 
1 & 1 & 
1 & 0 & 
KARE-NO ME-NO NAKA-NIWA \underline{KANASHIMI}-GA ARIMASHITA. (There was \underline{sadness} in his eyes.) \\
11 &
When a noun has a coordinate noun followed by ``GA'', &
1 & 2 & 
1 & 1 & 
1 & 0 & 
HITORI-NO OTOKO-NO \underline{HITO}-TO HITORI-NO ONNA-NO HITO-GA ANATA-NO GAISHUTSUCHUU-NI TAZUNETE KIMASHITA. (A \underline{man} and a woman came to see you when you were gone.) \\
12 &
When a noun is modified by a pronoun, &
0 & 0 & 
1 & 3 & 
0 & 0 & 
SONO \underline{JIKO}-GA HASSEI-SHITE-KARA YAJIUMA-GA ATSUMATTE KIMASHITA. (A crowd gathered after the \underline{accident}.)\\
13 &
When a noun is modified by ``SUBETENO (all)'', &
1 & 0 & 
1 & 0 & 
1 & 2 & 
SUBETE-NO \underline{GEIJUTSUKA}-GA UTSUKUSHII MONO-WO BYOUSHA SHIYOU-TO SURU-TOWA KAGIRIMASEN. (Not all \underline{artists} seek to portray the beautiful.) \\\hline
\end{tabular}
\end{center}
\end{table}

\begin{table}[p]
\center{Table \ref{app:tab:rule_ref}: Rule for referential property (cont.)}
  \footnotesize
  \begin{center}
\begin{tabular}[c]{|r|p{3cm}|r|r|r|r|r|r|p{5cm}|}\hline
  & Condition & \multicolumn{2}{c|}{Indef} & \multicolumn{2}{c|}{Def} & \multicolumn{2}{c|}{Gener} & Example\\\hline
14 &
When a noun is modified by ``SUBETE-NO (all)'' 
and is followed by a particle ``GA (subject)'', &
1 & 0 & 
1 & 1 & 
1 & 2 & 
SUBETE-NO \underline{GEIJUTSUKA}-GA UTSUKUSHII MONO-WO BYOUSHA SHIYOU-TO SURU-TOWA KAGIRIMASEN. (Not all \underline{artists} seek to portray the beautiful.) \\
15 &
When a noun is modified by 
``DOKUJI-NO (of one's own)'' or ``ONAJI-NO (the same)'', &
0 & 0 & 
1 & 2 & 
0 & 0 & 
CHUUGOKUJIN-WA DOKUJI-NO \underline{MOJI}-WO HATSUMEI SHIMASHITA. (The Chinese invented their own \underline{writing system}.) \\
16 &
When a noun is adjacent to and modified by a pronoun, &
1 & 0 & 
1 & 3 & 
1 & 0 & 
KARE-NO \underline{OKUSAN}-WA FUJIWARAKE-NO SHUSSHIN-DESU. (His \underline{wife} is a Fujiwara.)\\
17 &
When a noun is modified by a pronoun, &
1 & 0 & 
1 & 2 & 
1 & 0 & 
\\
18 &
When a noun is modified by a word 
which indicates  location such as ``UE (the upper)'' and ``TONARI (the neighbor)'', &
1 & 0 & 
1 & 2 & 
1 & 0 & 
\\
19 &
When a noun is a word 
which indicates a location such as ``NEMOTO (the base)'', &
1 & 0 & 
1 & 2 & 
1 & 0 & 
\\
20 &
When a noun is ``JIKOKU (one's country)'' or ``HATSU (first)'', &
1 & 0 & 
1 & 2 & 
1 & 0 & 
\\\hline
\end{tabular}
\end{center}
\end{table}

\begin{table}[p]
\center{Table \ref{app:tab:rule_ref}: Rule for referential property (cont.)}
  \footnotesize
  \begin{center}
\begin{tabular}[c]{|r|p{3cm}|r|r|r|r|r|r|p{5cm}|}\hline
  & Condition & \multicolumn{2}{c|}{Indef} & \multicolumn{2}{c|}{Def} & \multicolumn{2}{c|}{Gener} & Example\\\hline
21 &
When a noun is modified by 
the past form of the verb $+$ ``ATO (after)'', &
1 & 0 & 
1 & 3 & 
1 & 0 & 
\\
22 &
When a noun is modified by a word 
which indicates the superlative 
such as ``MOTTOMO (the best)'' and ``ICHIBAN (the first)'', &
0 & 0 & 
1 & 2 & 
0 & 0 & 
KOKO-NI ARU KURUMA-NO NAKA-DE KORE-WA ICHIBAN TAKAI \underline{KURUMA} DESU. (This is the most expensive \underline{car} in this lot.)\\
23 &
When a noun is modified by an ordinal number, &
0 & 0 & 
1 & 2 & 
0 & 0 & 
MITTSU-NO SHIGOTO-GA ARIMASHITA-GA KARE-WA NIBANME-NO \underline{SHIGOTO}-WO HIKIUKERU KOTO-NI SHIMASHITA. (He was offered three jobs and he decided to take the second \underline{job}.)\\
24 &
When a noun is as 
``HUTATSU-NO-UCHI-NO OOKII-HOU (the bigger one of two things)'', &
0 & 0 & 
1 & 2 & 
0 & 0 & 
WATASHI-WA HUTARI KYOUDAI-NO-UCHI \underline{WAKAI} \underline{HOU}-DESU. (I am the \underline{younger} of two brothers.) \\
25 &
When a noun is modified by a past predicative clause, &
1 & 0 & 
1 & 1 & 
1 & 0 & 
KORE-WA WATASHI-GA KARE-KARA KARITA \underline{JISHO}-DESU. (This is the \underline{dictionary} that I borrowed from him.) \\
26 &
When a noun is modified by a past predicative clause 
which contains a definite noun phrase 
followed by a particle such as ``GA'' or ``WA'', &
1 & 0 & 
1 & 3 & 
1 & 0 & 
KORE-WA WATASHI-GA KARE-KARA KARITA \underline{JISHO}-DESU. (This is the \underline{dictionary} that I borrowed from him.) \\\hline
\end{tabular}
\end{center}
\end{table}

\begin{table}[p]
\center{Table \ref{app:tab:rule_ref}: Rule for referential property (cont.)}
  \footnotesize
  \begin{center}
\begin{tabular}[c]{|r|p{3cm}|r|r|r|r|r|r|p{5cm}|}\hline
  & Condition & \multicolumn{2}{c|}{Indef} & \multicolumn{2}{c|}{Def} & \multicolumn{2}{c|}{Gener} & Example\\\hline
27 &
When a noun is modified by a verb 
modified by a definite noun phrase 
followed by a particle such as ``GA'' or ``WA'', &
1 & 1 & 
1 & 3 & 
1 & 0 & 
KARE-GA WATASHI-NI KURETA \underline{JOGEN}-WA HIJOU-NI YAKUDACHI-MASHITA. (The \underline{advice} he gave me was very helpful.) \\
28 &
When a noun is modified by a verb 
which contains a definite noun phrase 
followed by a particle such as ``GA'' or ``WA'', &
1 & 0 & 
1 & 1 & 
1 & 0 & 
WATASHI-GA AGETA \underline{SHOUSASSHI}-WO MADA MOTTE IMASU-KA. (Do you still have the \underline{booklet} I gave you?) \\
29 &
When a noun is modified by a clause 
which contains a definite noun phrase 
followed by a particle such as ``NI'' or ``DE'', &
1 & 0 & 
1 & 1 & 
1 & 0 & 
KOKO-NI ARU \underline{KURUMA}-NO NAKA-DE KORE-WA ICHIBAN TAKAI KURUMA-DESU. (This is the most expensive car of all \underline{the cars} in this lot.) \\
30 &
When a noun is modified by a verb ``ARU'' 
which contains a definite noun phrase 
followed by a particle ``NI'' or ``DE'', &
1 & 0 & 
1 & 1 & 
1 & 0 & 
KOKO-NI ARU \underline{KURUMA}-NO NAKA-DE KORE-WA ICHIBAN TAKAI KURUMA-DESU. (This is the most expensive car of all \underline{the cars} in this lot.) \\
31 &
When a noun is modified by a verb 
modified by a definite noun phrase 
followed by a particle ``GA'' or ``NO'', &
1 & 0 & 
1 & 2 & 
1 & 0 & 
\\\hline
\end{tabular}
\end{center}
\end{table}

\begin{table}[p]
\center{Table \ref{app:tab:rule_ref}: Rule for referential property (cont.)}
  \footnotesize
  \begin{center}
\begin{tabular}[c]{|r|p{3cm}|r|r|r|r|r|r|p{5cm}|}\hline
  & Condition & \multicolumn{2}{c|}{Indef} & \multicolumn{2}{c|}{Def} & \multicolumn{2}{c|}{Gener} & Example\\\hline
32 &
When a noun is adjacent to and modified by 
a definite noun followed by a particle ``NO'', &
1 & 0 & 
1 & 1 & 
1 & 0 & 
KARE-WA SONO BENGOSHI-NO \underline{MUSUKO}-NO HITORI-DESU. (He is one of \underline{the sons} of that lawyer.) \\
33 &
When a noun is modified by 
a definite noun followed by a particle ``NO'', &
1 & 0 & 
1 & 1 & 
1 & 0 & 
KARE-WA SONO BENGOSHI-NO \underline{MUSUKO}-NO HITORI-DESU. (He is one of \underline{the sons} of that lawyer.) \\
34 &
When a noun is modified by 
an expression containing a pronoun, &
1 & 0 & 
1 & 1 & 
1 & 0 & 
SEKIYU JIGYOU-WA WATASHI-GA TE-WO DASHITAKU-NAI \underline{JIGYOU}-NO HITOTSU-DESU. (The oil business is a \underline{business} that I don't wish to get into.)\\
35 &
When a noun is followed by a particle 
``MADE (to)'', ``KARA (from)'', or ``HE (to)'', &
1 & 0 & 
1 & 2 & 
1 & 0 & 
SHIAWASE-SOUNA DAIANA-JOU-WA KEKKON-SHIKI-GA OWARU-TO \underline{JIIN}-KARA DETE KIMASHITA. (A radiant Lady Diana came out of the \underline{cathedral} after the wedding.) \\
36 &
When a noun is followed by a particle 
``GA'', ``MADE'', ``KARA'', or ``HE'', 
and the topic of the sentence is a person's name, &
1 & 0 & 
1 & 2 & 
1 & 0 & 
SHIAWASE-SOU-NA DAIANA-JOU-WA KEKKON-SHIKI-GA OWARU-TO \underline{JIIN}-KARA DETE KIMASHITA. (A radiant Lady Diana came out of the \underline{cathedral} after the wedding.) \\
37 &
When a noun has a coordinate noun followed by a particle ``MADE'', 
``KARA'' or ``HE'', &
1 & 0 & 
1 & 2 & 
1 & 0 & 
\\\hline
\end{tabular}
\end{center}
\end{table}

\begin{table}[p]
\center{Table \ref{app:tab:rule_ref}: Rule for referential property (cont.)}
  \footnotesize
  \begin{center}
\begin{tabular}[c]{|r|p{3cm}|r|r|r|r|r|r|p{5cm}|}\hline
  & Condition & \multicolumn{2}{c|}{Indef} & \multicolumn{2}{c|}{Def} & \multicolumn{2}{c|}{Gener} & Example\\\hline
38 &
When a noun is followed by ``YOU (for)'',&
1 & 0 & 
1 & 0 & 
1 & 2 & 
SOUGON-NA FUJISAN-WA TAKUSAN-NO \underline{RYOKOU}YOU-NO PANHURETTO-NI NIHON-NO SHOUCHOU-TO SHITE DETE IMASU. (A majesty Mt.Fuji appears as a symbol of Japan on many brochures for \underline{travel}.)\\
39 &
When a noun is a clause 
containing a generic noun phrase 
followed by a particle ``WA'' 
and is not a pronoun or 
a numeral, &
1 & 0 & 
1 & 0 & 
1 & 2 & 
DAIGAKU-WA KOUDO-NO \underline{KYOUIKU}-WO UKERU TOKORO-DESU.(A college is an institution of higher \underline{learning}.)\\
40 &
When a noun is followed by a particle ``WA'' 
and it modifies an adjective, &
1 & 0 & 
1 & 3 & 
1 & 4 & 
KONO HEYA-NI HAITTE-KURU \underline{KUUKI}-WA TSUMETAI-DESU. (The \underline{air} that is being blown into this room is cold.) \\
41 &
When a noun is followed by a particle ``YORI'' 
and modifies an adjective, &
1 & 0 & 
1 & 3 & 
1 & 5 & 
KIKAI-DE SEIHUN-SARETA \underline{KONA}-YORI ISHIUSU-DE TSUKURARETA KONA-NO HOU-GA ANATA-NIWA IINO-DESU. (Stone grand flour is better for you than machine processed \underline{flour}.)\\
42 &
When a noun is followed by a particle ``GA'' 
and modifies an adjective ``YOI (good)'', &
1 & 0 & 
1 & 3 & 
1 & 6 & 
KIKAI-DE SEIHUN-SARETA KONA-YORI ISHIUSU-DE TSUKURARETA \underline{KONA-NO HOU}-GA ANATA-NIWA YOINO-DESU. (Stone grand \underline{flour} is better for you than machine processed flour.)\\\hline
\end{tabular}
\end{center}
\end{table}

\begin{table}[p]
\center{Table \ref{app:tab:rule_ref}: Rule for referential property (cont.)}
  \footnotesize
  \begin{center}
\begin{tabular}[c]{|r|p{3cm}|r|r|r|r|r|r|p{5cm}|}\hline
  & Condition & \multicolumn{2}{c|}{Indef} & \multicolumn{2}{c|}{Def} & \multicolumn{2}{c|}{Gener} & Example\\\hline
43 &
When a noun is followed by a particle ``GA'' 
and modifies an adjective ``SUKIDA (like)'', &
1 & 0 & 
1 & 2 & 
1 & 3 & 
\\
44 &
When a noun is followed by a particle ``WO'' 
and modifies a verb ``TANOSHIMU (enjoy)'', &
1 & 0 & 
1 & 2 & 
1 & 3 & 
OITA JONSON-HUJIN-WA SOUCHO-NO \underline{SANPO}-WO TANOSHIMI-MASU. (Old Mrs Johnson enjoys her early morning \underline{walks}.) \\
45 &
When a noun is ``HOU (be more ... than ...)'' 
and modifies an adjective, &
1 & 0 & 
1 & 1 & 
1 & 4 & 
KIKAI-DE SEIHUN-SARETA KONA-YORI ISHIUSU-DE TSUKURARETA \underline{KONA-NO HOU}-GA ANATA-NIWA IINO-DESU. (Stone ground \underline{flour} is better for you than machine processed flour.) \\
46 &
When a noun is followed by a particle ``TOWA'' or ``TOIUNOWA'' 
which easily follows a generic noun phrase, &
0 & 0 & 
1 & 0 & 
1 & 2 & 
HONTOU-NO \underline{SHINSHI}-TO IU-NOWA SHUKUJO-NI ITSUMO SHINSETSU-DESU. (The perfect \underline{gentleman} is always courteous to a lady.)\\
47 &
When a noun is followed by a particle ``WA'' or ``MO'' 
and modifies a verb modified by an adverb 
such as ``ITSUMO (always)'' and ``IPPAN (generally)'', &
0 & 0 & 
1 & 0 & 
1 & 2 & 
\underline{SHINSHI}-WA HUTSUU SHUKUJO-NO TAME-NI DOA-WO AKEMASU. (The \underline{gentleman} usually opens the door for the lady.) \\\hline
\end{tabular}
\end{center}
\end{table}

\begin{table}[p]
\center{Table \ref{app:tab:rule_ref}: Rule for referential property (cont.)}
  \footnotesize
  \begin{center}
\begin{tabular}[c]{|r|p{3cm}|r|r|r|r|r|r|p{5cm}|}\hline
  & Condition & \multicolumn{2}{c|}{Indef} & \multicolumn{2}{c|}{Def} & \multicolumn{2}{c|}{Gener} & Example\\\hline
48 &
When a noun is followed by a particle ``WA'' or ``MO'' 
and modifies a verb modified by an adverb 
such as ``DENTOU (traditionally)'', &
0 & 0 & 
1 & 0 & 
1 & 2 & 
\\
49 &
When a noun is followed by a particle ``WA'' or ``MO'' 
and modifies a verb modified by a word 
such as ``MUKASHI-WA (in earlier times)'' and 
``IMA-WA (at present)'', &
0 & 0 & 
1 & 0 & 
1 & 2 & 
\\
50 &
When a noun is followed by a particle ``WA'' or ``MO'' 
and modifies a verb modified by a word 
such as ``MUKASHI (in earlier times)'' and 
``IMA (at present)'', &
0 & 0 & 
1 & 0 & 
1 & 2 & 
\\
51 &
When a noun is followed by a particle ``WA'' or ``MO'' 
and modifies a verb modified by a word 
followed by ``DEWA (topic)'', &
0 & 0 & 
1 & 0 & 
1 & 2 & 
\\\hline
\end{tabular}
\end{center}
\end{table}

\begin{table}[p]
\center{Table \ref{app:tab:rule_ref}: Rule for referential property (cont.)}
  \footnotesize
  \begin{center}
\begin{tabular}[c]{|r|p{3cm}|r|r|r|r|r|r|p{5cm}|}\hline
  & Condition & \multicolumn{2}{c|}{Indef} & \multicolumn{2}{c|}{Def} & \multicolumn{2}{c|}{Gener} & Example\\\hline
52 &
When a noun is followed by a particle ``WA'', ``MO'', or ``GA'' 
and modifies a verb ``DEKIRU (can)'' or 
a noun followed by a copula ``DA (be)'', &
1 & 0 & 
1 & 2 & 
1 & 4 & 
\underline{RAKUDA}-WA MIZU-WO NOMANAKU-TEMO NAGAI AIDA ARUKU-KOTO-GA DEKIMASU. (A \underline{camel} can go for a long time without water.) \\
53 &
When a noun is followed by a particle ``WA'', ``MO'', or ``GA'' 
and modifies a progressive form of a verb, &
1 & 2 & 
1 & 2 & 
1 & 0 & 
\underline{KURUMA}-WA MICHI-NO WAKI-NI CHUUSHA-SHITE ARIMASU. (\underline{Cars} are parked along the street.)\\
54 &
When a noun modifies a verb modified by a word such as 
``ITSUMO (always)'' and ``IPPAN (generally)'', &
1 & 0 & 
1 & 1 & 
1 & 2 & 
NIHON-DEWA \underline{SHINDA HITO}-WA TAITEI KASOU SAREMASU. (In Japan, \underline{the dead} are usually cremated.) \\
55 &
When a noun is a common noun or a verbal noun, &
1 & 1 & 
1 & 0 & 
1 & 0 & 
KANOJO-WA TEEBURU-NO HOKORI-WO TORINOZOKU-TAME-NI \underline{HUKIN}-WO TSUKAI-MASHITA.(She used a \underline{cloth} to dust the table.)\\
56 &
When a noun is followed by ``DEWA-NAI (be not)'', &
1 & 4 & 
1 & 2 & 
1 & 0 & 
\\\hline
\end{tabular}
\end{center}
\end{table}

\begin{table}[p]
\center{Table \ref{app:tab:rule_ref}: Rule for referential property (cont.)}
  \footnotesize
  \begin{center}
\begin{tabular}[c]{|r|p{3cm}|r|r|r|r|r|r|p{5cm}|}\hline
  & Condition & \multicolumn{2}{c|}{Indef} & \multicolumn{2}{c|}{Def} & \multicolumn{2}{c|}{Gener} & Example\\\hline
57 &
When a noun is ``BAAI (when)'', ``TOKORO (where)'' and ``KOTO (that)'', &
1 & 1 & 
1 & 1 & 
1 & 0 & 
SHITSUBOU-SHITA FOUDO-DAITOURYOU-WA JIBUN-GA DAITOURYOU SENKYO-NI YABURETA \underline{KOTO}-WO MITOME-MASHITA. (A disappointed President Ford admitted \underline{that} he was defeated in the election.)\\
58 &
When a noun is modified by an adjective ``ARU (a certain)'', &
1 & 2 & 
0 & 0 & 
0 & 0 & 
ARU \underline{GAKUDAN}-WA SONO KOUEN-DE ONGAKU-WO ENSOU SHIMASHITA. (A \underline{band} gave a performance at the park.)\\
59 &
When a noun is modified by a word 
such as ``HOKA-NO (other)'' and ``BETSU-NO (another)'', &
1 & 2 & 
0 & 0 & 
0 & 0 & 
\\
60 &
When a noun is followed by a copula ``DA (be)'' 
and it is not modified by a generic noun phrase 
followed by a particle ``WA'', &
1 & 1 & 
1 & 0 & 
1 & 1 & 
KARE-WA SONO-BENGOSHI-NO MUSUKO-DESU. 
(He is a son of that lawyer.)\\
61 &
What a noun is followed by a copula ``DA (be)'' 
and is modified by a generic noun phrase 
followed by a particle ``WA'', &
1 & 1 & 
1 & 0 & 
1 & 1 & 
INU-WA YAKU-NI TATSU \underline{DOUBUTSU}-DESU. (A dog is an useful \underline{animal}.)\\\hline
\end{tabular}
\end{center}
\end{table}

\begin{table}[p]
\center{Table \ref{app:tab:rule_ref}: Rule for referential property (cont.)}
  \footnotesize
  \begin{center}
\begin{tabular}[c]{|r|p{3cm}|r|r|r|r|r|r|p{5cm}|}\hline
  & Condition & \multicolumn{2}{c|}{Indef} & \multicolumn{2}{c|}{Def} & \multicolumn{2}{c|}{Gener} & Example\\\hline
62 &
When a noun is followed by a copula ``DA (be)'' 
and is not modified by a generic noun phrase 
followed by a particle ``WA'', &
1 & 2 & 
1 & 0 & 
1 & 1 & 
\\
63 &
When a noun is modified by a numeral, &
1 & 10 & 
0 & 0 & 
0 & 0 & 
SONO RESUTORAN-DEWA ICHINICHI-NI HITO-HUKURO-NO \underline{TAMANEGI}-WO TSUKAIMASU. (That restaurant uses a bag of \underline{onions} a day.)\\
64 &
When a noun is a numeral 
and is not followed by a particle ``WA'', &
1 & 10 & 
0 & 0 & 
0 & 0 & 
KARE-WA SONO BENGOSHI-NO MUSUKO-NO \underline{HITORI}-DESU. (He is \underline{one} of the sons of that lawyer.)\\
65 &
When a noun is a numeral 
and is not followed by a particle ``WA'', &
1 & 4 & 
1 & 0 & 
1 & 0 & 
KARE-WA SONO BENGOSHI-NO MUSUKO-NO \underline{HITORI}-DESU. (He is \underline{one} of the sons of that lawyer.)\\
66 &
When a noun is modified a bunsetsu 
followed by a particle ``TOIU (called)'', &
1 & 2 & 
1 & 0 & 
1 & 0 & 
KURASU-NI IKEDA-TOIU \underline{HITO}-GA HITORI IRU. (We have \underline{one person} called Ikeda in our class.)\\
67 &
When a noun is followed by a particle 
``WA'', ``MO'', ``GA'', or ``WO'', 
and it modifies a verb modified by a numeral, &
1 & 10 & 
1 & 0 & 
1 & 0 & 
SONO IE-NIWA \underline{SHININ}-GA HITORI DEMASHITA. (There was a \underline{death} in the family.)\\\hline
\end{tabular}
\end{center}
\end{table}

\begin{table}[p]
\center{Table \ref{app:tab:rule_ref}: Rule for referential property (cont.)}
  \footnotesize
  \begin{center}
\begin{tabular}[c]{|r|p{3cm}|r|r|r|r|r|r|p{5cm}|}\hline
  & Condition & \multicolumn{2}{c|}{Indef} & \multicolumn{2}{c|}{Def} & \multicolumn{2}{c|}{Gener} & Example\\\hline
68 &
When the same noun appears previously in the same sentence 
and is indefinite, &
1 & 0 & 
1 & 2 & 
1 & 1 & 
KARE-WA JOUYOUSHA-TO TORAKKU-WO ICHIDAI-ZUTSU MOTTE IMASU-GA KARE-WA \underline{JOUYOUSHA}-NI-SHIKA HOKEN-WO KAKETE IMASEN.(He has a car and a truck but only the \underline{car} is insured.)\\
69 &
When the same noun appears previously in the same sentence 
and is definite, &
1 & 0 & 
1 & 4 & 
1 & 2 & 
\\
70 &
When the same noun appears previously in the same sentence 
and is generic, &
1 & 0 & 
1 & 3 & 
1 & 2 & 
KIKAI-DE SEIHUN-SARETA KONA-YORI ISHIUSU-DE TSUKURARETA \underline{KONA-NO-} \underline{HOU}-GA ANATA-NIWA IINO-DESU. (Stone ground \underline{flour} is better for you than machine processed flour.) \\
71 &
When the same noun appears previously 
in a coordinate structure in the same sentence 
and is not generic, &
1 & 0 & 
1 & 3 & 
1 & 0 & 
KARE-WA JOUYOUSHA-TO TORAKKU-WO ICHIDAI-ZUTSU MOTTE IMASU-GA KARE-WA \underline{JOUYOUSHA}-NI-SHIKA HOKEN-WO KAKETE IMASEN.(He has a car and a truck but only the \underline{car} is insured.)\\
72 &
When the same noun appears in the previous five sentences 
and is indefinite, &
1 & 1 & 
1 & 3 & 
1 & 0 & 
\\
73 &
When the same noun appears in the previous five sentences 
and is definite, &
1 & 0 & 
1 & 4 & 
1 & 2 & 
\\\hline
\end{tabular}
\end{center}
\end{table}

\begin{table}[p]
\center{Table \ref{app:tab:rule_ref}: Rule for referential property (cont.)}
  \footnotesize
  \begin{center}
\begin{tabular}[c]{|r|p{3cm}|r|r|r|r|r|r|p{5cm}|}\hline
  & Condition & \multicolumn{2}{c|}{Indef} & \multicolumn{2}{c|}{Def} & \multicolumn{2}{c|}{Gener} & Example\\\hline
74 &
When the same noun appears in the previous five sentences 
and is generic, &
1 & 0 & 
1 & 3 & 
1 & 2 & 
\\
75 &
When the same noun appears 
in a coordinate structure 
in the previous five sentences 
and is not generic, &
1 & 0 & 
1 & 3 & 
1 & 0 & 
\\
76 &
When a noun is followed by a particle ``DE'' or ``TO'', 
it modifies a verb, 
and the noun modified by the verb 
is generic, &
1 & 0 & 
1 & 0 & 
1 & 2 & 
\underline{KIKAI}-DE SEIHUN-SARETA KONA-YORI \underline{ISHIUSU}-DE TSUKURARETA KONA-NO HOU-GA ANATA-NIWA IINO-DESU. (\underline{Stone} ground flour is better for you than \underline{machine} processed flour.)\\
77 &
When a noun is followed by a particle ``GA'' 
and modifies a clause containing a word 
such as ``ITSUMO (always)'' and ``IPPAN (generally)'', &
1 & 0 & 
1 & 1 & 
1 & 2 & 
KOKO-WA MAITOSHI \underline{KOUZUI}GA TAKUSAN OKORU TOKORO-DESU. (This is an area where there are many \underline{floods} every year.)\\
78 &
When a noun is followed by a particle ``GA'' 
and is modified by a definite noun phrase 
followed by a particle ``NO'', &
1 & 0 & 
1 & 1 & 
1 & 0 & 
KOKO-NI WATASHI-NO \underline{KIPPU}-GA ARIMASU,SHASHOU-SAN (Here is my \underline{ticket} , conductor.)\\\hline
\end{tabular}
\end{center}
\end{table}

\begin{table}[p]
\center{Table \ref{app:tab:rule_ref}: Rule for referential property (cont.)}
  \footnotesize
  \begin{center}
\begin{tabular}[c]{|r|p{4cm}|r|r|r|r|r|r|p{4cm}|}\hline
  & Condition & \multicolumn{2}{c|}{Indef} & \multicolumn{2}{c|}{Def} & \multicolumn{2}{c|}{Gener} & Example\\\hline
79 &
When a noun is ``HAIKEI-NI (background)'' or ``TAISHOU-NI (target)'' 
and follows a noun followed by a particle ``WO'', &
1 & 0 & 
1 & 0 & 
1 & 2 & 
\\
80 &
When a noun is ``HAIKEI-NI (background)'' or ``TAISHOU-NI (target)'' 
and modifies a verb modified by a noun followed by a particle ``WO'', &
1 & 0 & 
1 & 0 & 
1 & 2 & 
\\
81 &
When a noun is followed by a particle ``NO'' 
and modifies a proper noun, &
1 & 0 & 
1 & 0 & 
1 & 1 & 
\\
82 &
When a noun is followed by a particle ``NO'' 
and modifies a noun, &
1 & 0 & 
1 & 2 & 
1 & 3 & 
OOKU-NO WAKAI \underline{OTOKO}-NO HITO-TACHI-WA RIKUGUN-NI HEIEKI-SHIMASU. (Many young \underline{people} serve in the army.)\\
83 &
When a noun is followed by a particle ``TO-IU'', &
1 & 0 & 
1 & 2 & 
1 & 0 & 
KURASU-NI \underline{IKEDA-TO} \underline{IUU HITO}-GA HITORI IRU. (We have an \underline{Ikeda} in our class.)\\
84 &
When a noun is ``NANI (what)'', &
1 & 3 & 
1 & 0 & 
1 & 0 & 
\\
85 &
When a noun is followed by a particle ``NO-YOUNA (such as or like)'', &
1 & 0 & 
1 & 2 & 
1 & 3 & 
\\
86 &
When a noun is followed by a particle ``WA'' 
and modifies a numeral, &
1 & 1 & 
1 & 1 & 
0 & 0 & 
\\\hline
\end{tabular}
\end{center}
\end{table}

\clearpage

\begin{table}[p]
  \footnotesize
  \caption{Rule for number}
  \label{app:tab:rule_num}
  \begin{center}
\begin{tabular}[c]{|r|p{3cm}|r|r|r|r|r|r|p{5cm}|}\hline
  & Condition & \multicolumn{2}{c|}{Sing} & \multicolumn{2}{c|}{Plur} & \multicolumn{2}{c|}{Uncnt} & Example\\\hline
1 &
When a noun is a noun predicate, 
and the subject of the noun predicate is singular, &
1 & 3 & 
1 & 0 & 
1 & 0 & 
KARE-WA IZEN \underline{MINSHUTOU-NO TOUIN}-DE ATTA. 
(He used to be \underline{a Democrat}.)\\
2 &
When a noun is a noun predicate, 
and the subject of the noun predicate is plural, &
1 & 0 & 
1 & 3 & 
1 & 0 & 
\\
3 &
When a noun is a noun predicate, 
and the subject of the noun predicate is uncountable, &
1 & 0 & 
1 & 0 & 
1 & 3 & 
KORE-WA \underline{JUNKIN}-DESU.(This is \underline{pure gold}.)\\
4 &
When a noun is a singular pronoun such as ``KARE (he)'' and ``WATASHI (we)'', &
1 & 3 & 
0 & 0 & 
0 & 0 & 
\underline{KANOJO}-WA KEEKI-WO IKKO PIKUNIKKU-HE MOTTE YUKIMASHITA.(\underline{She} took a cake to the picnic.)\\
5 &
When a noun is a singular demonstrative 
such as ``KORE (this)'' and ``ARE (that)'', &
1 & 3 & 
1 & 0 & 
1 & 0 & 
\underline{KOKO}-NI ARU KURUMA-NO NAKA-DE KORE-WA ICHIBAN TAKAI KURUMA-DESU. (This is the most expensive car in \underline{this lot}.) \\

6 &
When a noun is ``HITORI (one person)'', ``HITOTSU (one)'', or ``IPPIKI (one)'', &
1 & 3 & 
1 & 0 & 
1 & 0 & 
KARE-WA SONO BENGOSHI-NO MUSUKO-NO \underline{HITORI}-DESU. (\underline{He} is \underline{one} of the sons of that lawyer.)\\
7 &
When a noun is a singular numeral, &
1 & 3 & 
1 & 0 & 
1 & 0 & 
WATASHI-WA KONO KINJO-NO \underline{ICHI-KAZOKU}-SAE SHIRIMASEN. (I don't know \underline{a family} in this neighborhood.)\\\hline
\end{tabular}
\end{center}
\end{table}

\begin{table}[p]
\center{Table \ref{app:tab:rule_num}: Rule for number (cont.)}
  \footnotesize
  \begin{center}
\begin{tabular}[c]{|r|p{3cm}|r|r|r|r|r|r|p{5cm}|}\hline
  & Condition & \multicolumn{2}{c|}{Sing} & \multicolumn{2}{c|}{Plur} & \multicolumn{2}{c|}{Uncnt} & Example\\\hline
8 &
When a noun is not generic, &
1 & 1 & 
1 & 0 & 
1 & 0 & 
KARE-WA SONO \underline{BENGOSHI}-NO MUSUKO-NO HITORI-DESU. (He is one of the sons of that \underline{lawyer}.)\\
9 &
When a noun is definite, &
1 & 1 & 
1 & 0 & 
1 & 0 & 
KARE-WA SONO \underline{BENGOSHI}-NO MUSUKO-NO HITORI-DESU. (He is one of the sons of that \underline{lawyer}.)\\
10 &
When a noun is modified by 
a demonstrative adjective 
such as ``SONO (the)'', ``ANO (of that)'' and ``KONO (of this)'', &
1 & 3 & 
1 & 0 & 
1 & 1 & 
KARE-WA SONO \underline{BENGOSHI}-NO MUSUKO-NO HITORI-DESU. (He is a son of the \underline{lawyer}.)\\
11 &
When a noun is modified by 
``HITORI (one person)'', ``HITOTSU (one)'', or ``IPPIKI (one)'', &
1 & 3 & 
1 & 0 & 
1 & 0 & 
KURASU-TOWA JUGYOU-WO ISSHO-NI TOTTE-IRU GAKUSEI-TACHI-NO HITOTSU-NO \underline{GURUUPU}-DESU.(A class is a \underline{group} of students taking a course together.)\\
12 &
When a noun is modified by 
a singular numeral, &
1 & 3 & 
0 & 0 & 
0 & 0 & 
SONO RESUTORAN-DEWA ICHINICHI-NI HITO-HUKURO-NO \underline{TAMANEGI}-WO TSUKAIMASU. (That restaurant uses a bag of \underline{onions} a day.)\\
13 &
When a noun contains 
a prefix which is a singular numeral, &
1 & 2 & 
1 & 0 & 
1 & 0 & 
WATASHI-WA KONO KINNJO-NO \underline{ICHI-KAZOKU}-SAE SHIRIMASEN. (I don't know \underline{a family} in this neighborhood.)\\\hline
\end{tabular}
\end{center}
\end{table}

\begin{table}[p]
\center{Table \ref{app:tab:rule_num}: Rule for number (cont.)}
  \footnotesize
  \begin{center}
\begin{tabular}[c]{|r|p{4cm}|r|r|r|r|r|r|p{4cm}|}\hline
  & Condition & \multicolumn{2}{c|}{Sing} & \multicolumn{2}{c|}{Plur} & \multicolumn{2}{c|}{Uncnt} & Example\\\hline
14 &
When a noun is followed by a particle 
``WA'', ``WO'', ``GA'', or ``MO'', and 
modifies a verb modified by a singular numeral, &
1 & 1 & 
1 & 0 & 
1 & 0 & 
KANOJO-WA \underline{KEEKI}-WO IKKO PIKUNIKKU-HE MOTTE YUKIMASHITA.(She took a \underline{cake} to the picnic.)\\
15 &
When a noun is followed by a particle 
``WA'', ``WO'', ``GA'', or ``MO'', and 
modifies a verb modified by a singular numeral, &
1 & 1 & 
1 & 0 & 
1 & 0 & 
SANDOICCHI-NI \underline{NIKU}-GA HITOKIRE HOSHII-DESU. (I'd like a slice of \underline{meat} on my sandwich.)\\
16 &
When a noun is as ``HITOBITO (people)'', &
0 & 0 & 
1 & 3 & 
0 & 0 & 
\\
17 &
When a noun is modified by a word ``SUBETE-NO (all)'', &
0 & 0 & 
1 & 2 & 
1 & 0 & 
SUBETE-NO \underline{GEIJUTSUKA}-GA UTSUKUSHII MONO-WO BYOUSHA SHIYOU-TO SURU-TOWA KAGIRIMASEN. (Not all \underline{artists} seek to portray beautiful-things.) \\
18 &
When a noun is modified by a plural numeral, &
0 & 0 & 
1 & 3 & 
0 & 0 & 
\\
19 &
When a noun is modified by a plural numeral, &
0 & 0 & 
1 & 3 & 
0 & 0 & 
KARE-WA ISSEN-NIN-NO \underline{CHOUSHUU}-NI ENZETSU-WO SHIMASHITA. (He gave a speech to an audience of 1,000 \underline{people}.)\\
20 &
When a noun is a plural numeral, &
0 & 0 & 
1 & 2 & 
0 & 0 & 
\\\hline
\end{tabular}
\end{center}
\end{table}

\begin{table}[p]
\center{Table \ref{app:tab:rule_num}: Rule for number (cont.)}
  \footnotesize
  \begin{center}
\begin{tabular}[c]{|r|p{4cm}|r|r|r|r|r|r|p{4cm}|}\hline
  & Condition & \multicolumn{2}{c|}{Sing} & \multicolumn{2}{c|}{Plur} & \multicolumn{2}{c|}{Uncnt} & Example\\\hline
21 &
When a noun is a plural numeral, &
0 & 0 & 
1 & 2 & 
0 & 0 & 
\\
22 &
When a noun is a plural pronoun, &
0 & 0 & 
1 & 3 & 
0 & 0 & 
KAZOKU-NO HITOBITO-GA \underline{WAREWARE}-WO TAZUNE-NI KIMASHITA. (A family came to visit \underline{us}.)\\
23 &
When a noun is followed by a suffix 
which indicates plurality  
such as ``TACHI'' and ``RA'', &
1 & 0 & 
1 & 3 & 
0 & 0 & 
ISHA-WA \underline{BYOUNIN-TACHI}-NO SEWA-WO SHIMASU. (Doctors take care of \underline{patients}.)\\
24 &
When a noun is followed by a particle ``DE'' 
and modifies a verb modified by a generic noun phrase 
followed by a particle ``WA'', &
1 & 0 & 
1 & 2 & 
1 & 1 & 
NUNOJI-WA \underline{SENSHOKU-KOUJOU}-DE TSUKURARE-MASU. (Cloth is produced by \underline{textile mills}.)\\
25 &
When a noun is followed by a particle ``WA'' or ``GA'', 
and 
modifies a verb such as ``KOERU (be over)'', ``KOSU (be over)'', and ``TASSURU (amount to)'', &
1 & 0 & 
1 & 3 & 
1 & 0 & 
\\
26 &
When a noun is followed by a particle ``WO'' 
and modifies a verb ``ATSUMERU (gather)'', &
0 & 0 & 
1 & 3 & 
0 & 0 & 
\\
27 &
When a noun is followed by a particle ``GA'' 
and modifies a verb such as 
``ATSUMARU (come together)'' and ``RANRITSU SURU (be flooded)'', &
0 & 0 & 
1 & 3 & 
0 & 0 & 
SONO JIKO-GA HASSEI-SHITE-KARA \underline{YAJIUMA}-GA ATSUMATTE KIMASHITA. (A \underline{crowd} gathered after the accident.)\\\hline
\end{tabular}
\end{center}
\end{table}

\begin{table}[p]
\center{Table \ref{app:tab:rule_num}: Rule for number (cont.)}
  \footnotesize
  \begin{center}
\begin{tabular}[c]{|r|p{4cm}|r|r|r|r|r|r|p{4cm}|}\hline
  & Condition & \multicolumn{2}{c|}{Sing} & \multicolumn{2}{c|}{Plur} & \multicolumn{2}{c|}{Uncnt} & Example\\\hline
28 &
When a noun is followed by a particle ``WO'' 
and modifies a verb such as 
``SAITEN SURU (mark)'' and ``MOTARASU (bring)'', &
1 & 0 & 
1 & 2 & 
1 & 0 & 
\\
29 &
When a noun is followed by a particle ``WO'' or ``NI'' 
and 
modifies a verb modified by 
``IKURADEMO (as much ...)'' or ``NANKAIDEMO (as many times as ...)'', &
1 & 0 & 
1 & 2 & 
1 & 0 & 
\\
30 &
When a noun is followed by a particle 
``WA'', ``WO'', ``GA'', or ``MO'', 
and modifies a verb 
modified a plural noun, &
1 & 0 & 
1 & 2 & 
1 & 0 & 
WATASHI-WA SENSHUU \underline{HON}-WO NISATSU YOMIMASHITA. (I read two \underline{books} last week.)\\
31 &
When a noun is followed by a particle ``WA'', ``WO'', ``GA'', or ``MO'', 
and modifies a verb modified by a plural noun, &
1 & 0 & 
1 & 2 & 
1 & 0 & 
\\
32 &
When a noun is followed by a particle ``WA'', ``WO'', ``GA'', or ``MO'', 
and it modifies a verb modified by ``OOZEI'' etc., &
1 & 0 & 
1 & 2 & 
1 & 0 & 
\\
33 &
When a noun is Noun X in 
``Noun X NO HITORI (one of Noun X)'', &
0 & 0 & 
1 & 3 & 
0 & 0 & 
KARE-WA SONO BENGOSHI-NO \underline{MUSUKO}-NO HITORI-DESU. (He is one of \underline{the sons} of that lawyer.)\\\hline
\end{tabular}
\end{center}
\end{table}

\begin{table}[p]
\center{Table \ref{app:tab:rule_num}: Rule for number (cont.)}
  \footnotesize
  \begin{center}
\begin{tabular}[c]{|r|p{4cm}|r|r|r|r|r|r|p{4cm}|}\hline
  & Condition & \multicolumn{2}{c|}{Sing} & \multicolumn{2}{c|}{Plur} & \multicolumn{2}{c|}{Uncnt} & Example\\\hline
34 &
When a noun follows 
``... NO ICHIBU (part of)'' or ``... NO UCHINO (of)'', &
1 & 0 & 
1 & 3 & 
1 & 2 & 
\\
35 &
When a noun is followed by a particle ``GA'' 
and modifies a verb ``SUKIDA (like)'', &
1 & 0 & 
1 & 2 & 
1 & 0 & 
\\
36 &
When a noun is followed by a particle ``WO'' 
and modifies a verb ``TANOSHIMU (enjoy)'', &
1 & 0 & 
1 & 2 & 
1 & 0 & 
OITA JONSON-HUJIN-WA SOUCHO-NO \underline{SANPO}-WO TANOSHIMI-MASU. (Old Mrs Johnson enjoys her early morning \underline{walks}.)\\
37 &
When a noun is an uncountable noun 
which does not have a modifier, &
1 & 0 & 
1 & 0 & 
1 & 3 & 
RAKUDA-WA \underline{MIZU}-WO NOMANAKU-TEMO NAGAI AIDA ARUKU-KOTO-GA DEKIMASU. (A camel can go for a long time without \underline{water}.)\\
38 &
When a noun is an uncountable noun such as water, &
1 & 0 & 
1 & 0 & 
1 & 2 & 
RAKUDA-WA \underline{MIZU}-WO NOMANAKU-TEMO NAGAI AIDA ARUKU-KOTO-GA DEKIMASU. (A camel can go for a long time without \underline{water}.)\\
39 &
When a noun is an uncountable noun 
modified by ``HODO-NO (extent)'' or ``... TEKI-DA (-cal)'', &
1 & 2 & 
1 & 2 & 
1 & 0 & 
KANOJO-WA SONO MOUJIN-GA WASURE-RARE-NAI HODO-NO MAGOKORO-NO KOMOTTA \underline{SHINSETSU}-WO SONO MOUJIN-NI SHITE YARIMASHITA. (She showed a \underline{kindness} toward the blind man that he never forget.)\\\hline
\end{tabular}
\end{center}
\end{table}

\begin{table}[p]
\center{Table \ref{app:tab:rule_num}: Rule for number (cont.)}
  \footnotesize
  \begin{center}
\begin{tabular}[c]{|r|p{4cm}|r|r|r|r|r|r|p{4cm}|}\hline
  & Condition & \multicolumn{2}{c|}{Sing} & \multicolumn{2}{c|}{Plur} & \multicolumn{2}{c|}{Uncnt} & Example\\\hline
40 &
When a noun is ``MONO (thing)'' modified by an adjective, &
1 & 0 & 
1 & 0 & 
1 & 2 & 
SUBETE-NO GEIJUTSUKA-GA \underline{UTSUKUSHII MONO}-WO BYOUSHA SHIYOU-TO SURU-TOWA KAGIRIMASEN. (Not all artists to portray \underline{beautiful-things}.) \\
41 &
When a noun is followed by a particle ``WA'', ``WO'', ``GA'', or ``MO'', 
and follows an adverb 
such as ``TAKUSAN (a lot)'' and ``IPPAI (a lot)'', &
1 & 0 & 
1 & 3 & 
1 & 2 & 
\\
42 &
When a noun is followed by a particle ``WA'', ``WO'', ``GA'', or ``MO'', 
and modifies a verb modified by an adverb 
such as ``TAKUSAN (a lot)'' and ``IPPAI (a lot)'', &
1 & 0 & 
1 & 3 & 
1 & 2 & 
\\
43 &
When a noun is modified by 
``TAKUSAN-NO (a lot of)'' or ``IPPAI-NO (a lot of)'', &
0 & 0 & 
1 & 3 & 
1 & 2 & 
SOUGON-NA FUJISAN-WA TAKUSAN-NO \underline{RYOKOUYOU-NO} \underline{PANHURETTO}-NI NIHON-NO SHOUCHOU-TO SHITE DETE IMASU. (A majestic Mt.Fuji appears as a symbol of Japan on many \underline{travel brochures}.)\\
44 &
When a noun is modified by ``TAKUSAN-NO (a lot of)'', &
0 & 0 & 
1 & 3 & 
1 & 2 & 
\\
45 &
When a noun is followed by a particle ``WO'' and 
modifies a verb ``ABIRU (be covered)'', &
0 & 0 & 
1 & 2 & 
1 & 1 & 
\\\hline
\end{tabular}
\end{center}
\end{table}

\begin{table}[p]
\center{Table \ref{app:tab:rule_num}: Rule for number (cont.)}
  \footnotesize
  \begin{center}
\begin{tabular}[c]{|r|p{4cm}|r|r|r|r|r|r|p{4cm}|}\hline
  & Condition & \multicolumn{2}{c|}{Sing} & \multicolumn{2}{c|}{Plur} & \multicolumn{2}{c|}{Uncnt} & Example\\\hline
46 &
When a noun is followed by a particle ``GA'' 
and modifies a verb 
such as ``NARABU (be in line)'' 
and ``ZOKUSHUTSU SURU (appear one after another)'', &
0 & 0 & 
1 & 2 & 
1 & 1 & 
\\
47 &
When a noun is followed by a particle ``WA'' and 
modifies a noun predicate such as ``Noun X DA (be Noun X)'', 
and Noun X is plural, &
1 & 0 & 
1 & 5 & 
1 & 0 & 
\\
48 &
When a noun is followed by a particle ``WA'' 
and modifies a noun predicate such as ``Noun X DA (be Noun X)'', 
and Noun X is uncountable, &
1 & 0 & 
1 & 0 & 
1 & 6 & 
\underline{KORE}-WA JUNKIN-DESU. (\underline{This} is pure gold.)\\\hline
\end{tabular}
\end{center}
\end{table}

\chapter{Rule for Pronouns} \label{c:ap:pronoun}

\section{Rule for Demonstratives}

We made 50 {\it Candidate enumerating rule}s and 
10 {\it Candidate judging rule}s for analyzing demonstratives. 
All the rules are given below. 

\subsection{{Candidate Enumerating Rule}}

\begin{enumerate}
\item 
  When a pronoun is 
  a demonstrative followed by the particle ``GA'' 
  and 
  a non-{\it ga}-case zero pronoun is not yet recovered, 
  the system analyzes 
  the non-{\it ga}-case zero pronoun 
  before the analysis of the demonstrative. 

\item 
  When a pronoun is ``{\it so}-series demonstrative adjective + noun 
  $\alpha$,''\\
  \{
  (the noun phrase containing a noun $\alpha$, \,$45$)\\
  (the topic which is a subordinate of the noun $\alpha$ 
  and which has the weight $W$ and the distance $D$, \,$W-D*2+10$)\\
  (the focus which is a subordinate of the noun $\alpha$ 
  and which has the weight $W$ and the distance $D$, \,$W-D*2+10$)\}

\item 
  When a pronoun is ``{\it ko}-series demonstrative adjective + noun 
  $\alpha$,''\\
  \{
  (the noun phrase containing a noun $\alpha$, \,$45$)\\
  (the topic which is a subordinate of the noun $\alpha$ 
  and which has the weight $W$ and the distance $D$, \,$W-D+30$)\\
  (the focus which is a subordinate of the noun $\alpha$ 
  and which has the weight $W$ and the distance $D$, \,$W-D+30$)\}

\item 
  When a pronoun is ``{\it a}-series demonstrative adjective + noun 
  $\alpha$,''\\
  \{
  (the noun phrase containing a noun $\alpha$, \,$45$)\\
  (the topic which is a subordinate of the noun $\alpha$ 
  and which has the weight $W$ and the distance $D$, \,$W-D*0.4+30$)\\
  (the focus which is a subordinate of the noun $\alpha$ 
  and which has the weight $W$ and the distance $D$, \,$W-D*0.4+30$)\}

\item 
  When a pronoun is ``SORE (it)/ARE (that)/KORE (this)'' or 
  a demonstrative adjective and 
  the previous bunsetsu contains 
  the expression of 
  the predicative form of a verb 
  or the expression of 
  enumerating examples such as ``{TOKA} (and so on),''
  \{(the expression, \,$40$)\}

\item 
  When a pronoun is  ``SORE/ARE/KORE'' or 
  a demonstrative adjective,\\
  \{(
  The previous sentence (or the verb phrase 
  in the conditional form containing a conjunctive particle such as
  ``GA (but)'', `` DAGA (but)'', and ``KEREDO (but)''
  if the verb phrase is in the same sentence), 
   \,$15$)\}

\item 
  When a pronoun is ``KORE-WA/SORE-WA/KORE-DE/SORE-DE'', 
  is the first word of the sentence, 
  and is not a case component of a verb, \\
  \{(
  The previous sentence (or the verb phrase 
  in the conditional form containing a conjunctive particle such as
  ``GA (but)'', `` DAGA (but)'', and ``KEREDO (but)''
  if the verb phrase is in the same sentence), 
   \,$5$)\}

\item 
  When a pronoun is ``KORE-WA/SORE-WA/KORE-DE/SORE-DE'' 
  and is the first word of the sentence, \\
  \{(
  The previous sentence (or the verb phrase 
  in the conditional form containing a conjunctive particle such as
  ``GA (but)'', `` DAGA (but)'', and ``KEREDO (but)''
  if the verb phrase is in the same sentence), 
   \,$5$)\}

\item 
  When a pronoun is ``(KORE (this)/SORE (it))(HODO (extent)/DAKE (only)/DEMO (even)/KOSO (just))'', \\
  \{(
  The previous sentence (or the verb phrase 
  in the conditional form containing a conjunctive particle such as
  ``GA (but)'', `` DAGA (but)'', and ``KEREDO (but)''
  if the verb phrase is in the same sentence), 
   \,$5$)\}

\item 
  When a pronoun is ``KOUIU (like this)'', ``SOUIU (like it)'', 
  ``KON'NA (like this)'', etc., \\
  \{(
  the previous sentence (or the verb phrase 
  in the conditional form containing a conjunctive particle such as
  ``GA (but)'', `` DAGA (but)'', and ``KEREDO (but)''
  if the verb phrase is in the same sentence), 
   \,$5$)\}

\item 
  When a pronoun is ``KON'NA (like this)'', \\
  \{(the next sentences, \,$20$)\}

\item 
  When a pronoun is ``KON'NA (like this)'' and 
  ``KON'NA (like this)'' $+$ noun is followed by a particle ``NI/DE/SURA/WA/NO'', \\
  \{(the next sentences, \,$1$)\}

\item 
  When a pronoun is ``KON'NA (like this)'' and 
  ``KON'NA (like this)'' $+$ noun is followed by a particle ``WO/MO/DENAI'', \\
  \{(the previous sentences, \,$1$)\}

\item 
  When a pronoun is 
  ``(SONO (the)/KONO (this))(TAME (for)/UE (in)/ HOKA (other)/KOTO (thing)/
  BAAI (case)/TSUDO (every time))'', \\
  \{(
  the previous sentence (or the verb phrase 
  in the conditional form containing a conjunctive particle such as
  ``GA (but)'', `` DAGA (but)'', and ``KEREDO (but)''
  if the verb phrase is in the same sentence), 
   \,$30$)\}

\item 
  When a pronoun is 
  ``(SONO (its)/KONO (this))(IMI (meaning)/GEN'IN (cause)/KEKKA (result)/HAIKEI(background)/KOUKA (effect))'', \\
  \{(
  the previous sentence (or the verb phrase 
  in the conditional form containing a conjunctive particle such as
  ``GA (but)'', `` DAGA (but)'', and ``KEREDO (but)''
  if the verb phrase is in the same sentence), 
   \,$5$)\}
  \footnote{
      \label{foot:yanagi}
    This rule is based on Yanagi's method\cite{yanagi}.}

\item 
  When a pronoun is 
  ``ANO/SONO/AN'NA/SON'NA (like it)'' $+$ 
  noun which indicates time, \\
  \{(
  the previous sentence (or the verb phrase 
  in the conditional form containing a conjunctive particle such as
  ``GA (but)'', `` DAGA (but)'', and ``KEREDO (but)''
  if the verb phrase is in the same sentence), 
   \,$30$)\}

\item 
  When a pronoun is 
  ``KONO/KON'NA'' $+$ 
  noun which indicates time, \\
  \{(the present time, \,$5$)\}

\item 
  When a pronoun is 
  ``(KONO/KON'NA)(CHI (place)/ KUNI (country)/ SHAKAI (society))'', \\
  \{(the present place, \,$5$)\}

\item 
  When a pronoun is 
  ``SONO (the or its)'' 
  in ``Noun X TO SONO Noun (Noun X and the Noun)'' or 
  ``Noun X YA SONO Noun (Noun X or the Noun)'',\\
    \{(Noun X, \,$50$)\}
  $^{\ref{foot:yanagi}}$

\item 
  When a pronoun is ``SONO(its)'' in 
  ``Noun X NO(of) SONO(its) Noun'', \\
    \{(Noun X, \,$30$)\}

\item 
  When a pronoun is ``AA (oh)/SORE/KORE/ARE'' followed 
  by a comma, \\
    \{(it is regarded as an exclamation, \,$30$)\}

\item 
  When a pronoun is 
  ``SOU/KON'NA/KON'NANI/SON'NANI/SOREHODO'' 
  and 
  it modifies an adjective or an adverb,\\
  \{(Introduced as indefinite, \,$30$)\}
  $^{\ref{foot:yanagi}}$

\item 
  When 
  a pronoun is such as ``ARE-YA KORE-YA'', \\
  \{(an idiomatic expression, \,$50$)\}
  $^{\ref{foot:yanagi}}$

\item 
  When a pronoun is a demonstrative pronoun, 
  a demonstrative adverb, or a demonstrative adjective,\\
  \{(Introduce an individual, \,$10$)\}

\item 
  When a pronoun is a demonstrative in quotations,\\
  \{(Introduce an individual, \,$5$)\}

\item 
  When a pronoun is a {\it a}-series demonstrative, \\
  \{(Introduce an individual, \,$5$)\}

\item 
  When a pronoun is ``KOU/KON'NAHUUNISHITE/KOUSHITE'', \\
  \{(the previous sentences, \,$25$)\}

\item 
  When a pronoun is ``KOU/KON'NAHUUNISHITE/KOUSHITE'', \\
  \{(the next sentences, \,$26$)\}

\item 
When a pronoun is a part of 
  ``KOU/KON'NAHUUNI'' + conditional form or ``KOU SHITE'' 
  and is not the last word in the sentence,\\
  \{(the previous sentences, \,$7$)\}

\item 
When a pronoun is ``KON'NA HUUNI (like this)'', 
  and is not the last word in the sentence,\\
  \{(the previous sentences, \,$2$)\}

\item 
When a pronoun is ``KOUDA'' or ``KON'NA-HUUDAN'', \\
  \{(the next sentences, \,$3$)\}

\item 
  When a pronoun is 
  a demonstrative which does not indicate location 
  and 
  the previous sentence is a quotation, 
  \{(the previous sentences, \,$3$)\}

\item 
  When a pronoun is 
  a demonstrative which does not indicate location, \\
  \{(the previous sentences, \,$1$)\}

\item 
  When a pronoun is 
    a demonstrative which does not indicate location 
    and 
    the next sentence is a quotation, \\
    \{(the next sentences, \,$3$)\}

\item 
  When a pronoun is 
  a demonstrative which does not indicate location,\\
    \{(the next sentences, \,$1$)\}

\item 
  What a pronoun is ``AA (like that)'', \\
  \{(the previous sentence, \,$20$)\}

\item 
  When an anaphora is ``SOU (so)/SOUSHITE (do so)/SONOYOUNI (like it)'',\\
  \{(the previous sentences, \,$30$)\}

\item 
  When an anaphor is ``SOU/SOUSHITE/SONOYOUNI'' 
  and
  is in the subordinate clause 
  which has a conjunctive particle such as
  ``GA (but)'', `` DAGA (but)'', and `` KEREDO (but)'' 
  or an adjective conjunction such as ``YOUNI (as)'',\\
  \{(the main clause, \,$45$)\}
  \footnote{
    \label{foot:matsuoka}
    This rule is based on Matsuoka's method\cite{matsuoka_nl}.}

\item 
  When a pronoun is 
  ``KON'NANI/AN'NANI/SON'NA-HUUNI/AN'NA- HUUNI'' 
  and 
  does not modify an adjective or an adverb,\\
  \{(the previous sentence, \,$25$)\}

\item 
  When a pronoun is 
  ``KOKODE (here)/SOKODE (there)'' and 
  the first word of the sentence,\\
  \{(the previous sentence, \,$5$)\}

\item 
  When a pronoun is ``KOKODE (here)/SOKODE (there)'', 
  is the first word of the sentence, 
  and is not a case component of a verb,\\
  \{(the previous sentence, \,$5$)\}

\item 
  When a pronoun is ``KOKO (here)/SOKO (there)'', \\
  \{(the present place, \,$15$)\}
  
\item 
  When a pronoun is ``KOKO (here)/SOKO (there)'' $+$ noun 
  which indicates time, \\
  \{(the present time, \,$50$)\}
  
\item 
  When a pronoun is ``(ARE/KORE/SORE)(KARA (from)/MADE (to))'', \\
  \{(the present time, \,$15$)\}
  
\item 
  When a pronoun is ``KOCHIRA (this gentleman)'' and 
  is in a quotation, \\
  \{(the first person, \,$25$)\}
  
\item 
  When a pronoun is ``KOCHIRA (this gentleman)'' 
  which is not in a quotation, \\
  \{(the first person, \,$13$)\}

\item 
  When a pronoun is ``SOCHIRA (the other)'' 
  which is in a quotation, \\
  \{(the second person, \,$13$)\}

\item 
  When a demonstrative is 
  the subject of a noun/adjective predicative sentence 
  and 
  the predicate is 
  a word which signifies judgment 
  such as ``JISSEKI-DA (result)'', ``ZAN'NEN-DA (unfortunate)'', 
  ``KAKUJITSU-DA (sure)'',   and ``...TEKI-DA (-cal)'', \\
  \{(the previous sentences, \,$50$)\}
  $^{\ref{foot:yanagi}}$
  
\item 
  When a demonstrative is 
  in a subordinate clause 
  containing ``YOUNI (as)'', ``GA (but)'', and ``KEREDOMO (but)'', \\
  \{(the main clause, \,$10$)\}
  $^{\ref{foot:yanagi}}$
  $^{\ref{foot:matsuoka}}$
  
\item 
  When a pronoun is a demonstrative pronoun or 
  ``SONO (of it) / KONO (of this) / ANO (of that)'',\\
  \{(A topic which has the weight $W$ and the distance $D$, \, $W-D-2$)\\
  (A focus which has the weight $W$ and the distance $D$, \, $W-D+4$)\}

\end{enumerate}

\subsection{{Candidate Judging Rule}}

\begin{enumerate}
\item 
  When a pronoun is a demonstrative pronoun and 
  a candidate referent has a semantic marker {\sf HUM} (human), 
  it is given $-10$. 
  We use 
  Noun Semantic Marker Dictionary\cite{imiso-in-BGH} 
  as a semantic marker dictionary. 

\item 
  When a pronoun is a demonstrative pronoun, 
  a candidate referent is given the points 
  in Table \ref{tab:hininshoudaimeisi_ruijido} 
  by using the highest semantic similarity 
  between the candidate referent and 
  the codes 
  \{5200003010 5201002060 5202001020 5202006115 5241002150 5244002100\}
  in BGH \cite{BGH} which signify human beings. 
  
\item 
  When a pronoun is ``KOKO (here) / SOKO (there) / ASOKO (over there)'' 
  and a candidate referent has 
  a semantic marker {\sf LOC}, which indicates location, 
  the candidate referent is given $10$ points. 

\item 
  When a pronoun is ``KOKO/SOKO/ASOKO'', 
  a candidate referent is given the points in 
  Table \ref{tab:bashomeisi_ruijido} 
  by using the semantic similarity 
  between the candidate referent and 
  the codes 
  \{6563006010 6559005020 9113301090 9113302010 6471001030 6314020130\}
  which indicate locations in BGH \cite{BGH}. 
  
\item 
  When a pronoun is a {\it so}-series demonstrative adjective, 
  the system consults examples of 
  the form ``noun X {NO} noun Y'' 
  whose noun Y is modified by the pronoun, 
  and gives a candidate referent 
  the point 
  in Table~\ref{tab:sokei_meishi_anob_ruijido} 
  by the similarity between 
  the candidate referent and noun X. 
  The Japanese Co-occurrence Dictionary\cite{edr_kyouki_2.1} serves 
  as a source of examples for ``X NO Y''. 

\item 
  When a pronoun is a non-{\it so}-series demonstrative adjective, 
  the system consults examples of 
  the form ``Noun X {NO(of)} Noun Y (Y of X)'' 
  whose Noun Y is modified by the pronoun, 
  and gives a candidate referent 
  the point 
  in Table \ref{tab:akei_meishi_anob_ruijido}
  by the similarity between 
  the candidate referent and noun X. 

\item 
  When a candidate referent of a pronoun 
  does not satisfy the semantic marker of the case component 
  in the case frame, 
  it is given $-5$. 

\item 
  A candidate referent of a pronoun is 
  given the points in Table \ref{tab:yourei_ruijido}
  by using the highest semantic similarity 
  between the candidate referent and examples 
  of the case component in the case frame. 
  
\item 
  When a pronoun is a demonstrative 
  followed by ``GA Noun X NI-NARU (become Noun X)'', 
  it is given the points in Table \ref{tab:yourei_ruijido}
  by using the semantic similarity 
  between the candidate referent and Noun X. 

\begin{table}[t]
  \leavevmode
    \caption{Point given by the similarity of the verb}
    \label{tab:yougen_ruijido}
  \begin{center}
\begin{tabular}[c]{|l|r|r|r|r|r|r|r|r|}\hline
Similarity level & 0 & 1 & 2 & 3 & 4 & 5 & 6 & Exact Match\\\hline
Point  & 0 & 0 & 1 & 1.5 & 2& 3 & 3.5 & 4\\\hline
\end{tabular}
\end{center}
\end{table}

\item 
  When a pronoun is given 
  the points in Table \ref{tab:yougen_ruijido}
  by using the semantic similarity 
  between 
  the verb modified by the demonstrative 
  and the verb modified by a candidate referent. 

\end{enumerate}

\section{Rule for Personal Pronouns}

We made 
4 {\it Candidate enumerating rule}s and 
6 {\it Candidate judging rule}s for analyzing personal pronouns. 
All the rules are given below. 

\subsection{{ Candidate Enumerating Rule}}

\begin{enumerate}
\item 
  When an anaphor is a first personal pronoun, \\
  \{(the first person (the speaker) in the context, \,$25$)\}

\item 
  When an anaphor is a second personal pronoun, \\
  \{(the second person (the hearer) in the context, \,$25$)\}

\item 
  When an anaphor is a third personal pronoun, \\
  \{(a first person, \,$-10$) (a second person, \,$-10$)\}

\item 
  When an anaphor is a personal pronoun,\\ 
  \{(a topic which has the weight $W$ and the distance $D$,
\,$W-D-2$)\\
(a focus which has the weight $W$ and the distance $D$, \,$W-D+4$)\}
\end{enumerate}

\subsection{{Candidate Judging Rule}}

\begin{enumerate}
\item 
  When an anaphor is a personal pronoun 
  and a candidate referent has 
  a semantic marker {\sf HUM}, 
  the candidate referent is given $10$ points. 

\item 
  When an anaphor is a personal pronoun, 
  a candidate referent is given the points 
  in Table \ref{tab:ninshoudaimeisi_ruijido} 
  by using the semantic similarity 
  between the candidate referent and 
  the code 
  \{5200003010 5201002060 5202001020 5202006115 5241002150 5244002100\}
  which indicates human being in BGH\cite{BGH}. 

\item 
  When a candidate referent of a personal pronoun 
  does not satisfy the semantic marker of the case component 
  in the case frame, 
  it is given $-5$. 

\item 
  A candidate referent of a personal pronoun is 
  given the points in Table \ref{tab:yourei_ruijido}
  by using the highest semantic similarity 
  between the candidate referent and examples 
  of the case component in the case frame. 
  
\item 
  When a pronoun is a personal pronoun 
  followed by ``GA Noun X NI-NARU (become Noun X)'', 
  it is given the
  points in Table \ref{tab:yourei_ruijido}
  by using the semantic similarity 
  between the candidate referent and Noun X. 

\item 
  When a pronoun is given the
  points in Table \ref{tab:yougen_ruijido}
  by using the semantic similarity 
  between 
  the verb modified by the demonstrative 
  and the verb modified by a candidate referent. 

\end{enumerate}

\section{Rule for Zero Pronouns}

We made 
19 {\it Candidate enumerating rule}s and 
4 {\it Candidate judging rule}s for analyzing zero pronouns. 
All the rules are given below. 

\subsection{{Candidate Enumerating Rule}}

\begin{enumerate}
\item \label{enum:zero:kureru_ni_saki}
  When an anaphor is a {\it ga}-case zero pronoun 
  whose verb is followed by the auxiliary verbs 
  such as ``{KURERU}'' and ``{KUDASARU}'' 
  and there is a {\it ni}-case zero pronoun in the verb, 
  the {\it ni}-case zero pronoun is analyzed first. 
  With respect to the {\it ga}-case zero pronoun,  
  \{(do not fill a zero pronoun, \,$-5$)\}

\item \label{enum:zero:toiu}
  When a zero pronoun is not in a quotation 
  and 
  is a case component of a verb 
  whose {\it ga}-case is easily filled 
  by a first person (speaker) such as ``OMOU (think)'' and ``HOSHII (want)'', 
  \{(a first person,  \,$50$)\}

\item 
  In a quotation, when an anaphor is a {\it ga}-case zero pronoun 
  which is easily filled with a first person, 
  whose verb is such as ``{YARU} (give)'', ``{SHITAI} (want)'', and 
  ``{IKU (go)},'' 
  \{(the first person, \,$5$)\}

\item 
  When a zero pronoun is 
  a {\it ga}-case zero pronoun 
  which is not easily filled with a first person, 
  whose verb is such as ``{DAROU}'', ``{YOUDA}'', and 
  ``SOUDA'',\\
  \{(the first person, \,$-20$)\}

\item 
In a quotation, when an anaphor is a {ga}-case zero pronoun 
which is easily filled with a second person, 
whose verb is such as ``{KURERU} (give)'', ``{NASARU} (do)'', and 
``{KURU} (come)'', 
or whose verb is in an imperative sentence or 
an interrogative sentence, \\
\{(the first person, \,$-30$)(the second person, \,$25$)\}

\item 
In a quotation, when an anaphor is a {ga}-case zero pronoun, \\
\{(the first person, \,$15$)\}

\item 
  When an anaphor is a {\it ga}-case zero pronoun 
  of ``Y {DA} (is Y)'' in the expression 
  of ``X WO Y DA TO MINASU (consider X as Y)'', \\
  \{(Noun X, \,$50$)\}

\item 
  When a zero pronoun is 
  the subject of a noun predicative sentence and 
  the predicate is 
  ``KU (phrase)'', ``HAIKU (haiku)'', ``UTA (song)'' 
  and ``TANKA (tanka)'', \\
  \{(the previous sentence, \,$25$)\}
  $^{\ref{foot:yanagi}}$

\item 
  When a zero pronoun is 
  the subject of a noun predicative sentence and 
  the predicate is 
  a word which indicates time, \\
  \{(the time of the previous sentence, \,$25$)\}

\item 
  When a zero pronoun is a {\it ga}-case of the main (or subordinate) clause 
  in a complex sentence, 
  the complex sentence is connected 
  by the conjunctive particle indicating 
  disagreement of subjects in a complex sentence 
  such as ``{NODE} (because)'' and ``{NARABA} (if)'' 
  and the subject of the subordinate (or main) clause is not omitted 
  and is followed by the particle ``{GA},''\\
  \{(the subject of the subordinate (or main) clause, \,$-30$)\}

\item 
  When a zero pronoun is 
  the subject of a noun predicative sentence and 
  the predicate is 
  a word which indicates action, \\
  \{(the previous sentence, \,$21$)(the next sentence, \,$21$)\}

\item 
  When 
  the next sentence is a quotation,\\
  \{(the next sentence, \,$1$)\}

\item 
  When a zero pronoun is a {\it ga}-case component, \\
  \{(A topic which has the weight $W$ and the distance $D$, \,$W-D*2+1$)\\
  (A focus which has the weight $W$ and the distance $D$, \,$W-D+1$)\\
  (A subject of a clause coordinately connected to the clause containing the anaphor, \,25)\\
  (A subject of a clause subordinately connected to the clause containing the anaphor, \,23)\\
  (A subject of a main clause whose 
  embedded clause contains the anaphor, \,22)\}

\item 
  When a zero pronoun is not a {\it ga}-case component, \\
  \{(A topic which has the weight $W$ and the distance $D$, \,$W-D*2-3$)\\
  (A focus which has the weight $W$ and the distance $D$, \,$W-D*2+1$)\}

\item 
  When there is ``Noun $\alpha$'' in another case component 
  of the verb which has the analyzed case component 
  (the analyzed zero pronoun), \\
  \{(Noun $\alpha$, \,$-20$)\}

\item 
  When a zero pronoun is a case component 
  of a verb 
  which modifies a noun phrase 
  and is not modified by any phrase, \\
  \{(the system does not analyze the zero pronoun, \,3)\}

\item 
  When a zero pronoun is an optional case component, \\
  \{(the system does not analyze the zero pronoun, \,3)\}

\item 
  When a zero pronoun is a {\it ga}-case component, \\
  \{(the system does not analyze the zero pronoun, \,15)\}

\item 
  When a zero pronoun is not a {\it ga}-case component, \\
  \{(the system does not analyze the zero pronoun, \,18)\}
\end{enumerate}

\subsection{{ Candidate Judging Rule}}

\begin{enumerate}
\item 
  When a candidate referent of a case component (a zero pronoun) 
  does not satisfy the semantic marker of the case component 
  in the case frame, 
  it is given $-5$. 

\item 
  A candidate referent of a case component ( a zero pronoun ) is 
  given the points in Table \ref{tab:yourei_ruijido}
  by using the highest semantic similarity 
  between the candidate referent and examples 
  of the case component in the case frame. 
  
\item 
  When a zero pronoun is a subject 
  of ``GA Noun X NI-NARU (become Noun X)'', 
  it is given the points in Table \ref{tab:yourei_ruijido}
  by using the semantic similarity 
  between the candidate referent and Noun X. 
  
\item 
  When a zero pronoun is given the
  points in Table \ref{tab:yougen_ruijido}
  by using the semantic similarity 
  between 
  the verb having the zero pronoun 
  and the verb modified by a candidate referent.  

\end{enumerate}

\begin{themajorpublications}{9}

\bibitem{dummy1}
\newblock Murata,M. and Nagao,M.:
\newblock Determination of referential property and number of nouns 
in Japanese sentences for machine translation into English
\newblock {\em Proceedings of the 5th TMI\/},  
\newblock pp.218-225, 
\newblock 1993.

\bibitem{dummy2}
\newblock Murata,M. and Nagao,M.:
\newblock An Estimate of Referent of Nouns in Japanese Sentences
 with Referential Property of Nouns
\newblock (in Japanese), 
\newblock {\em Journal of ANLP\/},  
\newblock Vol.3 No.1, pp.67-81, 
\newblock 1996.

\bibitem{dummy3}
\newblock Murata,M., Kurohashi,S., and Nagao,M.:
\newblock An Estimate of Referential Property and Number 
of Japanese Noun Phrases from  Surface Expressions
\newblock (in Japanese), 
\newblock {\em Journal of ANLP\/},  
\newblock Vol.3 No.4, pp.31-48, 
\newblock 1996.

\bibitem{dummy4}
\newblock Murata,M. and Nagao,M.:
\newblock An Estimate of Referents of Pronouns in Japanese Sentences
using Examples and Surface Expressions
\newblock (in Japanese), 
\newblock {\em Journal of ANLP\/},  
\newblock to be published. 

\bibitem{dummy5}
\newblock Murata,M. and Nagao,M.:
\newblock Indirect Anaphora Resolution in Japanese Nouns
using Semantic Constraint
\newblock (in Japanese), 
\newblock {\em Journal of ANLP\/},  
\newblock to be published. 

\bibitem{dummy6}
\newblock Murata,M. and Nagao,M.:
\newblock Indirect reference in Japanese sentences
\newblock Discourse Anaphora and Anaphor Resolution Colloquium, 
\newblock Lancaster University, July, 1996


\end{themajorpublications}

\begin{theotherpublications}{99}

\bibitem{dummy1}
\newblock Murata,M., Kurohashi,S., and Nagao,M.:
\newblock An Estimate of Referential Property and Number of Nouns 
   from Japanese Surface Expressions
\newblock (in Japanese), 
\newblock {\em IEICE-WGNLC} 93-5, 
\newblock pp.33-40,
\newblock 1993.

\bibitem{dummy3}
\newblock Watanabe,Y., Takeuchi,M., Murata,M., and Nagao,M.:
\newblock Document Classification Using Important Kanji Characters 
Extracted by $\chi^2$ Method 
\newblock (in Japanese), 
\newblock {\em IEICE-WGNLC} 94-25, 
\newblock pp.23-30,
\newblock 1994.

\bibitem{dummy5}
\newblock Murata,M. and Nagao,M.:
\newblock An Estimate of Referent of Nouns in Japanese Sentences
\newblock (in Japanese), 
\newblock {\em Proceedings of The First Annual Meeting of 
ANLP}, A4-4, p.109--112, 
\newblock 1995.

\bibitem{dummy6}
\newblock Matsuoka,M., Murata,M., Kurohashi,S., and Nagao,M.:
\newblock Automatic Extraction of Cataphoric Expressions 
    Using Surface Expressions in Japanese Sentences
\newblock (in Japanese), 
\newblock {\em IPSJ-WGNL} 108-6, 
\newblock 1995.

\bibitem{dummy7}
\newblock Murata,M. and Nagao,M.:
\newblock An Estimate of Referents of Pronouns in Japanese Sentences
    using Examples and Surface Expressions
\newblock (in Japanese), 
\newblock {\em IPSJ-WGNL} 108-7, 
\newblock 1995.

\bibitem{dummy8}
\newblock Murata,M. and Nagao,M.:
\newblock Indirect Anaphora Resolution in Japanese Nouns
\newblock (in Japanese), 
\newblock {\em Proceedings of The Second Annual Meeting of 
ANLP}, C2-1, p.309--312, 
\newblock 1996.

\bibitem{dummy9}
\newblock Watanabe,Y., Murata,M., Takeuchi,M., and Nagao,M.:
\newblock Document Classification Using Domain Specific Kanji Characters
Extracted by X-2 Method
\newblock {\em Proc. of 16th COLING\/}
\newblock 1996. 

\end{theotherpublications}

\begin{flushleft} \Large \bf
Abbreviations
\end{flushleft}

\begin{description}
\item[COLING] International Conference on Computational Linguistics
\end{description}

\begin{description}
\item[IEICE] The Institute of Electronics, Information and Communication Engineers
\begin{description}
\item[WGNLC] Natural Language Processing and Models of Communication
\end{description}
\item[IPSJ] Information Processing Society of Japan
\begin{description}
\item[WGIM] Information Media
\item[WGNL] Natural Language
\end{description}
\item[JSAI] Japan Society for Artificial Intelligence
\item[ANLP] The Association for Natural Language Processing
\end{description}

\end{document}